\documentclass[twoside]{article}
\usepackage{ecj,palatino,epsfig,latexsym,natbib}

\usepackage{algorithmic}
\usepackage{algorithm}
\usepackage{array}
\usepackage{stfloats}
\usepackage{subfig}
\usepackage{amsmath,amssymb,amsfonts}
\usepackage{booktabs}
\usepackage{multirow}
\usepackage{threeparttable}
\usepackage{adjustbox}
 
\usepackage{hyperref}

\usepackage{wrapfig}

\begin{document}

\ecjHeader{x}{x}{xxx-xxx}{2025}{Enhancing Generalization and Scalability for MOPs with Population Pre-Training}{Haokai Hong, Liang Feng, Min Jiang, Kay Chen Tan}
\title{\bf Enhancing Generalization and Scalability for Multi-Objective Optimization with Population Pre-Training}  

\author{\name{\bf Haokai~Hong} \hfill \addr{haokai.hong@connect.polyu.hk}\\ 
        \addr{The Department of Data Science and Artificial Intelligence, The Hong Kong Polytechnic University, Hong Kong SAR, P.R. China}
\AND
       \name{\bf Liang~Feng} \hfill \addr{liangf@cqu.edu.cn}\\
        \addr{The College of Computer Science, Chongqing University, Chongqing 400044, China.}
\AND
       \name{\bf Min~Jiang} \hfill \addr{minjiang@xmu.edu.cn}\\
        \addr{The Department of Artificial Intelligence, Key Laboratory of Digital Protection and Intelligent Processing of Intangible Cultural Heritage of Fujian and Taiwan, Ministry of Culture and Tourism, School of Informatics, Xiamen University, Fujian, China, 361005.}
\AND
       \name{\bf Kay Chen Tan} \hfill \addr{kctan@polyu.edu.hk}\\
        \addr{The Department of Data Science and Artificial Intelligence, The Hong Kong Polytechnic University, Hong Kong SAR, P.R. China}
}

\maketitle

\begin{abstract}

Multi-objective optimization problems (MOPs) require the simultaneous optimization of conflicting objectives. Real-world MOPs often exhibit complex characteristics, including high-dimensional decision spaces, many objectives, or computationally expensive evaluations. While population-based evolutionary computation has shown promise in addressing diverse MOPs through problem-specific adaptations, existing approaches frequently lack generalizability across distinct problem classes. Inspired by pre-training paradigms in machine learning, we propose a Population Pre-trained Model (PPM) that leverages historical optimization knowledge to solve complex MOPs within a unified framework efficiently. PPM models evolutionary patterns via population modeling, addressing two key challenges: (1) handling diverse decision spaces across problems and (2) capturing the interdependency between objective and decision spaces during evolution. To this end, we develop a population transformer architecture that embeds decision spaces of varying scales into a common latent space, enabling knowledge transfer across diverse problems. Furthermore, our architecture integrates objective-space features through objective fusion to enhance population prediction accuracy for complex MOPs. Our approach achieves robust generalization to downstream optimization tasks with up to 5,000 dimensions--five times the training scale and 200 times greater than prior work. Extensive evaluations on standardized benchmarks and out-of-training real-world applications demonstrate the consistent superiority of our method over state-of-the-art algorithms tailored to specific problem classes, improving the performance and generalization of evolutionary computation in solving MOPs.

\end{abstract}

\begin{keywords}
Evolutionary computation,
multi-objective optimization,
pre-trained model.
\end{keywords}

\section{Introduction}
Many optimization problems involve multiple conflicting objectives requiring simultaneous optimization~\citep{996017}. These multi-objective optimization problems (MOPs)~\citep{1597059} are prevalent across diverse domains, including machine learning~\citep{10.1145/3321707.3321735}, bioinformatics~\citep{10.1093/bioinformatics/btv582}, and scheduling~\citep{RN330}. As the field expands and applications broaden, real-world MOPs exhibit growing complexity~\citep{doi:10.1142/9789813143180_0004}, characterized by large-scale decision variables~\citep{gu2024large, RN208}, numerous objectives~\citep{wu2025solving, 10.1007/978-3-540-70928-2_57}, expensive evaluations~\citep{10433214}, and intricate constraints~\citep{neumann2024optimizing, 10477538}.

To address various difficulties of complex MOPs, researchers designed different population-based multi-objective evolutionary algorithms (MOEAs) to tackle distinct complexities~\citep{Qian_Yu_2017, RN89}. However, this ``one algorithm for one problem'' paradigm often necessitates prior knowledge of problem characteristics and intricate algorithm design. To enhance MOEA generalization, recent work, inspired by pre-training successes in natural language processing~\citep{NEURIPS2020_1457c0d6, NEURIPS2022_b1efde53} and computer vision~\citep{dosovitskiy2021an, 9716741}, has explored pre-training for general-purpose optimization~\citep{seiler2025deep, 2023pretrained}. Nevertheless, existing approaches predominantly focus on single-objective problems~\citep{2023pretrained} or small-scale decision spaces~\citep{seiler2025deep}, exhibiting poor scalability in both decision and objective spaces and consequently limited generalization to complex MOPs.

\begin{figure}[th]
\centering
\includegraphics[width=1\textwidth]{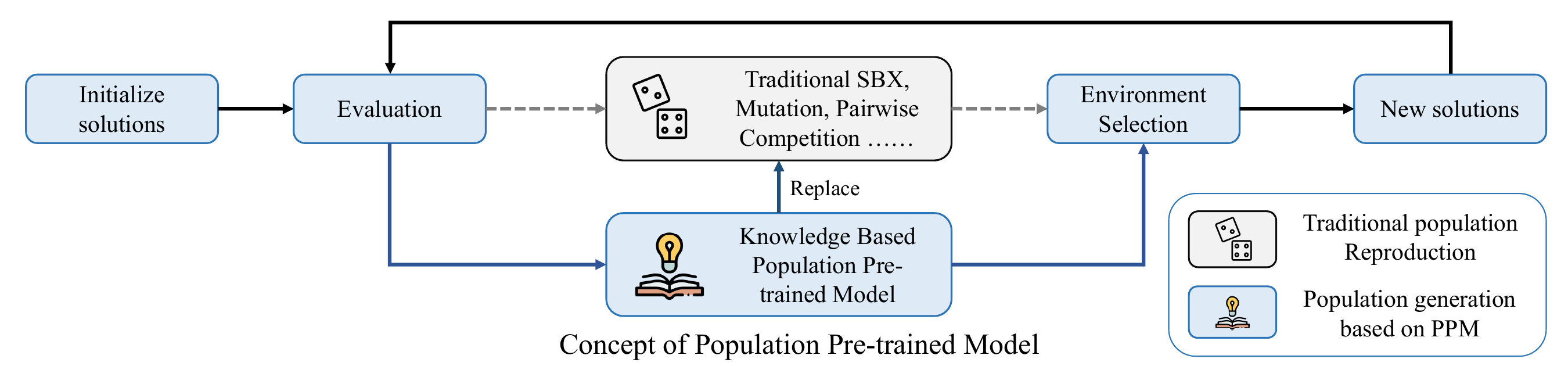}
\caption{\textbf{Illustration of the proposed population pre-trained model.} Utilize PPM to solve complex MOPs (Initialization - Evaluation - Reproducing - Selection - Fine-tuning). PPM could be regarded as the replacement of the process of generating a new population, which generally contains SBX, mutation, particle update, etc.}
\label{fig: idea}
\vspace{-0.5em}
\end{figure}

Taking this cue, we introduce a novel Population Pre-trained Model (PPM) for complex multi-objective optimization. \textit{Conceptually}, PPM is pre-trained on solution pairs derived from existing methods across diverse MOPs to generate promising solutions. This model provides a learning-based alternative to traditional, stochastic reproduction operators like simulated binary crossover (SBX)~\citep{SBX}, mutation~\citep{DE}, and swarm particle update~\citep{8681243}, leading to a new solution to improve the generalization of MOEAs. We provide a conceptual illustration for our idea in Figure~\ref{fig: idea}.

\textit{Technically}, we propose the population transformer based on the classical transformer architecture~\citep{NIPS2017_attention} to learn evolutionary knowledge from prior problems. Leveraging this, we introduce dimension embedding and objective fusion mechanisms to enhance the ability of our transformer to represent solutions with improved scalability and generalization across decision and objective spaces. First, dimension embedding maps problems with varying decision-space dimensionalities into a common latent space, enabling learning across scales and handling MOPs with up to 5,000 variables -- five times the scale of the training problem and nearly 200 times greater than prior work~\cite{2023pretrained}. Second, objective fusion integrates objective features with decision features to capture interdependencies between decision variables and objective values during evolution. This allows the model to learn evolutionary knowledge of decision variables under multi-objective convergence and diversity pressure, predicting subsequent populations with improved performance. Our main contributions are:

\textit{1)} This paper introduces a unified framework, the Population Pre-trained Model (PPM), for solving complex MOPs. Our model learns population-level patterns during evolution across diverse optimization problems, representing, to our knowledge, the first general-purpose solver for complex MOPs.

\textit{2)} We demonstrate the efficacy of the PPM by introducing a specific population transformer architecture. To tackle the challenges of existing approaches in modeling diverse and high-dimensional decision spaces and capturing evolutionary information from objective values, this architecture incorporates two novel mechanisms: a dimension embedding mechanism to handle variable scales of decision variables, and an objective fusion mechanism to integrate objective information into the model training. These mechanisms ensure robust performance and generalization across complex MOPs of diverse scales.

\textit{3)} Our population pre-trained model serves as a plugin module that can be integrated seamlessly into existing MOEAs, enhancing their ability to solve complex MOPs. Experimental results demonstrate state-of-the-art performance across diverse problem classes, including large-scale MOPs (up to 5,000 decision variables)~\citep{10.1145/3470971,9552479}, many-objective problems (up to 10 objectives)~\citep{10.1145/2792984}, computationally expensive problems (evaluated under a 1,000-function-evaluation budget)~\citep{Chugh2019emop}, and real-world constrained MOPs~\citep{8962275}.

The paper is structured as follows. Section \ref{sec: pre} establishes multi-objective optimization notation and complex MOP classifications. Section \ref{sec: et} details the PPM framework and its implementation. Section \ref{sec: exp} presents experimental validation and analysis, while Section \ref{sec: con} concludes with research contributions and future directions.
\par

\section{Preliminaries}
\label{sec: pre}
This section provides a concise review of complex MOPs, their inherent challenges, and the corresponding MOEAs. Subsequently, we discuss recent advances of pre-trained models for optimization problems.
\subsection{Complex MOPs and Corresponding MOEAs}
\textbf{Multi-objective optimization problems} (MOPs) can be mathematically formulated as follows:
\begin{equation}
\begin{aligned}
\label{alg: MOP}
&\textrm{minimize}\ \boldsymbol f(\boldsymbol x) := (f_1(\boldsymbol x), f_2(\boldsymbol x),\dots,f_m(\boldsymbol x))\\
&\textrm{subject to}\ \ \ \ \ \ \  \boldsymbol{x}  \in \Omega
\end{aligned}
\end{equation}
where $\boldsymbol x=(x_1,x_2,...,x_d)$ is $d$-dimensional decision vector, $\boldsymbol f=(f_1,f_2,\dots,f_m)$ is $m$-dimensional objective vector. 
Suppose $\boldsymbol x_1$ and $\boldsymbol x_2$ are two solutions of an MOP, solution $\boldsymbol x_1$ is known to Pareto dominate solution $\boldsymbol x_2$ (denoted as $\boldsymbol x_1 \prec \boldsymbol x_2$), if and only if $f_i(\boldsymbol x_1) \leqslant f_i(\boldsymbol x_2) (\forall i = 1,\dots,m)$ and there exists at least one objective $f_j (j \in \{1, 2, \dots , m\})$ satisfying $f_j(\boldsymbol x_1) < f_j(\boldsymbol x_2)$. The collection of all the Pareto optimal solutions in the decision space is called the Pareto optimal set (PS), and the projection of PS in the objective space is called the Pareto optimal front (PF).
\par
\subsubsection{Complex MOPs}
As research on MOPs advances, increasingly complex MOPs with different characteristics are being explored. These complex MOPs are frequently encountered in real-world scenarios. A notable example is the Transformation Ratio Error Estimation (TREE) problem~\citep{8962275}, which necessitates the simultaneous consideration of computationally expensive evaluation functions, constraints, and large-scale decision variables. This subsection introduces properties including large-scale decision variables, many objectives, expensive evaluation functions, and constraints, along with existing MOEAs designed to address these challenges.
\par
\paragraph{Large-scale Decision Variables.}
An MOP is classified as large-scale (LSMOP) when the number of decision variables $d \geq 100$~\citep{10.1145/3470971,9552479}. LSMOPs face the \textit{curse of dimensionality}~\citep{10.1145/3470971,10107418}, where computational difficulty grows exponentially with $d$, causing slow convergence in conventional algorithms. Recent MOEAs for LSMOPs fall into three categories: decision variable grouping-based~\citep{6557903}, decision space reduction-based~\citep{RN89}, and novel search strategy-based approaches~\citep{9138459,RN257}. 
\par
\paragraph{Many Objectives.}
An MOP becomes a many-objective optimization problem (MaOP) when objectives exceed three ($m > 3$)~\citep{10.1145/2792984}. Increasing $m$ challenges MOEAs through two effects: solution incomparability due to a rapidly rising proportion of non-dominated solutions~\citep{10.1007/978-3-540-70928-2_57}, and exponential growth in solutions needed to approximate high-dimensional Pareto fronts~\citep{6782742}. Many-objective evolutionary algorithms (MaOEAs) are categorized into seven classes based on their techniques~\citep{10.1145/2792984}, with detailed reviews available therein.
\par
\paragraph{Expensive Evaluation Functions.}
MOPs with objectives requiring extensive computation time per evaluation are termed computationally expensive MOPs (EMOPs)~\citep{Chugh2019emop,POLONI2000403}. Surrogate models (e.g., Kriging, artificial neural networks, polynomial regression) are commonly integrated into MOEAs to approximate these functions~\citep{8281523}, replacing costly exact evaluations.
\par
\paragraph{Constrained Optimization.}
Constrained MOPs (CMOPs) incorporate constraints into Eq.~(\ref{alg: MOP}) as $g_j(\boldsymbol{x}) \leq 0$ ($j=1,\dots,l$) for inequalities and $h_j(\boldsymbol{x}) = 0$ ($j=l+1,\dots,k$) for equalities, where $l$ and $k-l$ denote inequality/equality constraint counts~\citep{9723472}. Significant research focuses on constraint handling techniques (CHTs), driving the development of specialized constrained MOEAs (CMOEAs)~\citep{9723472}.
\par
\paragraph{Complex MOPs with Multiple Complexities.} 
Beyond single complexities, real-world optimization problems generally show multiple complexities~\citep{8962275} and recent MOEAs target combined challenges like large-scale MaOPs~\citep{9762228}, constrained large-scale MOPs~\citep{9311862,9861720}, and expensive MaOPs~\citep{8640100}. However, despite these advancements, the resolution of complex MOPs with more complexities, which encompass two or more of the complexities, continues to pose a significant challenge. For example, the exploration of expensive large-scale MOPs is largely uncharted due to the inefficiency of surrogate models such as Kriging~\citep{1665031} in learning the mapping between a multitude of decision variables and multiple objectives~\citep{10.1145/3470971}. This gap motivates our universal framework for MOPs with diverse complexities.

\subsection{Pre-trained Models in Optimization}
\paragraph{Pre-trained Models.} Pre-trained models encode visual~\citep{He_2016_CVPR, dosovitskiy2021an} or natural language~\citep{NEURIPS2020_1457c0d6} knowledge acquired from massive datasets. These models capture general patterns and semantic representations, serving as valuable resources for downstream tasks. For instance, pre-trained language models can solve diverse tasks previously requiring specialized models, including language translation, sentiment analysis, and question answering. This success motivates the development of pre-trained models for solving optimization problems.

\paragraph{Pre-training for Optimization.} Recent work introduced the Pre-trained Optimization Model (POM) for zero-shot single-objective optimization~\citep{2023pretrained}. POM concentrates on addressing the zero-shot optimization by leveraging knowledge gained from optimizing diverse tasks. Additionally, Deep-ELA~\citep{seiler2025deep} targets single- and multi-objective continuous optimization. However, current methodologies are constrained by limited decision spaces and fail to incorporate objective space information. To address these limitations, we propose a novel framework that generalizes to higher-dimensional search spaces and integrates objective values to guide population generation. This approach enhances both convergence and diversity, resulting in a more general solver for complex multi-objective problems.

\par
\section{Population Pre-trained Model}
\label{sec: et}
The pre-training of our proposed Population Pre-trained Model relies on three foundational components: the dataset, the training objective, and the model architecture. Specifically, (1) the dataset comprises a curated collection of populations generated by diverse MOEAs across a broad spectrum of MOPs; (2) the training objective is to predict the subsequent generation of a population given its predecessor; and (3) the model architecture is a bespoke Population Transformer designed for this purpose. 

This section first introduces the preparation of the pre-training dataset and the training objective, then outlines the design rationale and architecture details for the PPM, and finally details the implementation of integrating the PPM as a plugin into existing MOEAs.

\subsection{Pre-training dataset and objective}
To equip the Population Transformer Model with the capability to address Hull address complex MOPs, we pre-train it on a carefully curated dataset of populations generated by various MOEAs across diverse MOPs.

\begin{figure}[t]
    \centering
    \includegraphics[width=1\textwidth]{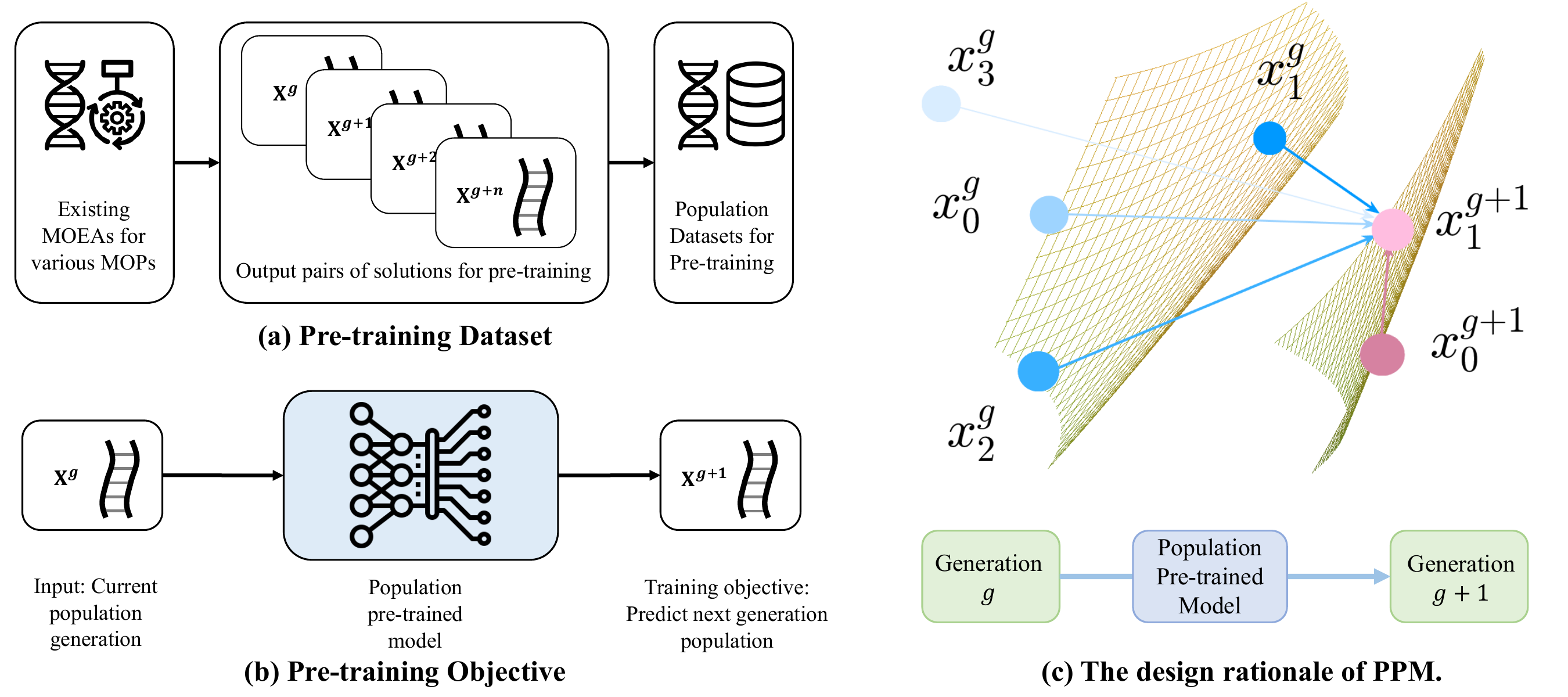}
    \caption{\textbf{Illustration of the proposed population pre-trained model: dataset, pre-training, and design rationale.} (a) The pre-training dataset is constructed from solution pairs generated by existing MOEAs across diverse MOPs. (b) The training objective of the PPM is to predict a next-generation population exhibiting better convergence and diversity, using the current population as input. (c) The figure illustrates the self-attention mechanism inherent in PPM. It stimulates the attention outcomes during the generation of the solution $\boldsymbol{x}^{g+1}_{1}$. The color of the circle and line corresponds to the magnitude of the attention score, with darker shades indicating larger scores.}
    \label{fig: PPM}
    \vspace{-0.5em}
\end{figure}

\paragraph{Dataset.} As illustrated in Figure~\ref{fig: PPM}~(a), the dataset consists of paired populations from consecutive generations, thereby capturing the evolutionary dynamics inherent in diverse MOPs. These pairs are derived from outputs of established MOEAs applied to various MOPs, providing a rich collection of solution sets that reflect real-world optimization scenarios. The inclusion of multiple algorithms and problems in the dataset promotes robustness, enabling the model to generalize across different optimization problems and objective landscapes.

\paragraph{Training Objective.} As depicted in Figure~\ref{fig: PPM}~(b), the pre-training objective of the PPM is to predict the population of the subsequent generation based on the current one. This predictive task mimics the iterative nature of evolutionary algorithms, training the model to anticipate improvements in solution quality over generations. Detailed implementation aspects of the dataset curation and pre-training process are elaborated in the experimental section (Section~\ref{sec: exp: pre}).


\subsection{Design Rationale of Population Transformer}

\textbf{Seq2Seq and Pop2Pop Adaptation.} Transformer architectures inherently process sequential input to generate sequential output. Within heuristic MOEAs, populations represent sequences of solutions, and offspring generation relies on specialized operators. Consequently, adapting attention-based models to function as population-to-population (pop2pop) mappings is a natural progression.

\paragraph{Attention Mechanism.} Transformers leverage self-attention within inputs and cross-attention between inputs and outputs. Analogously, the proposed PPM computes self-attention among solutions within a single generation and cross-attention between consecutive generations. By integrating objective-space information into solution representations and modeling attention across solutions, the quality of generated offspring is enhanced. Figure~\ref{fig: PPM} (c) provides a simulated illustration of this attention mechanism within the PPM.
\begin{figure}[th]
    \centering
    \includegraphics[width=1\textwidth]{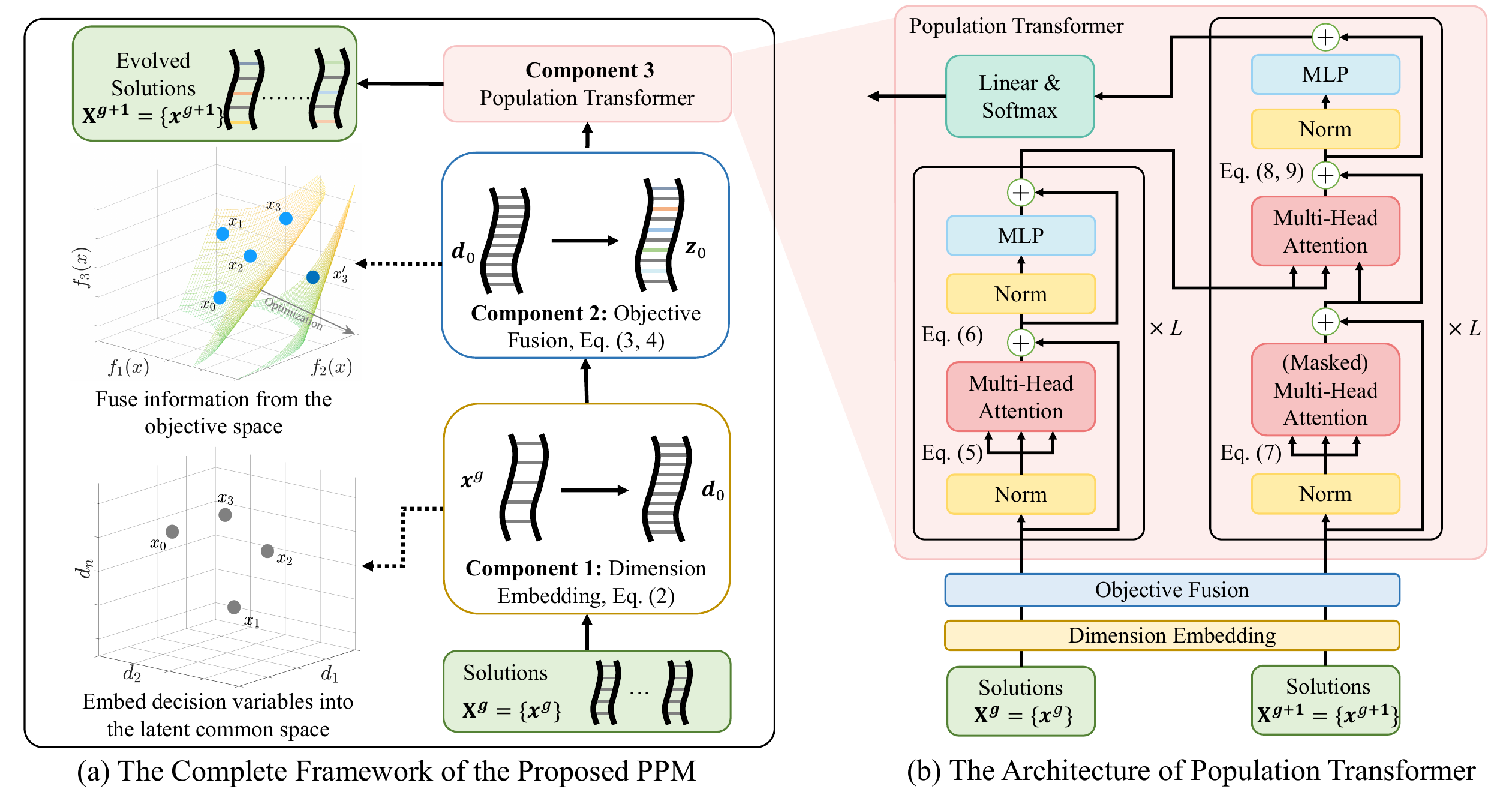}
    \caption{\small (a) The framework of the proposed Population Transformer, an instantiation of PPM, operates as follows: 1. Collect the population $\mathbf{X}^{g} = {\boldsymbol{x}^{g}}$ at generation $g$. 2. Embed the decision variable dimensions of $\boldsymbol{x}^g$ using Eq.~(\ref{equ: dim-embed}) to obtain $\boldsymbol{d}_0$. 3. Fuse objective values into $\boldsymbol{d}_0$ via Eq.~(\ref{equ: obj-embed}) and~(\ref{equ: input}) to obtain $\mathbf{Z}_{0}$. 4. Process $\mathbf{Z}_{0}$ using the PPM. 5. Output the next-generation population $\mathbf{X}^{g+1}={ \boldsymbol{x}^{g+1} }$. (b) The architecture of the population transformer. During pre-training, $\mathbf{X}^{g+1}=\{ \boldsymbol{x}^{g+1} \}$ are target solutions but masked for training. During fine-tuning, $\mathbf{X}^{g+1}=\{ \boldsymbol{x}^{g+1} \}$ are generated solutions by PPM, and once one solution $\boldsymbol{x}^{g+1} $ is generated, $\boldsymbol{x}^{g+1} $ will be evaluated and then input to the PPM.}
    \label{fig: framework}
    \vspace{-1em}
\end{figure}
\subsection{Architecture Details}

To develop a Population Transformer capable of processing and generating promising solutions, we employ the classical Transformer architecture~\citep{NIPS2017_attention, NEURIPS2020_1457c0d6, dit-2023}. 

As illustrated in Figure~\ref{fig: framework} (a), the complete population pre-trained model contains dimension embedding (component 1), objective fusion (component 2), and population transformer architecture (component 3). These components enable the transformer to process populations with varying dimensionalities and generate subsequent population states, guided by convergence and diversity information derived from objective spaces.

\subsubsection{Dimension Embedding}
The dimensionality of solutions varies across different MOPs. This variation complicates learning population representations from disparate decision spaces and limits the generalization capability of existing methods to high-dimensional problems. To address this, we propose dimension embedding. This technique projects solutions from diverse dimensional spaces into a higher-dimensional, common latent space, enabling the population transformer to handle MOPs with varying decision variable scales and generalize effectively to high-dimensional MOPs.

The standard transformer requires fixed-size token embeddings. To process variable-dimensional solutions, we first pad each solution $\mathbf{X}^{g} = \{\boldsymbol{x}_{0}^{g}, \dots, \boldsymbol{x}_{N}^{g}\}$ (where $\boldsymbol{x}_{i}^{g} \in \mathbb{R}^{d}$) in generation $g$ with zeros to a uniform dimension $\hat{d}$. Subsequently, these padded solutions, originating from diverse problems and decision spaces, are projected into a common latent space via a trainable linear transformation following the standard token embedding method. This linear transformation is implemented using a multi-layer perceptron, mapping $\mathbf{X}^{g}$ to a set of latent vectors $\mathbf{D}^{g}_{0} =\{ \boldsymbol{d}^{g}_{0},\dots,\boldsymbol{d}^{g}_{N}\}$ ($\boldsymbol{d}^{g}_{i} \in \mathbb{R}^{d'}$). Here, $d'$ denotes the predefined input size of the PPM, corresponding to the maximum dimension it can process. We term $\mathbf{D}^{g}_{0}$ the ``solution embeddings''. 

The PPM processes the entire population of $n$ solutions concurrently within a generation. This population size $n$ also defines the effective input sequence length for the transformer. The dimension embedding ensures the generalization across different decision spaces within $d'$ and is denoted by:

\begin{equation}
    \label{equ: dim-embed}
    \mathbf{D}^{g}_{0} = [\boldsymbol{x}^{g}_{0}\textbf{E}_{\textrm{dim}}; \boldsymbol{x}^{g}_{1}\textbf{E}_{\textrm{dim}}; \cdots \boldsymbol{x}^{g}_{N}\textbf{E}_{\textrm{dim}}], \textbf{E}_{\textrm{dim}} \in \mathbb{R}^{d\times d'},
\end{equation}
where $\textbf{E}_{\textrm{dim}}$ is the learnable linear projection.
\subsubsection{Objective Fusion}
The evaluation of solutions plays a crucial role in the evolving process of MOEAs. To enhance the learning capabilities of the proposed model in capturing solution evolution to output promising solutions, it is essential to incorporate the evaluation into the solution. Moreover, it is important to note that the proximity of two solutions in the decision space does not necessarily imply similarity in the objective space. Therefore, integrating evaluation information into the solution representation is important to capture relationships between objectives and improve the quality of evolved solutions.

In this regard, objective fusion is introduced to integrate evaluation information into the solution representation. By embedding the positions of solutions within the objective space, the model can better capture relationships between objectives and improve the quality of evolved solutions. With this approach, we combine the position of the solution in the objective space into solution embeddings $\mathbf{D}^{g}_{0}$, enabling the integration of evaluation information into the PPM.
\par
We use standard learnable position embeddings to obtain objective embeddings $\mathbf{O}^{g}_{0}$ in Eq.~(\ref{equ: obj-embed}). Afterward, objective embeddings are added to the solution embeddings to retain positional information in the objective space, and the resulting sequence of embedding vectors $\mathbf{Z}^{g}_{0}$ (Eq.~(\ref{equ: input})) serves as input to the encoder of the population transformer.
\begin{align}
    \label{equ: obj-embed}
    \mathbf{O}^{g}_{0} & = [\boldsymbol{f}^{g}_{0}\textbf{E}_{\textrm{obj}}; \boldsymbol{f}^{g}_{1}\textbf{E}_{\textrm{obj}}; \cdots \boldsymbol{f}^{g}_{N}\textbf{E}_{\textrm{obj}}], \textbf{E}_{\textrm{obj}} \in \mathbb{R}^{m\times d'}, \\
    \label{equ: input}
    \mathbf{Z}^{g}_{0} & = \mathbf{D}^{g}_{0} + \mathbf{O}^{g}_{0},
\end{align}
where $\boldsymbol{f}^{g}_{i}$ is $i$-th objective value of solution $g$ and $\mathbf{E}_{\textrm{obj}}$ is the learnable linear projection.
\subsubsection{Population Transformer}
%
As the PPM is designed to generate next-generation promising solutions, a Transformer with an encoder-decoder architecture is employed~\citep{NIPS2017_attention}. Both the encoder and decoder comprise a stack of $N$ identical layers (where $N = L$). As illustrated in Figure~\ref{fig: framework} (b), the model follows an auto-regressive approach~\citep{graves2013generating}, utilizing stacked self-attention and fully connected layers in both components.

\paragraph{Encoder.} The encoder receives $\mathbf{Z}^{g}_{0}$ as the input and maps to a sequence of continuous representations $\mathbf{Z}^{g}_{L}$. The encoder consists of layers of Multi-Headed Attention (MHA) and Multilayer Perceptron (MLP) blocks, and layer norm (LN) is applied before every block and residual connections after every block (Eq.~(\ref{equ: encode1},~\ref{equ: encode2})). 
\begin{align}
    \label{equ: encode1}
    \mathbf{Z'}^{g}_{l+1} & = \textrm{MHA} ( \textrm{LN} ( \mathbf{Z}^{g}_{l} )) + \mathbf{Z}^{g}_{l} \\
    \label{equ: encode2}
    \mathbf{Z}^{g}_{l+1} & = \textrm{MLP} ( \textrm{LN} ( \mathbf{Z'}^{g}_{l+1} )) + \mathbf{Z'}^{g}_{l+1} \\
    l & = 0,\dots,L-1 \nonumber.    
\end{align}

\paragraph{Decoder.} After fusing $\mathbf{X}^{g}$ into $\mathbf{Z}^{g}_{L}$, the decoder then generates one next generation ($g+1$) of solution $\boldsymbol{x}^{g+1}$ at a time. At each step, the model consumes the previously generated solutions as additional input when generating the next.
\par
Given the newly generated solutions $\boldsymbol{x}^{g+1}_{n}$ and accumulated solutions $\mathbf{X}^{g+1}$, to generate the next $\boldsymbol{x}^{g+1}_{n+1}$, the PPM firstly evaluates the $\boldsymbol{x}^{g+1}_{n}$ and obtain its objective $\boldsymbol{f}$. Second, the decoder receives $\mathbf{X}^{g+1}$ (including $\boldsymbol{x}^{g+1}_{n}$) as the input for dimension embedding and objective fusion and obtains $\mathbf{Z}^{g+1}_{0}$. Third, the decoder processes $\mathbf{Z}^{g+1}_{0}$ through LN and MHA via Eq. (\ref{equ: decode1}) to obtain $\mathbf{Z'}^{g+1}_{L}$.
\par
After obtaining $\mathbf{Z}^{g}_{l+1}$ and $\mathbf{Z'}^{g+1}_{l+1}$, the decoder employs the attention mechanism, which allows it to focus on relevant parts of the input solutions by computing a weighted sum of values based on the similarity between query and key vectors. Specifically, the attention mechanism calculates the cross-attention $\mathbf{C}^{g+1}_{l+1}$ by using the output of the previous decoder block ($\mathbf{Z'}^{g+1}_{l+1}$) as the query $\mathbf{Q}$, and the encoder's output ($\mathbf{Z}^{g}_{l+1}$) as the key $\mathbf{K}$ and value $\mathbf{V}$ (Eq. (\ref{equ: decode2})). This procedure is essentially an aggregation of the information from the whole solutions at generation $g$ and generated solutions at generation $g+1$.
\par
Finally, the decoder applies residual connections to the cross attention $\mathbf{C}^{g+1}_{l+1} $ and $ \mathbf{Z'}^{g+1}_{l+1}$ and input the result to the MLP after layer norm (Eq. (\ref{equ: decode3})). The masked MHA block of the decoder ensures that the generation for solution $\boldsymbol{x}^{g+1}_{n}$ can depend only on the known solutions before $n$.
\begin{align}
    \label{equ: decode1}
    \mathbf{Z'}^{g+1}_{l+1} & = \textrm{MHA} ( \textrm{LN} ( \mathbf{Z}^{g+1}_{l} )) + \mathbf{Z}^{g+1}_{l} \\
    \label{equ: decode2}
    \mathbf{C}^{g+1}_{l+1} & = \textrm{MHA} ( \mathbf{Z}^{g}_{l+1} (\mathbf{K}, \mathbf{V}),  \mathbf{Z'}^{g+1}_{l+1} (\mathbf{Q})) \\
    \label{equ: decode3}
    \mathbf{Z}^{g+1}_{l+1} & = \textrm{MLP}( \textrm{LN}( \mathbf{C}^{g+1}_{l+1} + \mathbf{Z'}^{g+1}_{l+1} )) + \mathbf{C}^{g+1}_{l+1}\\
    l & = 0,\dots,L-1 \nonumber
\end{align}
\par
The final output ($\mathbf{Z}^{g+1}_{L}$ ) of the decoder goes through the linear layer and the softmax layer, and finally, the $(n+1)$th solution of the $(g+1)$th generation is obtained.

\subsection{Method Modularity and Its Implementation}
Once the pre-trained PPM has been acquired, it can be integrated into any MOEAs to solve other complex MOPs. The PPM serves as a replacement for the conventional method of generating the next generation of solutions. Algorithm~\ref{alg: nsgaii-et} provides an example algorithm that combines PPM with the NSGA-II~\citep{996017}. Specifically, the "fast-non-dominated-sort" and "crowding-distance-selection" steps in this algorithm refer to the environmental selection process of NSGA-II. Instead of utilizing crossover and mutation operators, the creation of new populations is accomplished by $\mathbf{X}^{g+1}, \mathbf{F}^{g+1} \gets $ PPM($\mathbf{X}^{g}, \mathbf{F}$), leveraging the proposed pre-trained PPM.
\par
In addition to NSGA-II, the proposed PPM can collaborate with any available MOEA to harness pre-trained knowledge and facilitate the generation of improved populations, resulting in faster convergence to the Pareto frontier. At the $g$th iteration of any MOEA, PPM receives the solutions $\mathbf{X}^{g}$ and their corresponding evaluations, which are then passed as inputs to the encoder of PPM. During the generation process, the decoder of PPM initializes the $(g+1)$-th generation by accepting a randomly selected solution as the initialization token. Subsequently, PPM generates a solution for the $(g+1)$-th generation, evaluates it, and feeds it back into the decoder for further refinement. Ultimately, PPM delivers the complete population of the $(g+1)$-th generation, denoted as $\mathbf{X}_{g+1}$, to the cooperative MOEA.

The core idea of the proposed method is to learn evolutionary patterns between generations from existing algorithms, thereby enabling the model to generate promising solutions for new complex MOPs. The effectiveness of the proposed model is verified through extensive experiments, as discussed in the subsequent section.
\begin{algorithm}[t]\small
    \caption{NSGA-II with PPM}
    \label{alg: nsgaii-et}
        \textbf{Input}: $\mathbf{P}$: PPM.\\
        \textbf{Parameter}: $N$: Population size, $E$: Total function evaluations.\\
        \textbf{Output}: $\mathbf{X}^{G}$: Output solutions.
        \begin{algorithmic}[1] 
        \STATE Initialize a set of $N$ solutions $\mathbf{X}^{0}$;
        \STATE $\mathbf{F}^{0} \gets $ Evaluate ($\mathbf{X}^{0}$);
        \STATE $E \gets E - N$;
        \WHILE{$E > 0$}
            \STATE $\mathcal{F} \gets $ fast-non-dominated-sort($\mathbf{X}^{g}, \mathbf{F}^{g}$);
            \STATE $\mathbf{X}^{g} \gets $ crowding-distance-selection($\mathcal{F}$);
            \STATE $\mathbf{X}^{g+1}, \mathbf{F}^{g+1} \gets $ PPM($\mathbf{X}^{g}, \mathbf{F}$);
            \STATE PPM $\gets$ fine-tune(PPM, $\mathbf{X}^{g}, \mathbf{F}, \mathbf{X}^{g+1}, \mathbf{F}^{g+1}$);
            \STATE $E \gets E - N$;
            \STATE $g \gets g + 1$ ;
        \ENDWHILE
        \STATE \textbf{return} $\mathbf{X}^{G}$
    \end{algorithmic}
\end{algorithm}


\section{Experiments}
\label{sec: exp}

In this section, we first outline the experimental settings used for comparison. We then evaluate the performance of the proposed PPM in addressing complex MOPs, which include large-scale decision variables, many objectives, expensive evaluation functions, and constraints. The experiments are conducted on a variety of benchmarks, including real-world problems. Finally, we present an ablation study and analyze the convergence and computational efficiency of the proposed method.
\subsection{Experimental Setup}

\paragraph{Compared Algorithms.} NSGA-II~\citep{996017} is used as the baseline as it is used for our specific implementation. Besides, we choose SOTA methods that are specialized in solving large-scale, many objective, constrained, and expensive MOPs: CCGDE3~\citep{6557903} and WOF~\citep{RN89} are algorithms for solving LSMOPs, LMOCSO~\citep{8681243} can solve MOP efficiently~\citep{RN89}, DGEA~\citep{9138459} has advantages in solving LSMOPs and MaMOPs~\citep{9138459}, and CMOEAD~\citep{6600851} is designed for MOPs with constraints. CMOCSO~\citep{9861720} are designed for solving large-scale constrained MOPs. POCEA~\citep{9311862} shows promising performance in solving large-scale constrained MaOPs. Besides, we include two machine learning-based MOEAs, EmoDM~\citep{yan2024emodm} and MODE/D-LO~\citep{liu2023large}, for comparison. They utilize the diffusion model and a large language model to search for better solutions, respectively.

\paragraph{Parameter Setting.} General parameters and algorithm parameters are presented as follows: 
\begin{enumerate}
    \item {Population Size}. To ensure consistency across all test benchmarks, the population size is set to 100~\citep{8681243, 9138459}.
    \item {Termination Condition}. To test the performance of algorithms on expensive optimization problems, the number of maximum evaluations $E$ is set to 1,000 for all compared MOEAs~\citep{Chugh2019emop}.
    \item Algorithm Parameters. Parameters for the PPM are listed in Table~\ref{tab: para}.  All algorithms are implemented in PlatEMO \citep{RN92}, and the parameter settings for all compared algorithms follow their original publications to ensure fair comparison. The parameter settings for each algorithm are detailed in Table~\ref{tab: full-para} in Appendix~\ref{app:ps}.
\end{enumerate}


\begin{table}[tbp]
  \centering
  \caption{Parameters for The Proposed PPM.}
  \begin{adjustbox}{width=0.9\hsize,center}
    \begin{tabular}{ccccc}
    \toprule
    Dim embed size d' & Batch size & Layers & Hidden size D & Heads \\
    10,000 & 64    & 11    & 512   & 10 \\
    \midrule
    Optimizer & beta1 & beta2 & Weight decay & \# Parameters \\
    Adam~\citep{DBLP:journals/corr/KingmaB14}  & 0.9   & 0.999 & 0.1   & 61M \\
    \bottomrule
    \end{tabular}%
  \label{tab: para}%
    \end{adjustbox}
\end{table}%

\paragraph{Benchmarks.} The experiments utilize three widely adopted benchmark suites: ZDT~\citep{6787994}, LSMOP~\citep{7553457}, and TREE~\citep{8962275}. The ZDT suite is commonly used to test the scalability of algorithms with respect to different Pareto front shapes and solution complexities. The test problem suite LSMOP~\citep{7553457} is commonly employed for evaluating the performance of MOEAs in the presence of decision variables and objectives that scale differently. Lastly, the TREE problem represents a real-world constrained MOP, allowing us to assess how well the algorithms handle practical optimization challenges with strict constraints and computationally expensive objective functions. Specifically, TREE~\citep{8962275} is an important real-world task in modern power delivery systems, which aims to detect the voltage transformers' ratio error (RE). Based on the statistical and physical rules, the TREE is modeled as several computationally expensive, constrained, large-scale MOPs. Objectives include total time-varying, the sum of the RE variation, and the phase angle relationship among the true voltage values. Constraints contain topology, series, and phase constraints.

\subsection{Metrics}
We employ the Inverted Generational Distance (IGD)~\citep{RN98} and Hypervolume (HV)~\citep{1583625} metrics to rigorously assess the convergence and diversity of algorithms addressing complex MOPs. The IGD metric quantifies the convergence quality of solutions by measuring the average distance from a reference set to the obtained solutions, with lower values indicating superior convergence performance. The HV metric evaluates the diversity and dominance of a solution set by calculating the volume of the objective space collectively dominated by those solutions, where a larger HV signifies better overall performance.
\par
Each algorithm is run 20 times independently, and the Wilcoxon rank-sum~\citep{Haynes2013} is used to compare the statistical results obtained at a significance level of 0.05. In the tables, (+/-/=) indicate that compared algorithms perform significantly better, significantly worse, or indifferent compared to the PPM, respectively, in a statistically meaningful sense.

\subsection{Pre-Training}
\label{sec: exp: pre}
To evaluate the performance of the pre-training stage, we conducted experiments on the ZDT and LSMOP benchmark problems using NSGA-II, DGEA, and WOF. These algorithms were chosen because of their proven ability to explore and exploit solution spaces effectively. The goal of pre-training is to train the proposed PPM on a diverse set of MOPs to generate high-quality populations. Specifically, we executed NSGA-II, DGEA, and WOF on ZDT 1-5 and LSMOP 1-6, performing $1,000 \times d$ function evaluations for each. We varied the number of decision variables $d$, setting them as 100, 200, 500, and 1,000, while the number of objectives was set to 2 and 3.
\par
Throughout the runs, we recorded the populations at each generation. Considering generation $g$ and generation $g+1$ as a pair of populations for pre-training, PPM was utilized to generate a new population $g'$ based on the input of population $g$, followed by the calculation of the loss between populations $g'$ and $g+1$. The pre-training process using PPM, performed on one A40 GPU with a Platinum 8358P CPU, required approximately two days for 1000 epochs.

\subsection{Performance on Benchmarks}
\begin{table*}[tbp]
\scriptsize
  \centering
  \caption{IGD Values Obtained by Compared Algorithms on 16 Instances From ZDT and LSMOP Test Suite. The Best Result in Each Row is Highlighted in Bold. The Last Column Represents the Rate of Change of the Proposed PPM Compared to Suboptimal Results.}
  \begin{adjustbox}{width=1\hsize,left}
    \begin{tabular}{cccccccccccccccc}
    \toprule
    Problem & D     & NSGA-II & CCGDE3 & DGEA  & WOF   & LMOCSO & CMOEAD & POCEA & CMOCSO & ABSAEA & CSEA  & EmoDM & MOEA/D-LO & PPM    & ROC(\%) \\
    \midrule
    \multirow{4}[2]{*}{\parbox{1cm}{\centering ZDT6\\ M=2}} & 250   & 7.48e+00- & 7.63e+00- & 7.62e+00- & 7.67e+00- & 7.50e+00- & 7.66e+00- & 7.69e+00- & 7.47e+00- & 7.69e+00- & 7.76e+00- & 7.72e+00- & 7.70e+00- & \textbf{6.55e-01} & 91.23 \\
          & 2500  & 7.79e+00- & 7.81e+00- & 7.84e+00- & 7.80e+00- & 7.85e+00- & 7.86e+00- & 7.84e+00- & 7.72e+00- & 7.83e+00- & 7.87e+00- & 6.15e+01- & 9.08e+04- & \textbf{6.55e-01} & 91.52 \\
          & 3000  & 7.78e+00- & 7.85e+00- & 7.84e+00- & 7.79e+00- & 7.85e+00- & 7.86e+00- & 7.85e+00- & 7.80e+00- & 7.83e+00- & 7.87e+00- & 6.06e+01- & 2.11e+01- & \textbf{6.55e-01} & 91.58 \\
          & 5000  & 7.83e+00- & 7.86e+00- & 7.85e+00- & 7.83e+00- & 7.83e+00- & 7.90e+00- & 7.87e+00- & 7.82e+00- & 4.55e+01- & 4.55e+01- & 4.55e+01- & 6.12e+01- & \textbf{6.55e-01} & 91.62 \\
    \midrule
    \multicolumn{1}{c}{\multirow{4}[2]{*}{\parbox{1cm}{\centering LSMOP7\\ M=2}}} & 250   & 4.41e+04- & 3.74e+04- & 2.12e+04- & 5.55e+04- & 3.72e+03- & 4.17e+04- & 3.31e+03- & 3.04e+04- & 6.47e+04- & 8.53e+04- & 7.72e+00- & 7.75e+00- & \textbf{1.50e+00} & 80.58 \\
          & 2500  & 8.24e+04- & 8.06e+04- & 1.17e+04- & 7.95e+04- & 1.01e+04- & 8.13e+04- & 3.90e+03- & 3.50e+04- & 8.39e+04- & 9.05e+04- & 9.00e+04- & 9.19e+04- & \textbf{1.52e+00} & 99.96 \\
          & 3000  & 8.00e+04- & 7.13e+04- & 2.61e+04- & 8.32e+04- & 6.73e+03- & 8.48e+04- & 4.70e+03- & 2.40e+04- & 8.50e+04- & 9.36e+04- & 6.06e+01- & 2.11e+01- & \textbf{1.52e+00} & 92.79 \\
          & 5000  & 8.62e+04- & 8.03e+04- & 9.30e+03- & 8.59e+04- & 9.51e+03- & 8.45e+04- & 5.47e+03- & 2.26e+04- & 4.55e+01- & 4.55e+01- & 4.55e+01- & 6.12e+01- & \textbf{1.52e+00} & 96.66 \\
    \midrule
    \multicolumn{1}{c}{\multirow{4}[2]{*}{\parbox{1cm}{\centering LSMOP8\\ M=2}}} & 250   & 1.32e+01- & 1.62e+01- & 6.96e+00- & 3.69e+00- & 5.03e+00- & 1.55e+01- & 3.36e+00- & 9.82e+00- & 1.77e+01- & 1.88e+01- & 7.72e+00- & 7.75e+00- & \textbf{7.42e-01} & 77.91 \\
          & 2500  & 1.92e+01- & 1.85e+01- & 7.78e+00- & 1.94e+01- & 5.99e+00- & 1.99e+01- & 3.86e+00- & 8.58e+00- & 1.96e+01- & 2.11e+01- & 9.08e+04- & 9.19e+04- & \textbf{7.42e-01} & 80.79 \\
          & 3000  & 1.89e+01- & 1.85e+01- & 8.99e+00- & 4.63e+00- & 7.39e+00- & 2.03e+01- & 4.18e+00- & 8.74e+00- & 2.00e+01- & 2.07e+01- & 2.09e+01- & 2.04e+01- & \textbf{7.42e-01} & 82.24 \\
          & 5000  & 2.03e+01- & 1.81e+01- & 1.33e+01- & 2.11e+01- & 6.78e+00- & 2.10e+01- & 4.20e+00- & 8.38e+00- & 4.55e+01- & 4.55e+01- & 4.55e+01- & 6.12e+01- & \textbf{7.42e-01} & 82.33 \\
    \midrule
    \multirow{4}[2]{*}{\parbox{1cm}{\centering LSMOP9\\ M=2}} & 250   & 3.67e+01- & 4.08e+01- & 4.83e+01- & 4.17e+01- & 2.15e+01- & 3.09e+01- & 6.73e+00- & 2.89e+01- & 3.81e+01- & 5.26e+01- & 7.72e+00- & 7.75e+00- & \textbf{8.10e-01} & 87.96 \\
          & 2500  & 5.70e+01- & 5.45e+01- & 4.21e+01- & 5.43e+01- & 3.15e+01- & 5.49e+01- & 1.73e+01- & 4.06e+01- & 5.70e+01- & 6.15e+01- & 9.08e+04- & 9.19e+04- & \textbf{8.10e-01} & 95.31 \\
          & 3000  & 5.54e+01- & 5.42e+01- & 4.22e+01- & 6.11e+01- & 1.69e+01- & 5.55e+01- & 7.12e+00- & 4.29e+01- & 5.73e+01- & 6.06e+01- & 2.11e+01- & 2.04e+01- & \textbf{8.10e-01} & 88.63 \\
          & 5000  & 5.80e+01- & 5.53e+01- & 4.71e+01- & 1.25e+01- & 2.88e+01- & 5.82e+01- & 1.24e+01- & 4.50e+01- & 4.55e+01- & 4.55e+01- & 6.10e+01- & 6.08e+01- & \textbf{8.10e-01} & 93.44 \\
    \midrule
    \multicolumn{2}{c}{(+/-/=)} & {0/16/0} & {0/16/0} & {0/16/0} & {0/16/0} & {0/16/0} & {0/16/0} & {0/16/0} & {0/16/0} & {0/16/0} & {0/16/0} & {0/16/0} & {0/16/0} &       &  \\
    \bottomrule
    \end{tabular}%
    \end{adjustbox}
  \label{tab: fe-zdt-lsmop}%
\end{table*}%

\begin{table*}[tbp]
  \centering
  \caption{HV Values Obtained by Compared Algorithms on 36 Instances From LSMOP Test Suite. The Best Result in Each Row is Highlighted in Bold. The Last Column Represents the Rate of Change of the Proposed PPM Compared to Suboptimal Results.}
  \begin{adjustbox}{width=1\hsize,left}
    \begin{tabular}{cccccccccccccccc}
    \toprule
    Problem & D     & NSGAII & CCGDE3 & DGEA  & WOF   & LMOCSO & CMOEAD & POCEA & CMOCSO & ABSAEA & CSEA  & EmoDM & MODE/D-LO    & PPM & ROC (\%) \\
    \midrule

    \multirow{4}[2]{*}{\parbox{1cm}{\centering ZDT6\\ M=2}} & 250   & 0.00e+00- & 0.00e+00- & 0.00e+00- & 0.00e+00- & 0.00e+00- & 8.15e-02- & 5.95e-02- & 7.00e-02- & 6.95e-02- & 6.17e-02- & 0.00e+00- & 6.63e-02- & \textbf{1.91E-01} & \textbf{11.53} \\
          & 2500  & 0.00e+00- & 0.00e+00- & 2.13e-02- & 0.00e+00- & 1.64e-02- & 0.00e+00- & 8.65e-02- & 3.69e-02- & 6.28e-02- & 0.00e+00- & 0.00e+00- & 8.47e-02- & \textbf{9.01E-02} & \textbf{5.09} \\
          & 3000  & 0.00e+00- & 0.00e+00- & 0.00e+00- & 3.03e-02- & 1.66e-02- & 0.00e+00- & 0.00e+00- & 5.84e-02- & 0.00e+00- & 0.00e+00- & 0.00e+00- & 0.00e+00- & \textbf{7.09E-02} & \textbf{55.65} \\
          & 5000  & 0.00e+00- & 0.00e+00- & 0.00e+00- & 6.99e-03- & 6.69e-03- & 0.00e+00- & 8.64e-02- & 5.23e-02- & 4.86e-02- & 0.00e+00- & 3.02e-02- & 0.00e+00- & \textbf{8.09E-02} & \textbf{5.21} \\
    \midrule
    \multicolumn{1}{c}{\multirow{4}[2]{*}{\parbox{1cm}{\centering LSMOP7\\ M=2}}} & 250   & 0.00e+00= & 0.00e+00= & 0.00e+00= & 4.79E-03+ & 6.77e-02+ & 7.06e-02+ & 6.65e-02+ & 0.00e+00= & 0.00e+00= & 0.00e+00= & 1.13e-02+ & 0.00e+00= & 0.00E+00 & -100.00 \\
          & 2500  & 0.00e+00= & 0.00e+00= & 1.87e-03+ & 0.00e+00= & 3.40e-02+ & 3.77e-02+ & 1.79e-02+ & 8.84e-02+ & 3.00e-02+ & 0.00e+00= & 5.33e-02+ & 0.00e+00= & 0.00E+00 & -100.00 \\
          & 3000  & 0.00e+00= & 0.00e+00= & 7.91e-02+ & 4.18e-02+ & 7.88e-02+ & 3.28e-02+ & 7.94e-02+ & 6.59e-02+ & 8.80e-02+ & 7.63e-02+ & 0.00e+00= & 0.00e+00= & 0.00E+00 & -100.00 \\
          & 5000  & 0.00e+00= & 0.00e+00= & 2.80e-02+ & 0.00e+00= & 0.00e+00= & 7.86e-02+ & 8.70e-02+ & 0.00e+00= & 0.00e+00= & 0.00e+00= & 4.70e-02+ & 4.67e-02+ & 0.00E+00 & -100.00 \\
    \midrule
    \multicolumn{1}{c}{\multirow{4}[2]{*}{\parbox{1cm}{\centering LSMOP8\\ M=2}}} & 250   & 0.00e+00- & 0.00e+00- & 0.00e+00- & 0.00e+00- & 0.00e+00- & 6.94e-02- & 0.00e+00- & 2.64e-02- & 6.04e-02- & 6.86e-02- & 4.11e-02- & 1.43e-02- & \textbf{9.09E-02} & \textbf{30.98} \\
          & 2500  & 0.00e+00- & 0.00e+00- & 4.27e-02- & 0.00e+00- & 0.00e+00- & 1.70e-02- & 7.30e-02- & 0.00e+00- & 0.00e+00- & 0.00e+00- & 6.26e-02- & 4.91e-02- & \textbf{9.09E-02} & \textbf{24.52} \\
          & 3000  & 0.00e+00- & 0.00e+00- & 4.37e-02- & 0.00e+00- & 3.43e-04- & 1.98e-02- & 6.89e-02- & 0.00e+00- & 0.00e+00- & 2.53e-02- & 0.00e+00- & 0.00e+00- & \textbf{9.09E-02} & \textbf{31.93} \\
          & 5000  & 0.00e+00- & 0.00e+00- & 2.55e-02- & 0.00e+00- & 0.00e+00- & 3.84e-02- & 0.00e+00- & 0.00e+00- & 0.00e+00- & 0.00e+00- & 1.41e-02- & 3.94e-02- & \textbf{9.09E-02} & \textbf{136.72} \\
    \midrule
    \multirow{4}[2]{*}{\parbox{1cm}{\centering LSMOP9\\ M=2}} & 250   & 0.00e+00- & 0.00e+00- & 0.00e+00- & 0.00e+00- & 4.85e-02- & 0.00e+00- & 5.39e-02- & 0.00e+00- & 0.00e+00- & 5.78e-02- & 3.85e-02- & 3.39e-02- & \textbf{9.09E-02} & \textbf{57.27} \\
          & 2500  & 0.00e+00- & 0.00e+00- & 1.76e-02- & 0.00e+00- & 7.17e-02- & 5.60e-02- & 1.35e-02- & 0.00e+00- & 0.00e+00- & 4.23e-02- & 1.95e-03- & 3.65e-02- & \textbf{9.09E-02} & \textbf{26.78} \\
          & 3000  & 0.00e+00- & 0.00e+00- & 0.00e+00- & 0.00e+00- & 0.00e+00- & 0.00e+00- & 0.00e+00- & 6.62e-02- & 2.23e-02- & 0.00e+00- & 2.05e-02- & 0.00e+00- & \textbf{9.09E-02} & \textbf{37.31} \\
          & 5000  & 0.00e+00- & 0.00e+00- & 2.21e-02- & 2.91e-02- & 4.99e-02- & 0.00e+00- & 0.00e+00- & 8.75e-02- & 4.71e-03- & 0.00e+00- & 4.35e-02- & 0.00e+00- & \textbf{9.09E-02} & \textbf{3.89} \\
    \midrule
    \multicolumn{2}{c}{(+/-/=)} & {0/12/4} & {0/12/4} & {3/12/1} & {2/12/2} & {3/12/1} & {4/12/0} & {4/12/0} & {2/12/2} & {2/12/2} & {1/12/3} & {3/12/1} & {1/12/3} \\
    \bottomrule
    \end{tabular}%
    \end{adjustbox}
  \label{tab: fe-zdt-lsmop-hv}%
\end{table*}%

\begin{table*}[tbp]
  \centering
  \caption{IGD Values Obtained by Compared Algorithms on 36 Instances From LSMOP Test Suite. The Best Result in Each Row is Highlighted in Bold. The Last Column Represents the Rate of Change of the Proposed PPM Compared to Suboptimal Results.}
  \begin{adjustbox}{width=1\hsize,left}
    \begin{tabular}{cccccccccccccccc}
    \toprule
    Problem & D     & NSGAII & CCGDE3 & DGEA  & WOF   & LMOCSO & CMOEAD & POCEA & CMOCSO & ABSAEA & CSEA  & EmoDM & MOEA/D-LO & PPM    & ROC (\%) \\
    \midrule
    \multicolumn{1}{c}{\multirow{4}[2]{*}{\parbox{1.3cm}{\centering LSMOP7\\ M=3}}} & 250   & 1.75e+00- & 1.77e+00- & 1.33e+00- & 1.78e+00- & 1.72e+00- & 1.71e+00- & 1.74e+00- & 1.95e+00- & 1.04e+02- & 9.70e+02- & 1.71e+03- & 5.13e+02- & \textbf{1.27e+00} & \textbf{4.25} \\
          & 2500  & 1.00e+00- & 1.00e+00- & 9.70e-01= & 1.00e+00- & 1.00e+00- & 1.00e+00- & 9.97e-01- & 1.01e+00- & 1.49e+02- & 1.13e+03- & 1.49e+02- & \textbf{7.07e-01+} & 9.37E-01 & -32.53 \\
          & 3000  & 9.93e-01- & 9.91e-01- & 9.71e-01= & 1.03e+03- & 9.91e-01- & 9.92e-01- & 9.87e-01- & 9.97e-01- & 7.90e+01- & 2.15e+03- & 1.46e+02- & 1.51e+02- & \textbf{9.33e-01} & \textbf{4.01} \\
          & 5000  & 9.73e-01= & 9.73e-01= & 9.61e-01= & 9.73e-01= & 9.70e-01= & 1.82e+03- & 9.70e-01= & 9.75e-01- & 1.20e+02- & 1.20e+02- & 1.20e+02- & 1.20e+02- & \textbf{9.24e-01} & \textbf{3.79} \\
    \midrule
    \multicolumn{1}{c}{\multirow{4}[2]{*}{\parbox{1.3cm}{\centering LSMOP7\\ M=5}}} & 250   & 4.48e+00- & 4.85e+00- & 4.52e+00- & 2.06e+02- & 4.17e+00- & 4.35e+00- & 3.06e+00- & 2.11e+05- & 2.51e+02- & 2.51e+02- & 1.10e+03- & 2.19e+03- & \textbf{1.82e+00} & \textbf{40.70} \\
          & 2500  & 1.41e+00- & 1.42e+00- & 1.17e+00= & 1.43e+00- & 1.41e+00- & 1.39e+00- & 1.40e+00- & 2.26e+05- & 2.89e+02- & 2.89e+02- & 2.89e+02- & \textbf{9.02e-01+} & 1.15E+00 & -27.67 \\
          & 3000  & 1.36e+00- & 1.36e+00- & 1.16e+00= & 1.38e+00- & 1.37e+00- & 6.21e+03- & 1.34e+00- & 2.64e+05- & 3.38e+02- & 3.38e+02- & 3.38e+02- & 3.69e+02- & \textbf{1.14e+00} & \textbf{2.16} \\
          & 5000  & 4.29e+04- & 2.69e+04- & 2.96e+01- & 1.73e+04- & 5.11e+03- & 1.39e+04- & 2.92e+03- & 2.33e+05- & 7.82e+02- & 7.82e+02- & 7.82e+02- & 7.82e+02- & \textbf{2.02e+00} & \textbf{93.16} \\
    \midrule
    \multicolumn{1}{c}{\multirow{4}[2]{*}{\parbox{1.3cm}{\centering LSMOP7\\ M=10}}} & 250   & 3.10e+04- & 2.64e+04- & 8.54e+02- & 5.21e+04- & 1.88e+03- & 1.71e+03- & 1.76e+03- & 2.17e+05- & 7.22e+03- & 2.16e+04- & 2.28e+04- & 1.76e+04- & \textbf{1.88e+00} & \textbf{99.78} \\
          & 2500  & 7.73e+04- & 5.61e+04- & 2.81e+01- & 7.00e+04- & 2.97e+03- & 1.58e+04- & 3.19e+03- & 1.73e+05- & 7.34e+02- & 7.34e+02- & 7.34e+02- & 1.10e+01- & \textbf{2.02e+00} & \textbf{81.66} \\
          & 3000  & 2.98e+04- & 6.46e+04- & 3.80e+01- & 1.75e+04- & 4.41e+03- & 1.24e+04- & 1.86e+03- & 3.98e+05- & 8.20e+02- & 8.20e+02- & 8.20e+02- & 1.25e+03- & \textbf{2.02e+00} & \textbf{94.69} \\
          & 5000  & 4.29e+04- & 2.69e+04- & 2.96e+01- & 1.73e+04- & 5.11e+03- & 1.39e+04- & 2.92e+03- & 2.33e+05- & 7.82e+02- & 7.82e+02- & 7.82e+02- & 7.82e+02- & \textbf{2.02e+00} & \textbf{93.16} \\
    \midrule
    \multicolumn{1}{c}{\multirow{4}[2]{*}{\parbox{1.3cm}{\centering LSMOP8\\ M=3}}} & 250   & 9.90e-01- & 9.91e-01- & 7.15e-01= & 9.40e-01- & 7.37e-01- & 9.29e-01- & \textbf{6.78e-01=} & 9.93e-01- & 1.04e+02- & 7.94e-01- & 1.71e+03- & 5.13e+02- & 6.79E-01 & -0.10 \\
          & 2500  & 9.53e-01- & 7.35e-01- & 7.10e-01- & 9.53e-01- & 5.95e-01= & \textbf{5.80e-01=} & 5.98e-01= & 9.55e-01- & 1.49e+02- & 6.67e-01- & 7.07e-01- & 6.18e-01= & 6.16E-01 & -6.15 \\
          & 3000  & 9.52e-01- & 9.52e-01- & 6.45e-01- & 9.52e-01- & 5.82e-01= & 6.00e-01- & 6.74e-01- & 9.55e-01- & 7.90e+01- & 6.37e-01- & 1.46e+02- & 1.51e+02- & \textbf{5.38e-01} & \textbf{7.61} \\
          & 5000  & 7.11e-01- & 9.51e-01- & 5.80e-01= & 8.05e-01- & 5.70e-01= & 5.81e-01= & 6.31e-01- & 9.53e-01- & 1.20e+02- & 1.20e+02- & 1.20e+02- & 1.20e+02- & \textbf{5.47e-01} & \textbf{3.99} \\
    \midrule
    \multicolumn{1}{c}{\multirow{4}[2]{*}{\parbox{1.3cm}{\centering LSMOP8\\ M=5}}} & 250   & 1.29e+00- & 1.23e+00- & 1.14e+00- & 1.17e+00- & 1.16e+00- & 1.10e+00- & 1.16e+00- & 1.76e+01- & 2.51e+02- & 2.51e+02- & 1.10e+03- & 2.19e+03- & \textbf{9.92e-01} & \textbf{9.92} \\
          & 2500  & 1.13e+00- & 1.01e+00- & 1.00e+00- & \textbf{8.28e-01+} & 9.90e-01- & 8.65e-01= & 8.79e-01= & 1.13e+00- & 2.89e+02- & 2.89e+02- & 9.02e-01= & 9.34e-01= & 8.99E-01 & -8.54 \\
          & 3000  & 1.13e+00- & 1.13e+00- & 1.12e+00- & 9.50e-01= & 8.62e-01= & \textbf{7.94e-01+} & 8.42e-01+ & 2.73e+00- & 3.38e+02- & 3.38e+02- & 3.38e+02- & 3.69e+02- & 9.08E-01 & -14.33 \\
          & 5000  & 1.57e+01- & 1.40e+01- & 1.93e+00- & 1.37e+01- & 3.96e+00- & 8.87e+00- & 3.66e+00- & 3.25e+01- & 7.82e+02- & 7.82e+02- & 7.82e+02- & 7.82e+02- & \textbf{1.24e+00} & \textbf{35.72} \\
    \midrule
    \multicolumn{1}{c}{\multirow{4}[2]{*}{\parbox{1.3cm}{\centering LSMOP8\\ M=10}}} & 250   & 1.62e+01- & 1.37e+01- & 4.61e+00- & 9.46e+00- & 2.74e+00- & 3.09e+00- & 2.09e+00- & 3.86e+01- & 7.22e+03- & 2.16e+04- & 2.28e+04- & 1.76e+04- & \textbf{1.24e+00} & \textbf{40.62} \\
          & 2500  & 1.53e+01- & 1.63e+01- & 1.68e+00- & 1.55e+01- & 5.81e+00- & 8.03e+00- & 3.66e+00- & 2.75e+01- & 7.34e+02- & 7.34e+02- & 1.10e+01- & 1.10e+01- & \textbf{1.23e+00} & \textbf{26.56} \\
          & 3000  & 1.55e+01- & 1.40e+01- & 2.32e+00- & 1.22e+01- & 3.67e+00- & 8.69e+00- & 3.32e+00- & 1.72e+01- & 8.20e+02- & 8.20e+02- & 8.20e+02- & 1.25e+03- & \textbf{1.23e+00} & \textbf{46.84} \\
          & 5000  & 1.57e+01- & 1.40e+01- & 1.93e+00- & 1.37e+01- & 3.96e+00- & 8.87e+00- & 3.66e+00- & 3.25e+01- & 7.82e+02- & 7.82e+02- & 7.82e+02- & 7.82e+02- & \textbf{1.24e+00} & \textbf{35.72} \\
    \midrule
    \multicolumn{1}{c}{\multirow{4}[2]{*}{\parbox{1.3cm}{\centering LSMOP9\\ M=3}}} & 250   & 9.11e+01- & 1.06e+02- & 7.26e+01- & 8.63e+01- & 4.35e+01- & 8.13e+01- & 3.92e+01- & 1.04e+02- & 1.04e+02- & 1.33e+02- & 1.71e+03- & 5.13e+02- & \textbf{1.54e+00} & \textbf{96.07} \\
          & 2500  & 1.38e+02- & 1.34e+02- & 1.09e+02- & 1.33e+02- & 6.92e+01- & 1.32e+02- & 5.27e+01- & 9.42e+01- & 1.49e+02- & 1.49e+02- & 7.07e-01+ & \textbf{6.18e-01+} & 1.54E+00 & -148.84 \\
          & 3000  & 1.37e+02- & 1.39e+02- & 7.08e+01- & 1.40e+02- & 7.34e+01- & 1.37e+02- & 5.02e+01- & 7.90e+01- & 7.90e+01- & 1.46e+02- & 1.51e+02- & 1.51e+02- & \textbf{1.54e+00} & \textbf{96.93} \\
          & 5000  & 1.46e+02- & 1.39e+02- & 6.07e+01- & 1.39e+02- & 8.70e+01- & 1.41e+02- & 6.64e+01- & 9.88e+01- & 1.20e+02- & 1.20e+02- & 1.20e+02- & 1.20e+02- & \textbf{1.54e+00} & \textbf{97.47} \\
    \midrule
    \multicolumn{1}{c}{\multirow{4}[2]{*}{\parbox{1.3cm}{\centering LSMOP9\\ M=5}}} & 250   & 2.45e+02- & 2.62e+02- & 1.31e+02- & 2.57e+02- & 1.39e+02- & 1.80e+02- & 1.01e+02- & 2.51e+02- & 2.51e+02- & 2.51e+02- & 1.10e+03- & 2.19e+03- & \textbf{3.00e+00} & \textbf{97.02} \\
          & 2500  & 3.47e+02- & 3.38e+02- & 1.68e+02- & 1.34e+02- & 2.12e+02- & 3.24e+02- & 1.53e+02- & 2.89e+02- & 2.89e+02- & 2.89e+02- & \textbf{9.02e-01+} & 9.34e-01+ & 3.00E+00 & -232.79 \\
          & 3000  & 3.49e+02- & 3.35e+02- & 1.97e+02- & 1.25e+02- & 2.20e+02- & 3.11e+02- & 1.83e+02- & 3.38e+02- & 3.38e+02- & 3.38e+02- & 3.69e+02- & 3.80e+02- & \textbf{2.99e+00} & \textbf{97.61} \\
          & 5000  & 1.26e+03- & 1.14e+03- & 6.01e+02- & 1.22e+03- & 6.84e+02- & 1.10e+03- & 7.35e+02- & 7.82e+02- & 7.82e+02- & 7.82e+02- & 7.82e+02- & 7.82e+02- & \textbf{6.53e+00} & \textbf{98.91} \\
    \midrule
    \multicolumn{1}{c}{\multirow{4}[2]{*}{\parbox{1.3cm}{\centering LSMOP9\\ M=10}}} & 250   & 9.94e+02- & 1.05e+03- & 4.70e+02- & 1.42e+03- & 5.39e+02- & 5.37e+02- & 3.32e+02- & 8.55e+02- & 7.22e+03- & 2.16e+04- & 2.28e+04- & 1.76e+04- & \textbf{6.52e+00} & \textbf{98.04} \\
          & 2500  & 1.21e+03- & 1.17e+03- & 7.56e+02- & 1.14e+03- & 6.42e+02- & 9.63e+02- & 6.12e+02- & 7.34e+02- & 7.34e+02- & 7.34e+02- & 1.10e+01- & 1.10e+01- & \textbf{6.53e+00} & \textbf{40.60} \\
          & 3000  & 1.17e+03- & 1.15e+03- & 6.72e+02- & 1.29e+03- & 6.82e+02- & 9.73e+02- & 7.07e+02- & 8.20e+02- & 8.20e+02- & 8.20e+02- & 1.25e+03- & 1.28e+03- & \textbf{6.53e+00} & \textbf{99.03} \\
          & 5000  & 1.26e+03- & 1.14e+03- & 6.01e+02- & 1.22e+03- & 6.84e+02- & 1.10e+03- & 7.35e+02- & 7.82e+02- & 7.82e+02- & 7.82e+02- & 7.82e+02- & 7.82e+02- & \textbf{6.53e+00} & \textbf{98.91} \\
    \midrule
    \multicolumn{2}{c}{(+/-/=)} & {0/35/1} & {0/35/1} & {0/29/7} & {1/33/2} & {0/31/5} & {1/32/3} & {1/31/4} & {0/36/0} & {0/36/0} & {0/36/0} & {2/33/1} & {4/30/2} &       &  \\
    \bottomrule
    \end{tabular}%

    \end{adjustbox}
  \label{tab: igd-many}%
\end{table*}%

\begin{table*}[tbp]
  \centering
  \caption{HV Values Obtained by Compared Algorithms on 36 Instances From LSMOP Test Suite. The Best Result in Each Row is Highlighted in Bold. The Last Column Represents the Rate of Change of the Proposed PPM Compared to Suboptimal Results.}
  \begin{adjustbox}{width=1\hsize,left}
    \begin{tabular}{cccccccccccccccc}
    \toprule
    Problem & D     & NSGA-II & CCGDE3 & DGEA  & WOF   & LMOCSO & CMOEAD & POCEA & CMOCSO & ABSAEA & CSEA  & EmoDM & MOEA/D-LO & PPM    & ROC (\%) \\
    \midrule
    \multicolumn{1}{c}{\multirow{4}[2]{*}{\parbox{1.3cm}{\centering LSMOP7\\ M=3}}} & 250   & 0.00e+00= & 0.00e+00= & 0.00e+00= & 0.00e+00= & 0.00e+00= & 0.00e+00= & 0.00e+00= & 0.00e+00= & 0.00e+00= & 0.00e+00= & 0.00e+00= & 0.00e+00= & 0.00e+00 & NaN \\
          & 2500  & 0.00e+00= & 0.00e+00= & \textbf{4.78e-02=} & 0.00e+00= & 0.00e+00= & 0.00e+00= & 4.23e-03= & 0.00e+00= & 0.00e+00= & 0.00e+00= & 0.00e+00= & 0.00e+00= & 3.63E-02 & -24.14 \\
          & 3000  & 1.05e-02= & 1.28e-02= & 4.50e-02= & 0.00e+00= & 1.38e-02= & 1.23e-02= & 1.91e-02= & 3.43e-03= & 0.00e+00= & 0.00e+00= & 0.00e+00= & 0.00e+00= & \textbf{4.51e-02} & \textbf{0.04} \\
          & 5000  & 4.32e-02= & 4.25e-02= & 6.35e-02= & 4.17e-02= & 4.73e-02= & 0.00e+00- & 4.76e-02= & 3.84e-02= & 0.00e+00- & 0.00e+00- & 0.00e+00- & 0.00e+00- & \textbf{6.36e-02} & \textbf{0.01} \\
    \midrule
    \multicolumn{1}{c}{\multirow{4}[2]{*}{\parbox{1.3cm}{\centering LSMOP7\\ M=5}}} & 250   & 0.00e+00= & 0.00e+00= & 0.00e+00= & 0.00e+00= & 0.00e+00= & 0.00e+00= & 0.00e+00= & 0.00e+00= & 0.00e+00= & 0.00e+00= & 0.00e+00= & 0.00e+00= & 0.00e+00 & NaN \\
          & 2500  & 0.00e+00= & 0.00e+00= & \textbf{2.76e-03=} & 0.00e+00= & 0.00e+00= & 0.00e+00= & 0.00e+00= & 0.00e+00= & 0.00e+00= & 0.00e+00= & 0.00e+00= & 0.00e+00= & 0.00e+00 & -100.00 \\
          & 3000  & 0.00e+00= & 0.00e+00= & 8.89e-03= & 0.00e+00= & 0.00e+00= & 0.00e+00= & 0.00e+00= & 0.00e+00= & 0.00e+00= & 0.00e+00= & 0.00e+00= & 0.00e+00= & \textbf{1.76e-02} & \textbf{97.52} \\
          & 5000  & 0.00e+00= & 0.00e+00= & 0.00e+00= & 0.00e+00= & 0.00e+00= & 0.00e+00= & 0.00e+00= & 0.00e+00= & 0.00e+00= & 0.00e+00= & 0.00e+00= & 0.00e+00= & 0.00e+00     & NaN \\
    \midrule
    \multicolumn{1}{c}{\multirow{4}[2]{*}{\parbox{1.3cm}{\centering LSMOP7\\ M=10}}} & 250   & 0.00e+00= & 0.00e+00= & 0.00e+00= & 0.00e+00= & 0.00e+00= & 0.00e+00= & 0.00e+00= & 0.00e+00= & 0.00e+00= & 0.00e+00= & 0.00e+00= & 0.00e+00= & 0.00e+00 & NaN \\
          & 2500  & 0.00e+00= & 0.00e+00= & 0.00e+00= & 0.00e+00= & 0.00e+00= & 0.00e+00= & 0.00e+00= & 0.00e+00= & 0.00e+00= & 0.00e+00= & 0.00e+00= & 0.00e+00= & 0.00e+00 & NaN \\
          & 3000  & 0.00e+00= & 0.00e+00= & 0.00e+00= & 0.00e+00= & 0.00e+00= & 0.00e+00= & 0.00e+00= & 0.00e+00= & 0.00e+00= & 0.00e+00= & 0.00e+00= & 0.00e+00= & 0.00e+00 & NaN \\
          & 5000  & 0.00e+00= & 0.00e+00= & 0.00e+00= & 0.00e+00= & 0.00e+00= & 0.00e+00= & 0.00e+00= & 0.00e+00= & 0.00e+00= & 0.00e+00= & 0.00e+00= & 0.00e+00= & 0.00e+00 & NaN \\
    \midrule
    \multicolumn{1}{c}{\multirow{4}[2]{*}{\parbox{1.3cm}{\centering LSMOP8\\ M=3}}} & 250   & 1.40e-02- & 1.39e-02- & 2.11e-02- & 1.80e-02- & 1.73e-02- & 1.76e-02- & 1.75e-02- & 1.03e-02- & 0.00e+00- & 0.00e+00- & 0.00e+00- & 0.00e+00- & \textbf{1.11e-01} & \textbf{427.24} \\
          & 2500  & 7.80e-02- & 7.82e-02- & 8.36e-02- & 7.81e-02- & 8.14e-02- & 7.89e-02- & 8.17e-02- & 7.38e-02- & 0.00e+00- & 0.00e+00- & 0.00e+00- & 0.00e+00- & \textbf{1.68e-01} & \textbf{101.01} \\
          & 3000  & 7.91e-02- & 7.86e-02- & 8.47e-02- & 7.95e-02- & 8.06e-02- & 7.98e-02- & 8.16e-02- & 7.44e-02- & 0.00e+00- & 0.00e+00- & 0.00e+00- & 0.00e+00- & \textbf{1.68e-01} & \textbf{98.10} \\
          & 5000  & 8.19e-02- & 8.13e-02- & 8.37e-02- & 8.15e-02- & 8.51e-02- & 8.21e-02- & 8.42e-02- & 7.70e-02- & 0.00e+00- & 0.00e+00- & 0.00e+00- & 0.00e+00- & \textbf{1.70e-01} & \textbf{100.29} \\
    \midrule
    \multicolumn{1}{c}{\multirow{4}[1]{*}{\parbox{1.3cm}{\centering LSMOP8\\ M=5}}} & 250   & 0.00e+00- & 0.00e+00- & 0.00e+00- & 0.00e+00- & 0.00e+00- & 0.00e+00- & 1.57e-02- & 0.00e+00- & 0.00e+00- & 0.00e+00- & 0.00e+00- & 0.00e+00- & \textbf{9.09e-02} & \textbf{477.38} \\
          & 2500  & 5.59e-02- & 5.22e-02- & 7.58e-02- & 2.35e-02- & 2.17e-02- & 5.86e-02- & 5.86e-02- & 5.04e-02- & 0.00e+00- & 0.00e+00- & 0.00e+00- & 0.00e+00- & \textbf{1.64e-01} & \textbf{116.62} \\
          & 3000  & 6.09e-02- & 6.06e-02- & 7.83e-02- & 6.21e-02- & 6.27e-02- & 6.22e-02- & 6.29e-02- & 0.00e+00- & 0.00e+00- & 0.00e+00- & 0.00e+00- & 0.00e+00- & \textbf{1.63e-01} & \textbf{107.90} \\
          & 5000  & 0.00e+00- & 0.00e+00- & 0.00e+00- & 0.00e+00- & 0.00e+00- & 0.00e+00- & 0.00e+00- & 0.00e+00- & 0.00e+00- & 0.00e+00- & 0.00e+00- & 0.00e+00- & \textbf{9.09e-02} & \textbf{Inf} \\
    \midrule
    \multicolumn{1}{c}{\multirow{4}[0]{*}{\parbox{1.3cm}{\centering LSMOP8\\ M=10}}} & 250   & 3.83e-02- & 4.79e-02- & 4.82e-02- & 5.67e-03- & 4.68e-02- & 3.87e-02- & 2.24e-02- & 2.94e-02- & 3.91e-02- & 2.11e-02- & 1.97e-02- & 4.98e-03- & \textbf{9.09e-02} & \textbf{88.59} \\
            & 2500  & 2.17e-02- & 4.39e-02- & 3.34e-02- & 2.80e-02- & 2.63e-03- & 4.49e-02- & 3.62e-02- & 2.08e-02- & 3.14e-02- & 1.03e-02- & 8.98e-03- & 5.09e-04- & \textbf{9.09e-02} & \textbf{102.45} \\
            & 3000  & 1.90e-02- & 3.45e-02- & 4.39e-02- & 1.22e-03- & 2.50e-02- & 4.33e-02- & 3.61e-03- & 3.41e-02- & 3.75e-02- & 6.14e-04- & 4.44e-02- & 4.04e-02- & \textbf{9.09e-02} & \textbf{104.73} \\
            & 5000  & 2.92e-02- & 3.85e-02- & 3.44e-02- & 3.36e-02- & 2.19e-02- & 4.90e-02- & 4.57e-02- & 1.72e-02- & 3.58e-02- & 2.59e-02- & 1.92e-02- & 2.54e-04- & \textbf{9.09e-02} & \textbf{85.51} \\
    \midrule
    \multicolumn{1}{c}{\multirow{4}[0]{*}{\parbox{1.3cm}{\centering LSMOP9\\ M=3}}} & 250   & 2.26e-02- & 2.00e-02- & 1.91e-02- & 1.47e-02- & 4.31e-02- & 3.08e-02- & 3.92e-02- & 3.05e-02- & 2.13e-02- & 3.39e-02- & 4.75e-02- & 6.75e-03- & \textbf{9.09e-02} & \textbf{91.37} \\
            & 2500  & 2.63e-02- & 4.80e-03- & 1.32e-02- & 5.88e-03- & 2.83e-02- & 2.82e-02- & 2.90e-02- & 4.77e-02- & 8.36e-03- & 1.68e-02- & 4.03e-02- & 2.36e-02- & \textbf{9.09e-02} & \textbf{90.57} \\
            & 3000  & 1.45e-02- & 1.27e-02- & 2.55e-02- & 3.26e-04- & 1.32e-02- & 4.51e-03- & 3.99e-02- & 2.92e-02- & 1.81e-02- & 2.18e-02- & 3.05e-02- & 1.06e-02- & \textbf{9.09e-02} & \textbf{127.82} \\
            & 5000  & 3.35e-02- & 2.88e-02- & 3.43e-02- & 4.09e-03- & 1.22e-02- & 2.46e-02- & 8.17e-03- & 1.28e-02- & 4.63e-02- & 3.55e-02- & 2.80e-02- & 6.79e-03- & \textbf{9.09e-02} & \textbf{96.33} \\
    \midrule
    \multicolumn{1}{c}{\multirow{4}[0]{*}{\parbox{1.3cm}{\centering LSMOP9\\ M=5}}} & 250   & 3.64e-03- & 3.98e-03- & 3.82e-02- & 1.09e-02- & 3.02e-02- & 3.73e-03- & 1.43e-02- & 2.10e-02- & 4.82e-02- & 2.72e-02- & 4.54e-02- & 2.72e-02- & \textbf{9.09e-02} & \textbf{88.59} \\
            & 2500  & 4.48e-02- & 1.33e-02- & 4.61e-02- & 2.73e-02- & 2.39e-02- & 2.09e-03- & 2.66e-03- & 2.57e-02- & 7.38e-04- & 1.78e-02- & 1.68e-02- & 4.11e-02- & \textbf{9.09e-02} & \textbf{97.18} \\
            & 3000  & 2.36e-03- & 1.44e-02- & 8.32e-04- & 3.11e-03- & 2.87e-02- & 3.05e-02- & 1.66e-02- & 4.05e-03- & 7.38e-04- & 8.48e-03- & 4.49e-02- & 3.53e-03- & \textbf{9.19e-02} & \textbf{104.68} \\
            & 5000  & 4.78e-03- & 7.52e-03- & 3.16e-02- & 2.13e-02- & 1.15e-02- & 1.25e-02- & 1.35e-02- & 4.94e-02- & 4.88e-02- & 3.13e-02- & 2.04e-02- & 3.85e-02- & \textbf{9.09e-02} & \textbf{84.01} \\
    \midrule
    \multicolumn{1}{c}{\multirow{4}[1]{*}{\parbox{1.3cm}{\centering LSMOP9\\ M=10}}} & 250   & 3.72e-02- & 2.78e-02- & 3.34e-02- & 3.92e-02- & 5.73e-05- & 9.13e-03- & 2.06e-02- & 2.53e-02- & 1.34e-02- & 4.58e-02- & 1.68e-02- & 4.35e-02- & \textbf{9.16e-02} & \textbf{100} \\
            & 2500  & 4.86e-02- & 3.76e-02- & 4.75e-02- & 4.00e-02- & 2.48e-02- & 1.65e-02- & 2.24e-02- & 1.42e-02- & 4.09e-02- & 5.00e-02- & 4.92e-02- & 1.34e-02- & \textbf{9.09e-02} & \textbf{81.8} \\
            & 3000  & 1.96e-02- & 8.24e-03- & 1.53e-03- & 1.39e-02- & 3.19e-02- & 4.05e-02- & 1.80e-02- & 2.80e-02- & 2.23e-02- & 1.56e-02- & 3.42e-02- & 1.31e-02- & \textbf{9.09e-02} & \textbf{124.44} \\
            & 5000  & 3.41e-02- & 3.19e-02- & 2.80e-02- & 4.91e-02- & 3.02e-02- & 1.93e-02- & 1.11e-02- & 7.05e-03- & 2.84e-02- & 1.03e-02- & 2.60e-02- & 1.36e-02- & \textbf{9.09e-02} & \textbf{85.13} \\
    \midrule
    \multicolumn{2}{c}{(+/-/=)} & {0/24/12} & {0/24/12} & {0/24/12} & {0/24/12} & {0/24/12} & {0/25/11} & {0/24/12} & {0/24/12} & {0/25/11} & {0/25/11} & {0/25/11} & {0/25/11} &       &  \\
    \bottomrule
    \end{tabular}%
    \end{adjustbox}
  \label{tab: hv-many}%
\end{table*}%
\subsubsection{Experimental Results on Large-scale MOPs}
We evaluated the generalization performance of the PPM on various problems, including ZDT and LSMOP. The obtained IGD values and HV values for all competing algorithms are presented in Table \ref{tab: fe-zdt-lsmop} and Table \ref{tab: fe-zdt-lsmop-hv}, respectively. We tested on ZDT6 and LSMOP7-9, which are different from the training problems, and scaled the dimension up to 5,000.
\par
Table \ref{tab: fe-zdt-lsmop} reveals that PPM exhibits notable advantages in solving ZDT 6 with dimensions ranging from 250 to 5,000. Compared to the suboptimal algorithm, PPM showcases a remarkable improvement of 91\%. Conversely, most of the compared algorithms show worse performance. In particular, AB-SAEA~\citep{WANG2020317} and CSEA~\citep{8281523} fail to produce solutions even at dimension scales of 100 and 5,000. This observation validates the limitations of existing surrogate models in capturing the complex mapping between numerous decision variables and multiple objective functions \citep{10.1145/3470971}. 
\par
Table \ref{tab: fe-zdt-lsmop-hv} showcases the algorithm's performance in terms of HV. Notably, while all algorithms, including PPM, yield zero HV values for the LSMOP7 problem, the proposed algorithm achieves non-zero HV values for ZDT6, LSMOP8, and LSMOP9, outperforming the compared algorithms. This outcome demonstrates PPM's superior performance compared to other algorithms. The HV values attained by PPM remain relatively consistent, primarily due to the limited number of solutions that approximate the Pareto front. Although this suggests that PPM may not exhibit perfect diversity, it is worth noting that it still outperforms existing algorithms.

\subsubsection{Experimental Results on Large-scale MaOPs}
We further expand the complexity of complex MOPs, and we present the IGD and HV values obtained by all algorithms in solving expensive, large-scale, many-objective optimization problems in Tables \ref{tab: igd-many} and \ref{tab: hv-many}. The tested problem is scaled up to ten objectives.
\par
The results show that there are advantages in many-objective (up to 10 objectives), large-scale decision variables (250-5,000 dimensions), and computationally expensive functions (only 1,000 function evaluations). Table \ref{tab: igd-many} shows that the proposed PPM significantly outperforms compared algorithms on computationally expensive large-scale many-objective optimization problems (2,500 decision variables with 10 objectives and 1,000 function evaluations). We noticed that WOF, CMOEAD, and MOEA/D-LO achieved several of the best results with the IGD indicator, but their overall performance is still unsatisfactory.
\par
Although LSMOP~1-9 is a series of test sets, different test functions under this series have different properties. For example, different problems have different landscapes and their decision variables also have different separability~\citep{7553457}, while PPM can utilize learned patterns on evolutionary computation on LSMOP~1-6 and fine-tuning on LSMOP~7-9 problems, achieving superior results, which shows that the pre-training and fine-tuning are a valid paradigm in solving complex MOPs.
\par
\begin{figure*}[tbp]
    \centering
    \subfloat[{Problem ZDT6}]{\includegraphics[width=0.22\hsize]{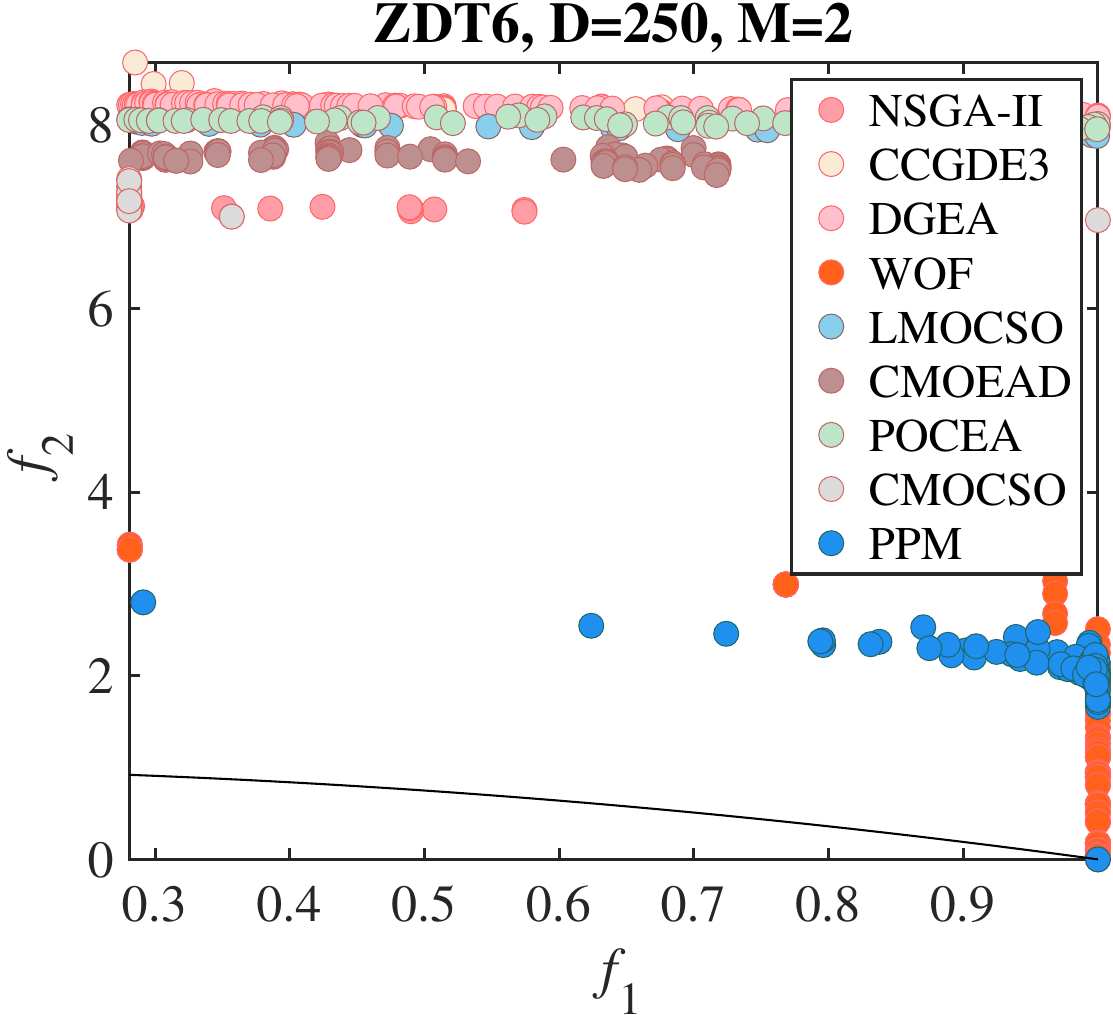}}
    \subfloat[{Problem LSMOP7}]{\includegraphics[width=0.24\hsize]{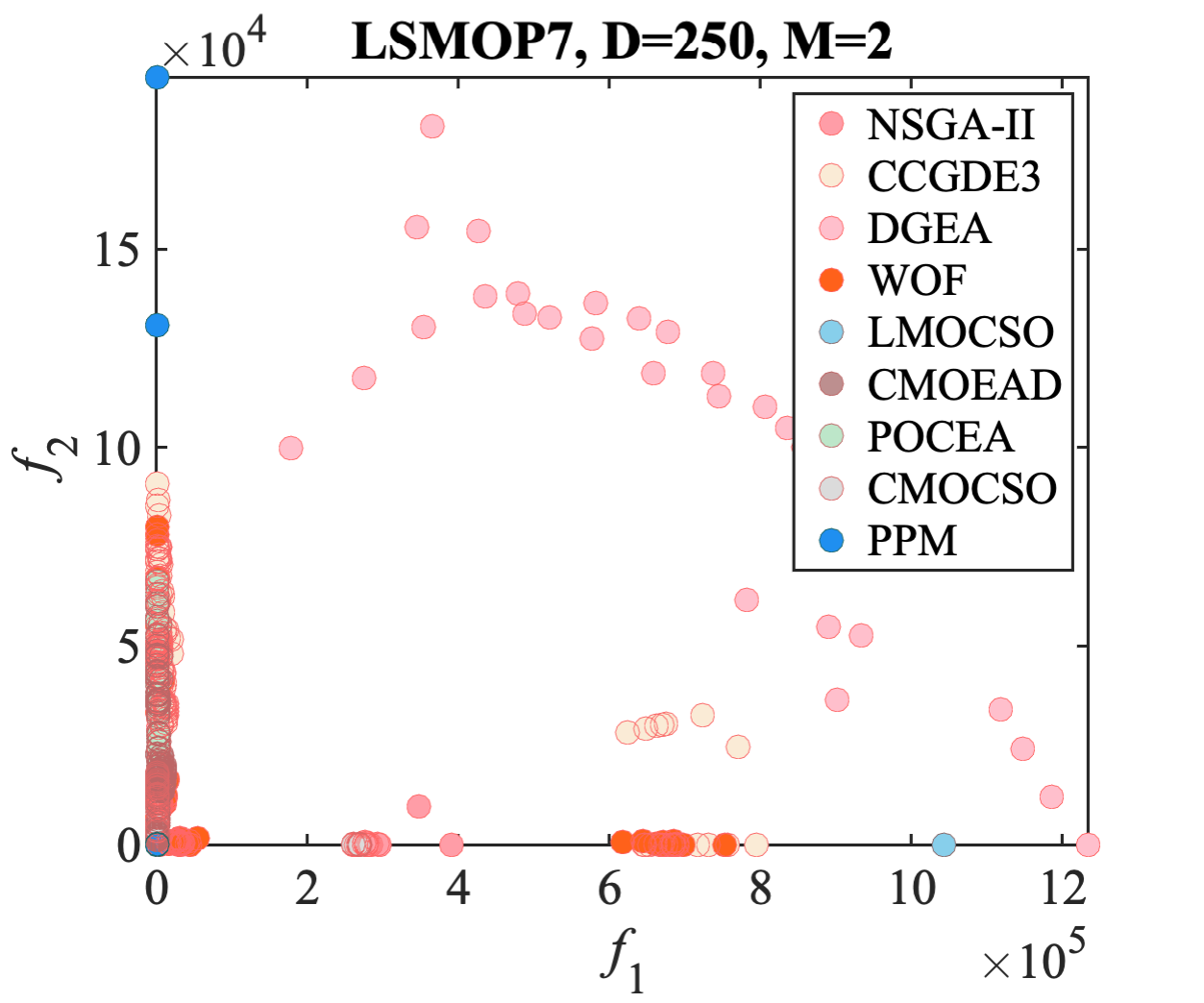}}
    \subfloat[{Problem LSMOP8}]{\includegraphics[width=0.24\hsize]{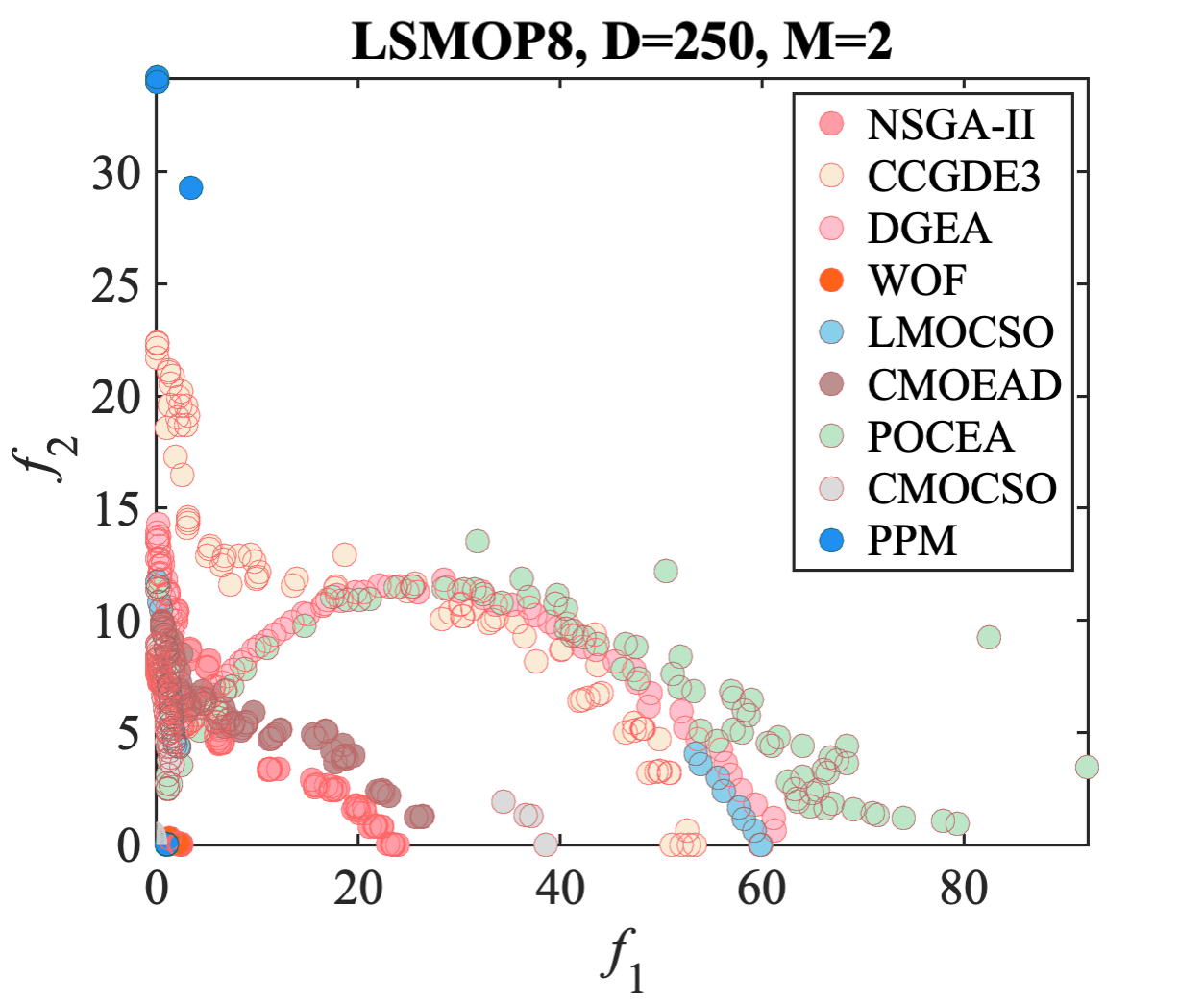}}
    \subfloat[{Problem LSMOP9}]{\includegraphics[width=0.24\hsize]{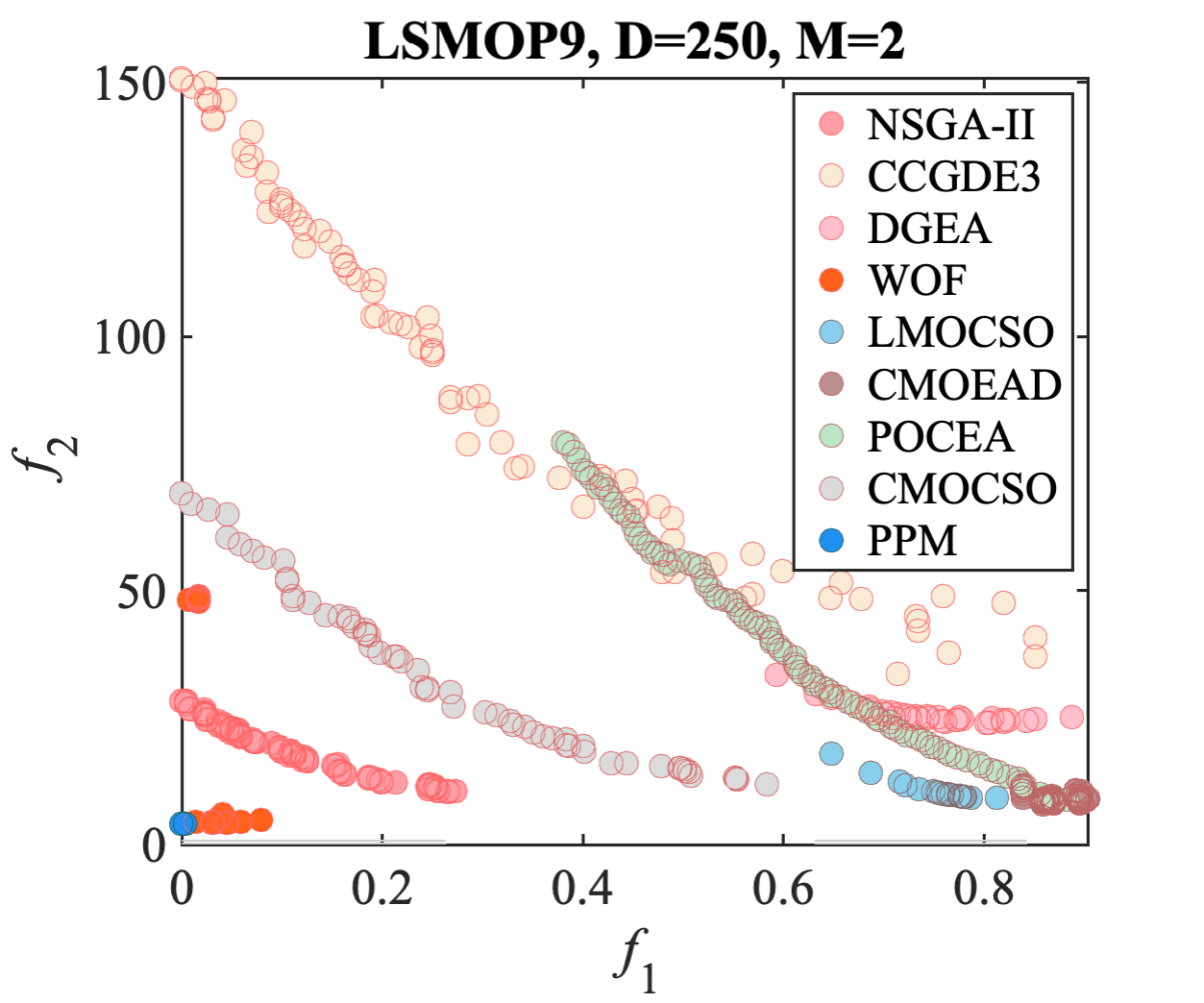}}
    \\
    \subfloat[{Problem ZDT6\\~~ (Local Enlarged) }]{\includegraphics[width=0.22\hsize]{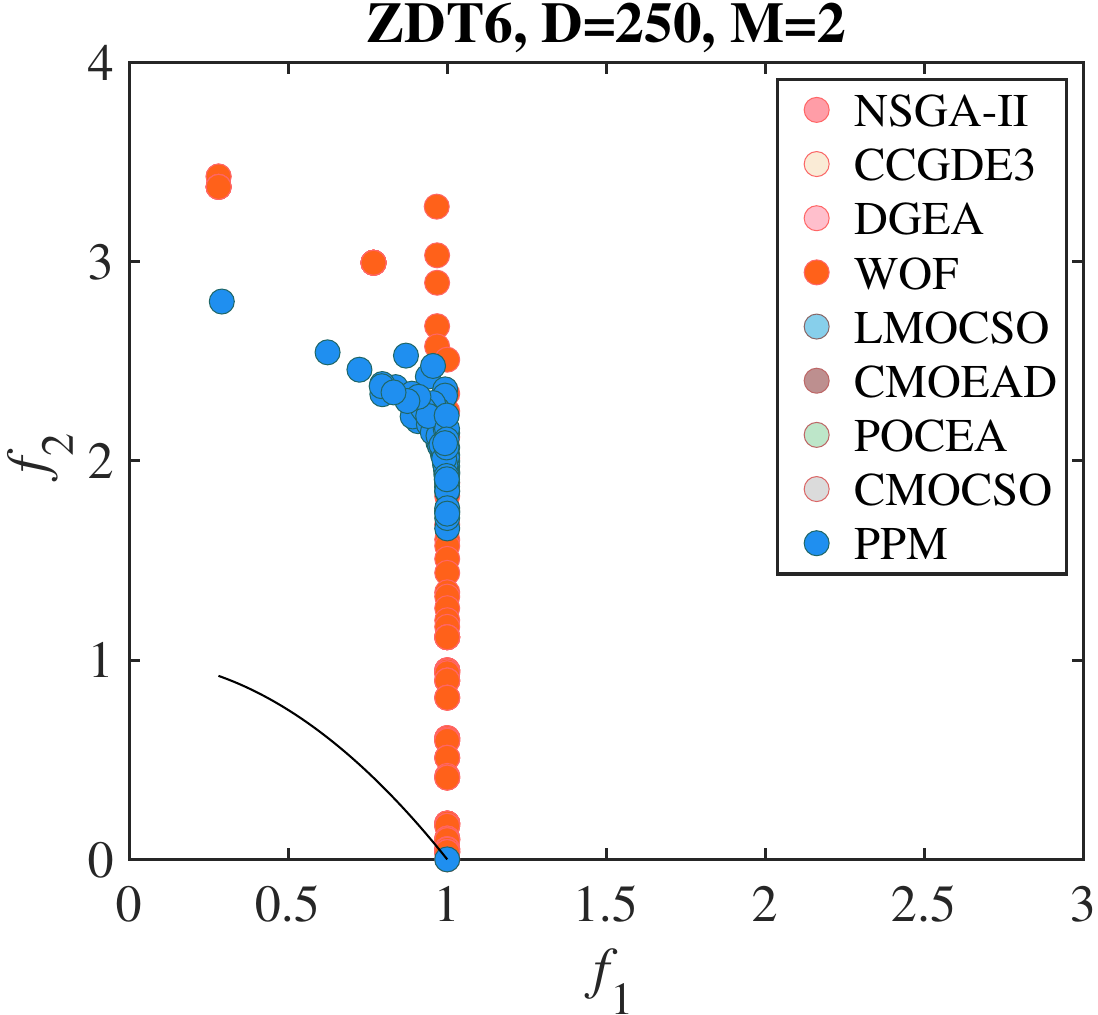}}
    \subfloat[{Problem LSMOP7\\~~ (Local Enlarged) }]{\includegraphics[width=0.24\hsize]{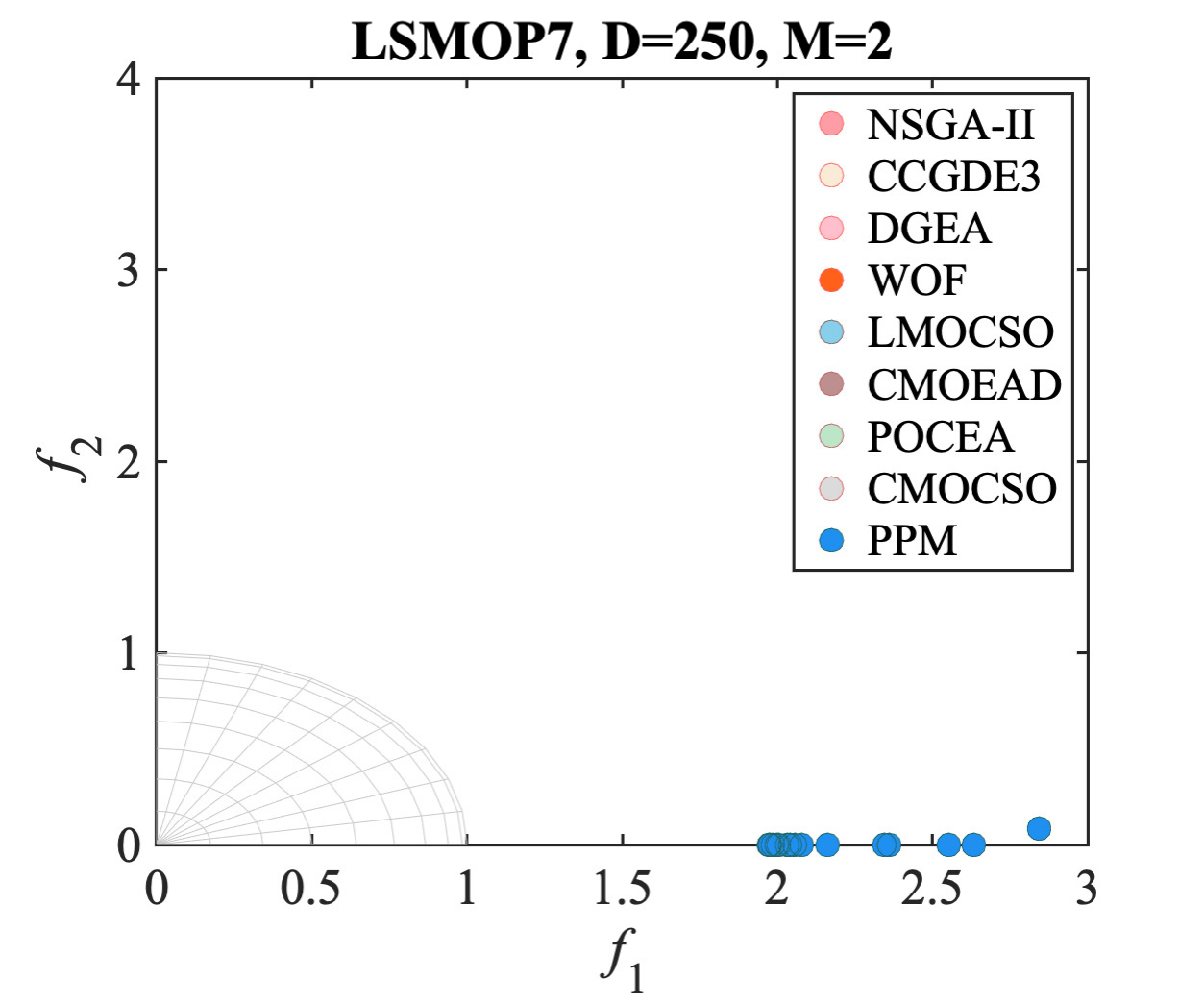}}
    \subfloat[{Problem LSMOP8\\~~ (Local Enlarged) }]{\includegraphics[width=0.24\hsize]{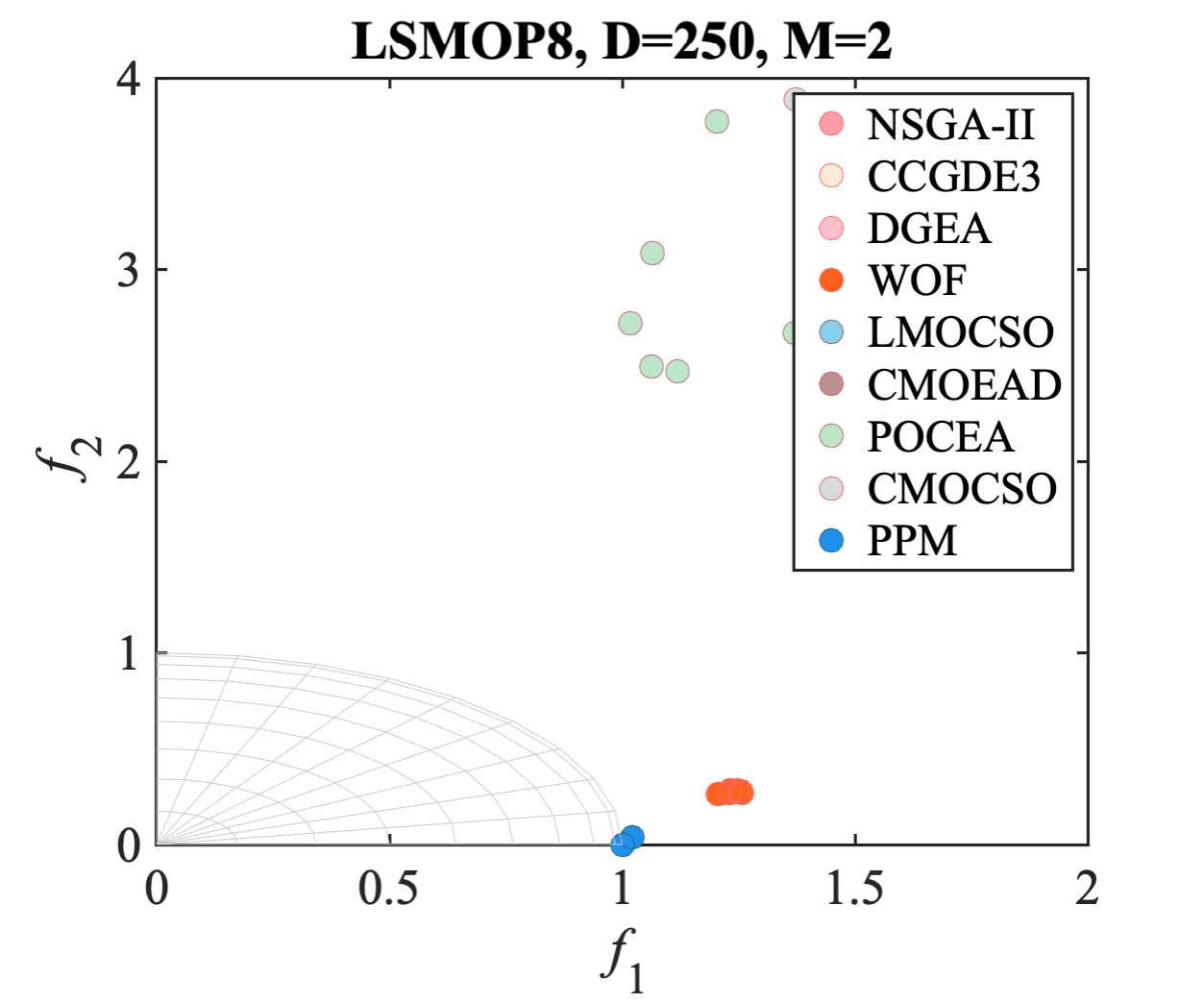}}
    \subfloat[{Problem LSMOP9\\~~ (Local Enlarged) }]{\includegraphics[width=0.24\hsize]{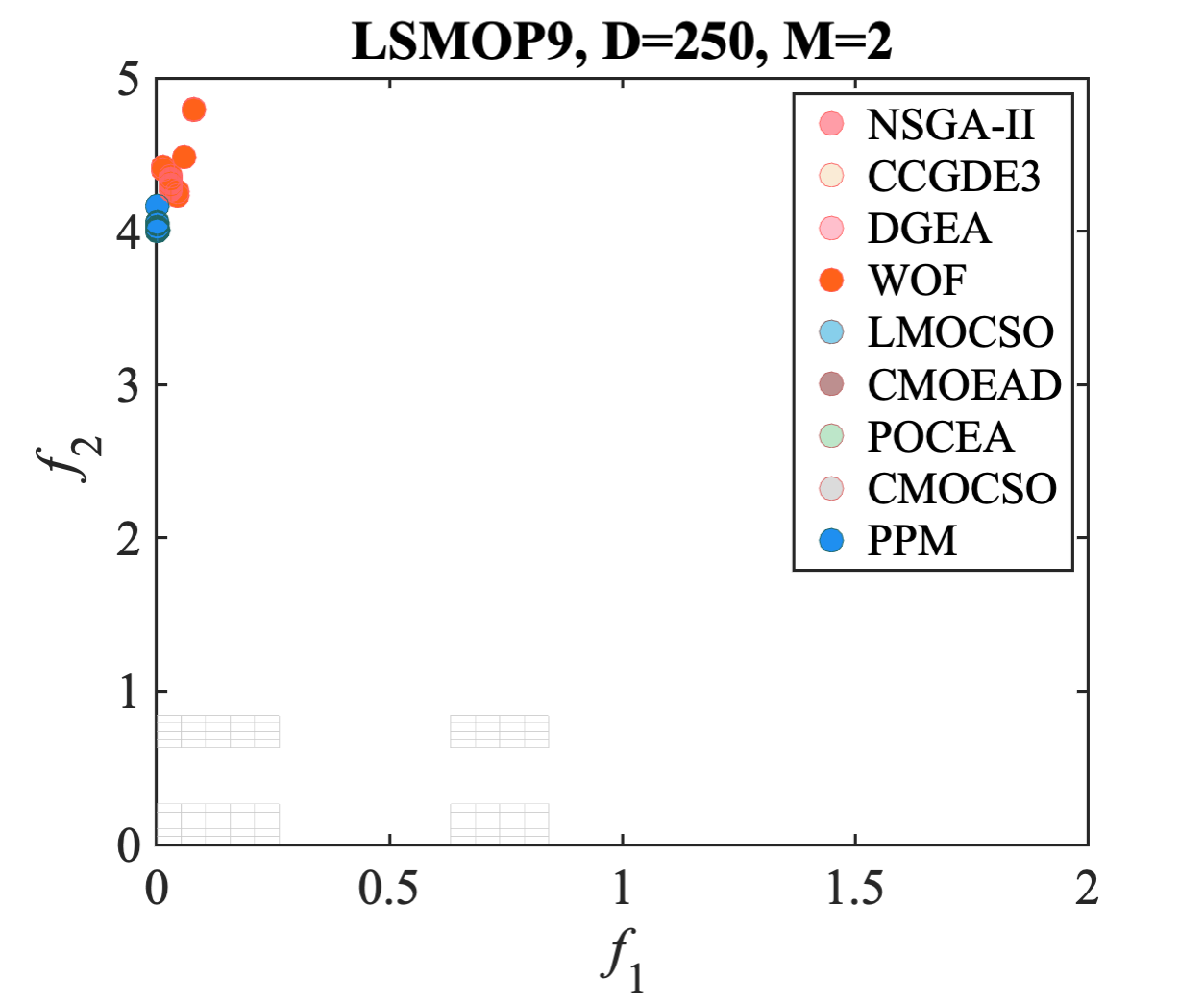}}
    \caption{Visualization of Non-dominated Solutions Obtained by Each Algorithms on LSMOP7, LSMOP8, and LSMOP9. The first row of the Figures is the visualization of the complete solutions, and the second row is the scaled near PF for better visualization.}
    \label{fig: nds}
\end{figure*}
\subsubsection{Visualization of Performance}
Figure~\ref{fig: nds} depicts the non-dominated solutions obtained by different algorithms, with points of different colors, and the PF is located in the lower right corner of the figure. The first row of the figure displays all the solutions obtained by all the algorithms, while the second row provides an enlarged view near the PF to enhance visualization. The visual results align with the quantitative results and indicate that, despite some limitations in diversity, the proposed PPM excels at generating solutions that approximate the PF.

\subsubsection{Experimental Results on Unseen Large-scale MOPs: LSMOP*}
We further evaluated the generalization performance to LSMOP* benchmark~\citep{10254122}, a variant of the original LSMOP benchmark that incorporates a translation transformation. Quantitative and visualization results are presented in Appendix~\ref{app:lsmops} and \ref{app:lsmops2}. The results show that our PPM also achieves the best performance when MOPs are different from the training problems.

\subsection{Performance Generalization to Real-world Applications}
In addition to the pre-trained test problems ZDT and LSMOP, we also evaluate our algorithm on the TREE test problem, which aims to estimate the state of voltage transformers in real time. The TREE test set, with its large-scale nature (3,000 dimensions) and constraints, presents a more complex scenario compared to ZDT and LSMOP. The combination of large-scale decision variables, costly function evaluation, and constraints results in a vast search space. The available evaluations are strictly limited, and the solutions found may not necessarily be feasible, making it challenging for the algorithm to locate the PF or even feasible solutions.
\par
Notably, the results presented in Table~\ref{tab: fe-tree} indicate that our proposed PPM model significantly outperforms compared algorithms on the TREE test set. Despite the triple challenges of large-scale decision variables, expensive function evaluation, and constraints, our algorithm delivers the best results on TREE~1-5. In contrast, some algorithms fail to obtain feasible results. The tests on the TREE test set demonstrate that the proposed PPM can not only achieve impressive optimization results on real-world MOPs but also exhibit potential emergent abilities on some non-pre-trained problems.
\begin{table*}[tbp]
\footnotesize
  \centering
  \caption{IGD Values Obtained By Compared Algorithms on Real World Test Suite,  TREE. The Best Result in Each Row is Highlighted in Bold. The Last Column Represents the Rate of Change of the Proposed PPM Compared to Suboptimal Results.}
    \begin{adjustbox}{width=1\hsize,center}
    \begin{tabular}{cccccccccccc}
    \toprule
    Problem & D     & NSGA-II & CCGDE3 & DGEA  & WOF   & LMOCSO & CMOEAD & POCEA & CMOCSO & PPM   & ROC(\%) \\
    \midrule
    TREE1 & 3000  & NaN+  & NaN+  & 9.58e+01- & 9.65e+01- & NaN+  & NaN+  & NaN+  & NaN+  & \textbf{1.65e+00} & \textbf{98.28\%} \\
    TREE2 & 3000  & NaN+  & NaN+  & 3.69e+02- & 3.99e+02- & NaN+  & NaN+  & NaN+  & NaN+  & \textbf{2.55e+02} & \textbf{30.90\%} \\
    TREE3 & 6000  & NaN+  & NaN+  & 1.20e+02- & 1.20e+02- & NaN+  & NaN+  & NaN+  & NaN+  & \textbf{2.77e+01} & \textbf{76.87\%} \\
    TREE4 & 6000  & NaN+  & NaN+  & 1.20e+02- & 1.19e+02- & NaN+  & NaN+  & NaN+  & NaN+  & \textbf{5.69e+01} & \textbf{52.10\%} \\
    TREE5 & 6000  & NaN+  & NaN+  & 1.84e+02- & NaN+  & NaN+  & NaN+  & NaN+  & NaN+  & \textbf{1.02e+02} & \textbf{44.54\%} \\
    \midrule
    \multicolumn{2}{c}{(+/-/=)} & {0/5/0} & {0/5/0} & {0/5/0} & {1/4/0} & {0/5/0} & {0/5/0} & {0/5/0} & {0/5/0} &       &  \\
    \bottomrule
    \end{tabular}%
    \end{adjustbox}
  \label{tab: fe-tree}%
\end{table*}%
\begin{table}[tbp]
\scriptsize
  \centering
  \caption{Ablation Study: IGD Values Obtained by Compared Algorithms and Untrained PPM.}
  \begin{threeparttable}
    \begin{tabular}{ccc|cc|cc}
    \toprule
    Problem & D     & M     & NSGA-II & +PPM*$^{\dagger}$ & WOF   & +PPM*$^{\dagger}$ \\
    \midrule
    ZDT1  & 150   & 2     & 2.44e+00- & \textbf{8.40e-01} & \textbf{1.88e+00+} & 2.37E+00 \\
    ZDT1  & 550   & 2     & 2.67e+00- & \textbf{8.40e-01} & \textbf{2.54e+00=} & 2.57E+00 \\
    \midrule
    LSMOP1 & 150   & 2     & 7.18e+00- & \textbf{7.07e-01} & 5.26e+00- & \textbf{2.09e+00} \\
    LSMOP1 & 150   & 10    & NaN+  & NaN   & NaN+  & NaN \\
    LSMOP1 & 550   & 2     & 1.05e+01- & \textbf{7.07e-01} & \textbf{2.37e+00+} & 8.53E+00 \\
    LSMOP1 & 550   & 10    & 8.82e+00- & \textbf{1.02e+00} & 8.56e+00= & \textbf{8.53e+00} \\
    \midrule
    LSMOP2 & 150   & 2     & \textbf{1.99e-01+} & 6.39E-01 & \textbf{1.94e-01=} & 1.95E-01 \\
    LSMOP2 & 150   & 10    & NaN+  & NaN   & NaN+  & NaN \\
    LSMOP2 & 550   & 2     & \textbf{6.76e-02+} & 7.07E-01 & 6.77e-02= & \textbf{6.73e-02} \\
    LSMOP2 & 550   & 10    & \textbf{4.56e-01+} & 9.79E-01 & \textbf{4.03e-01+} & 5.01E-01 \\
    \bottomrule
    \end{tabular}%
        \begin{tablenotes}
        \footnotesize
        \item[$\dagger$] +PPM* denotes embedding PPM into the NSGA-II and WOF without pre-training on any benchmarks. The untrained population transformer is used to generate new solutions for NSGA-II and WOF to validate the effectiveness of the pre-training in solving MOPs.
        \vspace{-1em}
    \end{tablenotes}
  \end{threeparttable}
  \label{tab: ablation}%
\end{table}%
\subsection{Ablation Study}
This section examines the efficacy of pre-training. The untrained PPM was integrated into NSGA-II and WOF, replacing the offspring reproduction operators. The outcomes are presented in Table~\ref{tab: ablation}. The experimental findings suggest that the untrained PPM is largely ineffective in resolving complex MOPs. The performance of the untrained PPM in population generation resembles the original reproduction operators of NSGA-II and WOF, which are essentially stochastic generating. This further illustrates the effectiveness of pre-training and fine-tuning in generating better solutions to solve complex MOPs.

\subsection{Convergence and Computational Efficiency}
\begin{figure*}[tbp]
    \centering
    \subfloat[{Problem ZDT6}]{\includegraphics[width=0.23\hsize]{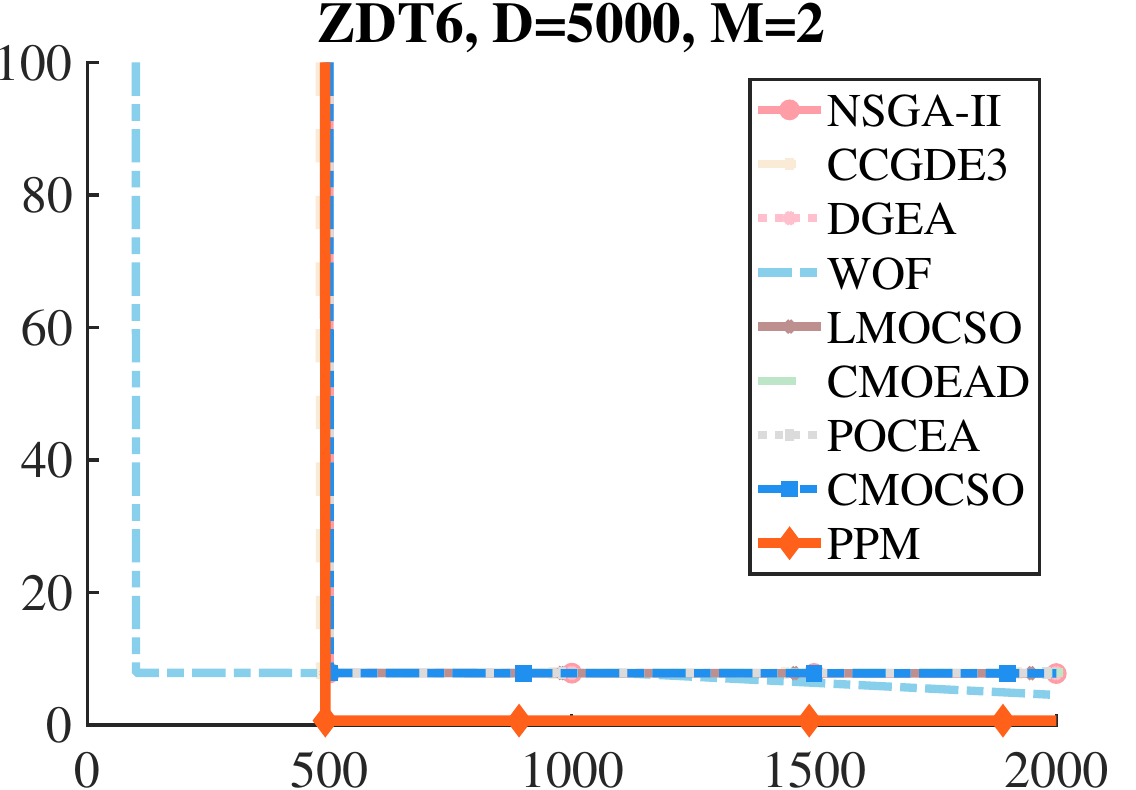}}
    \subfloat[{Problem LSMOP7}]{\includegraphics[width=0.23\hsize]{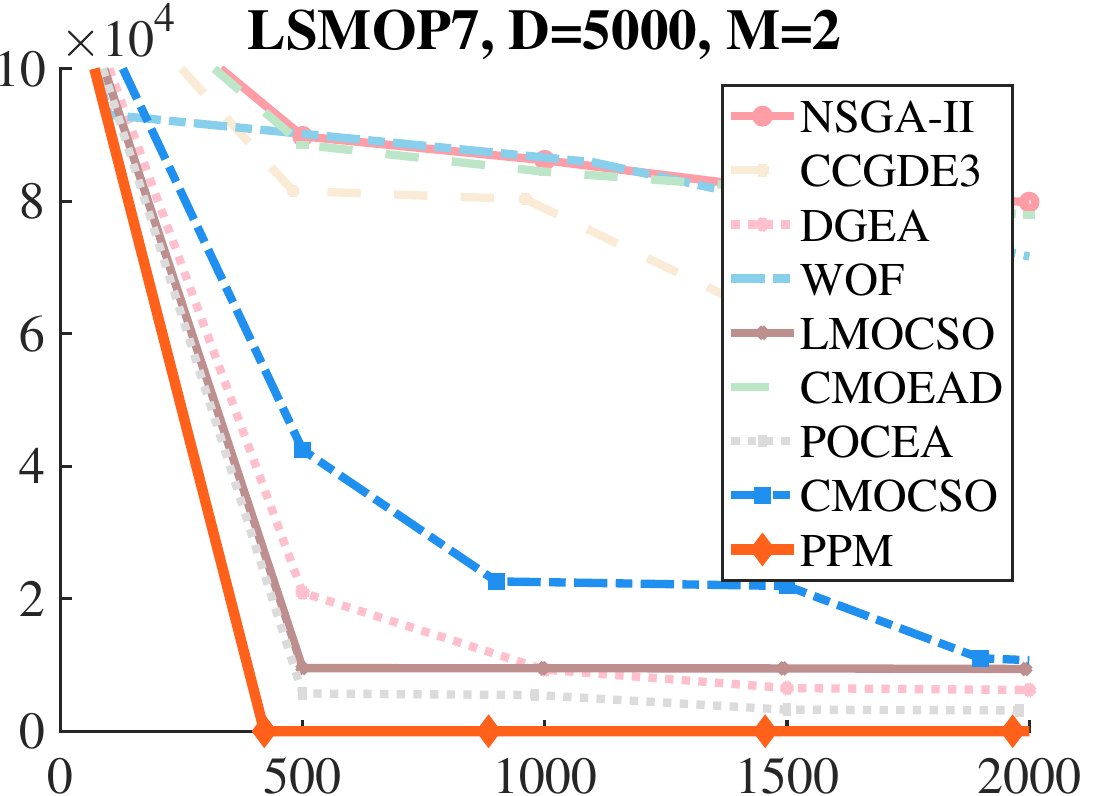}}
    \subfloat[{Problem LSMOP8}]{\includegraphics[width=0.23\hsize]{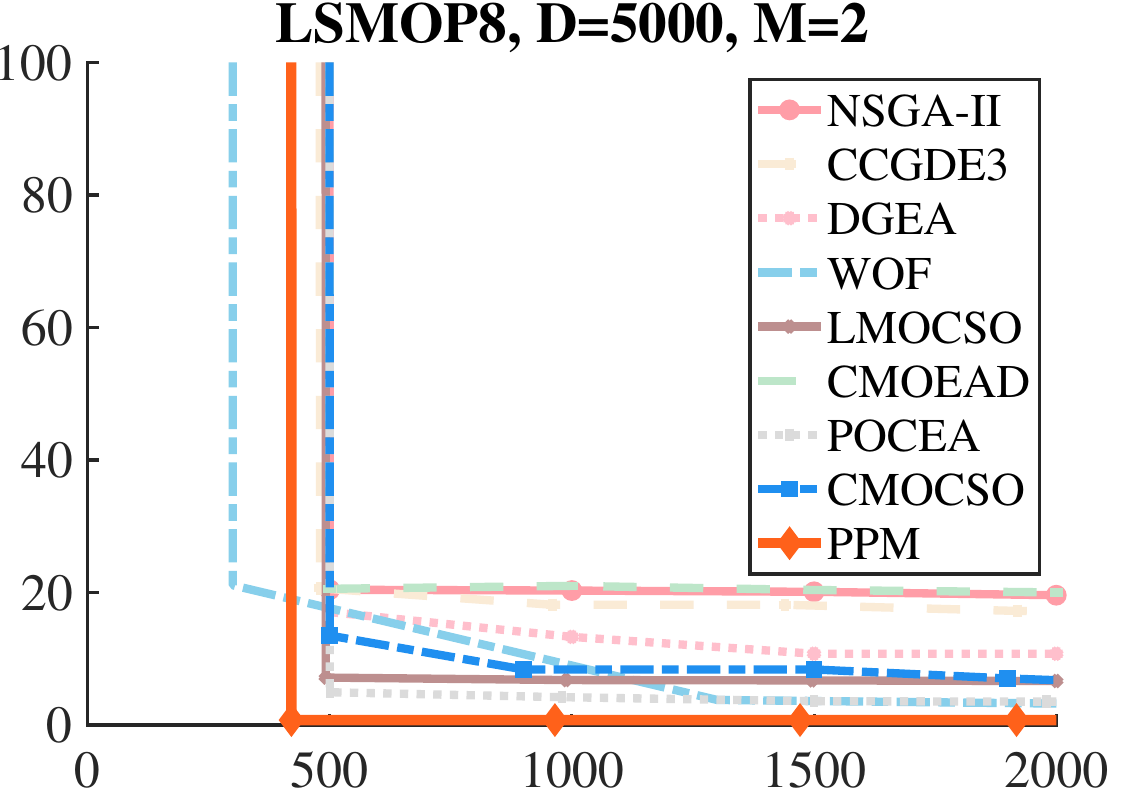}}
    \subfloat[{Problem LSMOP9}]{\includegraphics[width=0.23\hsize]{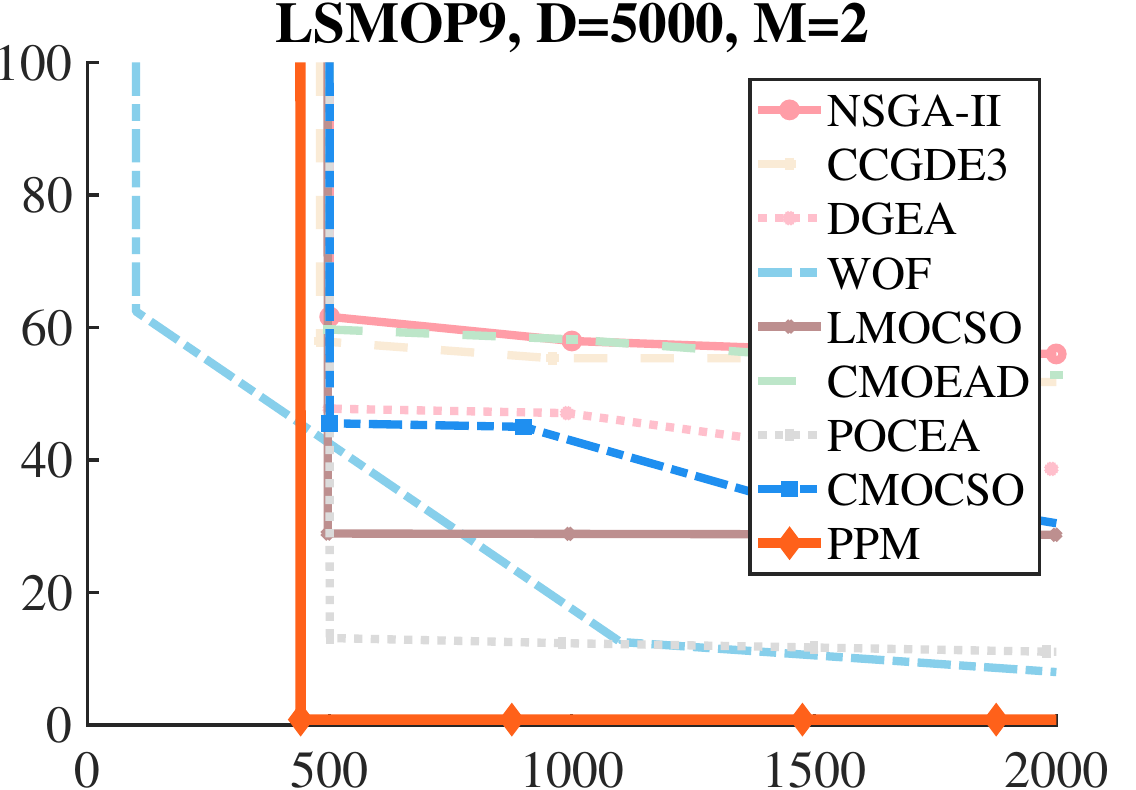}} \\
    \caption{Visualization of Convergence of the Proposed PPM and Compared Algorithms}
    \label{fig: con}
\end{figure*}

\begin{figure*}[tbp]
    \centering
    \subfloat[{Problem ZDT6}]{\includegraphics[width=0.23\hsize]{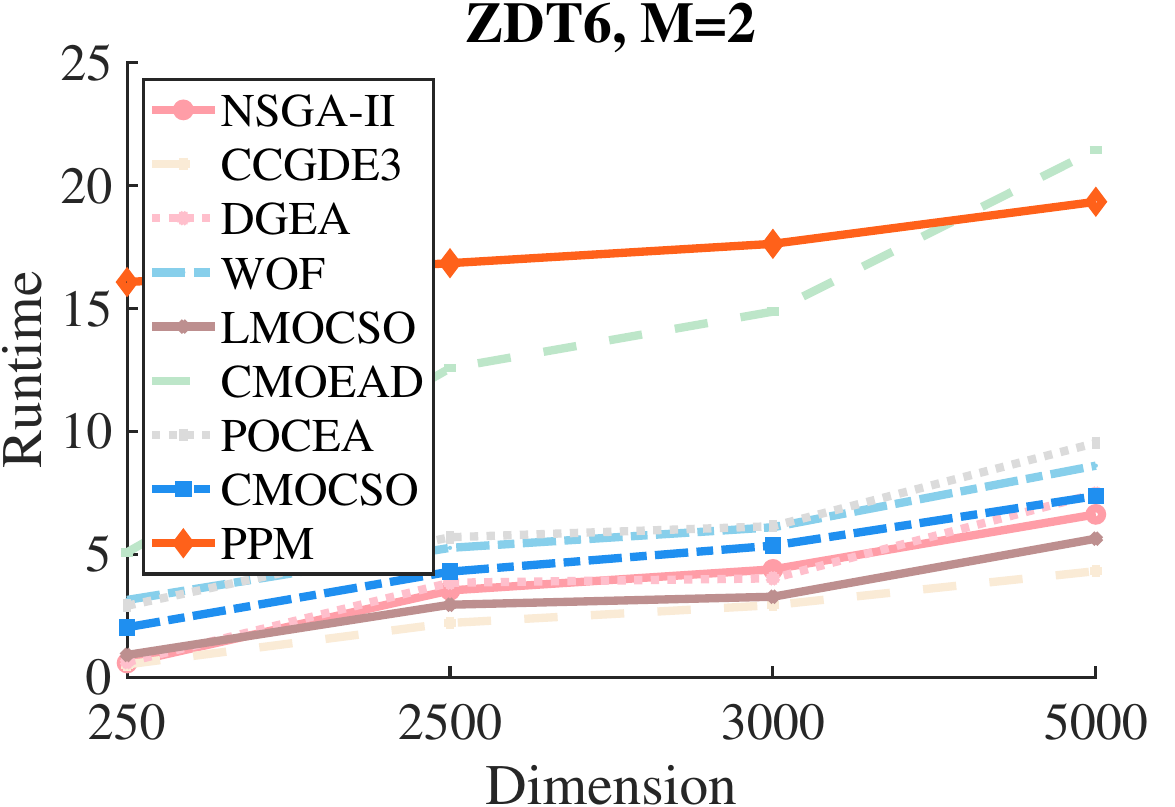}}
    \subfloat[{Problem LSMOP7}]{\includegraphics[width=0.23\hsize]{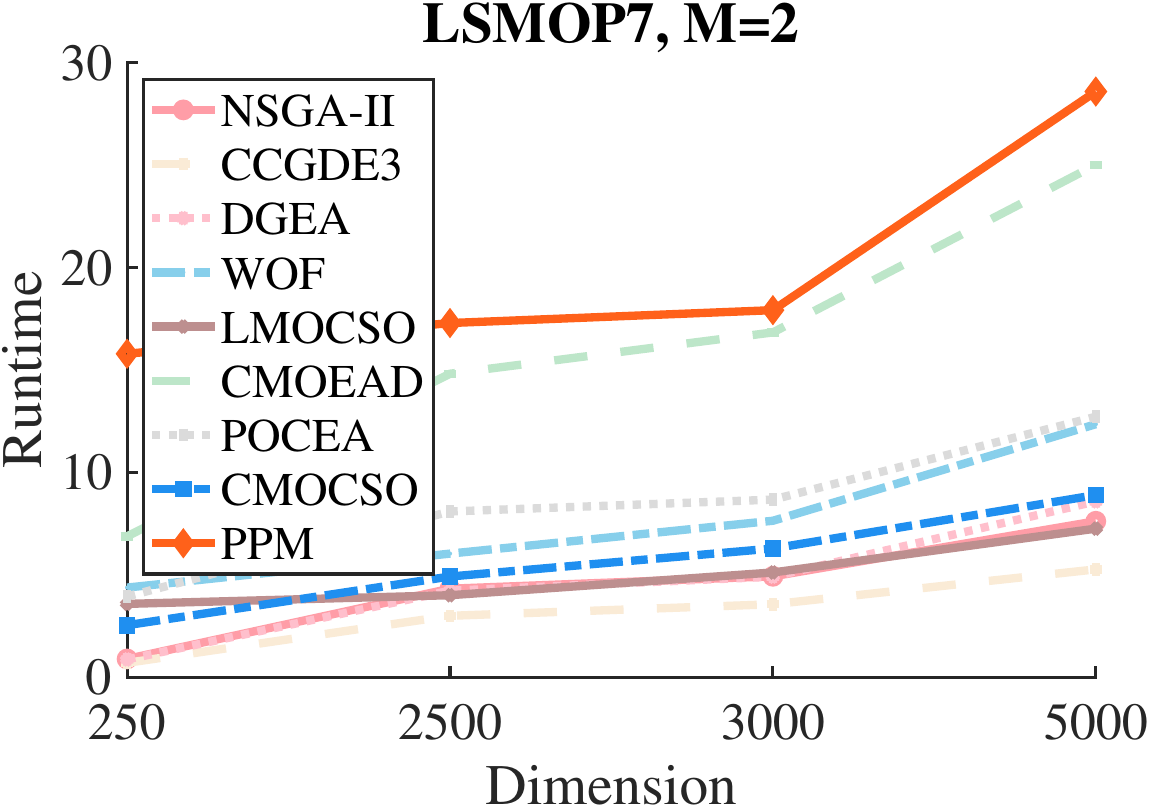}}
    \subfloat[{Problem LSMOP8}]{\includegraphics[width=0.23\hsize]{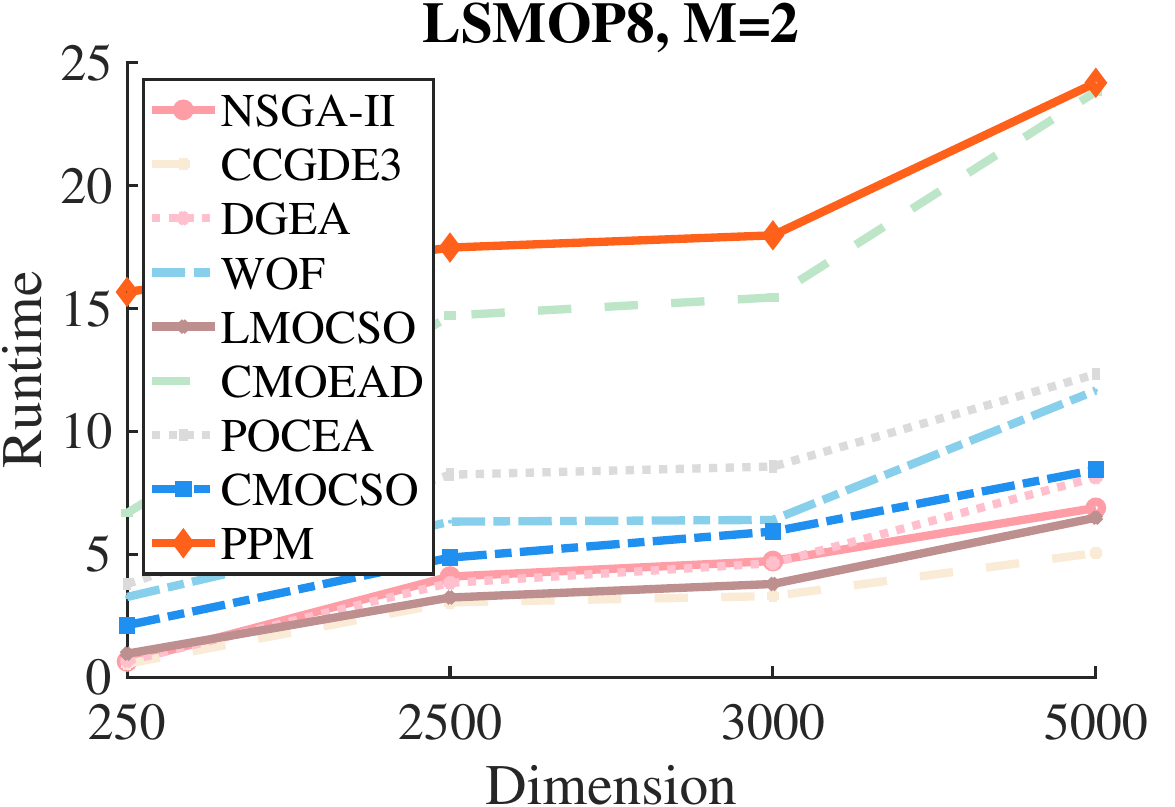}}
    \subfloat[{Problem LSMOP9}]{\includegraphics[width=0.23\hsize]{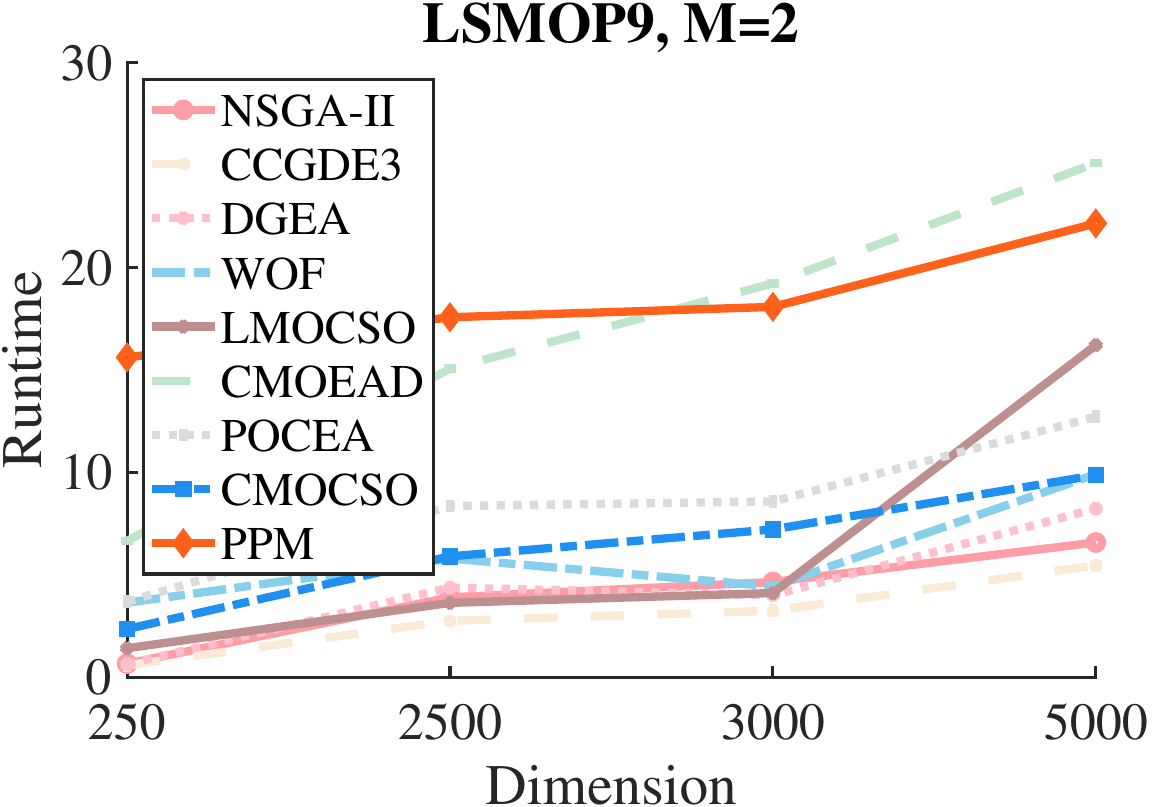}} \\
    \caption{Visualization of the Running Time of the Proposed PPM and Compared Algorithms.}
    \label{fig: time}
\end{figure*}

The progression of IGD values for all compared algorithms on the ZDT6, LSMOP7, and LSMOP8 benchmark problems is illustrated in Figure~\ref{fig: con}. As the figure indicates, the proposed algorithm achieves superior IGD values and exhibits the fastest convergence rate, which is particularly evident when the number of function evaluations reaches 1,000. The comparative algorithms generally fail to match the performance of the proposed algorithm even after 2,000 evaluations. Beyond convergence, computational efficiency is a crucial consideration for practical applications. Figure~\ref{fig: time} presents the average running time of all compared algorithms on ZDT6, LSMOP7, LSMOP8, and LSMOP9, with the number of decision variables ranging from 250 to 5,000.

The convergence analysis, coupled with the assessment of running time, underscores the advantages of our algorithm in addressing complex multi-objective optimization problems. Although the proposed PPM requires an additional pre-training phase, several important considerations justify this aspect. Firstly, during the fine-tuning process, the algorithm's time consumption is comparable to that of most non-surrogate MOEAs, as opposed to ABSAEA and CSEA, which often require over a day to solve a single complex MOP. Secondly, when evaluating the algorithm's quantitative performance in relation to the time spent on calculations, the overall computation time remains acceptable for practical problem-solving purposes. Thirdly, the time complexity of the PPM is not directly related to the dimension, thanks to the designed dimension embedding. Consequently, higher dimensions yield greater advantages. Lastly, the proposed algorithm utilizes a transformer model that can be accelerated by a GPU, opening the possibility of investigating faster models that leverage advanced inference speeds in future research endeavors~\citep{pmlr-v202-leviathan23a}.

\section{Conclusion and Future Work}
\label{sec: con}
Evolution and populations constitute the foundational mechanisms of heuristic MOEAs, where algorithmic efficacy critically depends on population quality. This study demonstrates the potential of population pre-training to learn evolutionary patterns from previously solved optimization problems, resulting in a unified framework capable of generating high-quality solutions for complex MOPs. Specifically, addressing limitations in existing methods concerning the modeling of large-scale populations and the integration of objective space information, we introduce dimension embedding and objective fusion techniques. The proposed Population Pre-trained Model represents an innovative approach integrating learned evolutionary pattern extraction via artificial neural networks into evolutionary computation.

Experimental results confirm the effectiveness of the proposed pre-training paradigm in solving a wide range of complex MOPs, including those characterized by large-scale decision variables, many objectives, constraints, and computationally expensive evaluation functions. The ability of PPM to generalize across diverse problem types, from constrained to large-scale MOPs, constitutes a significant advancement in evolutionary computation. Future research should explore larger-scale models and alternative neural architectures to address the challenges inherent in complex MOPs.

\small

\bibliographystyle{apalike}
\bibliography{pet}

\appendix
\section{Supplementary Material for Experiments}
\label{app:exp}
This is the supplementary material for the paper ``Enhancing Generalization and Scalability for Multi-Objective Optimization with Population Pre-Training'', and this file includes five sections, 3 tables, and 7 figures.
\par
This supplementary material is mainly the appendix of the experiment and is divided into two parts. Section \ref{app:ps} presents parameters in all algorithms. Section \ref{app:lsmop} supplements the visualization results on ZDT6 and LSMOP7-9. Sections \ref{app:lsmops} and \ref{app:lsmops2} present results on benchmark LSMOP*.


\section{Parameter Settings}
\label{app:ps}

The Table \ref{tab: full-para} lists the specific parameters for each compared algorithms.


\begin{table}[htbp]
  \centering
  \caption{Specific parameters for each compared algorithms}
  \begin{adjustbox}{width=1\hsize,left}
    \begin{tabular}{p{5em}p{5em}p{30em}p{10em}}
    \toprule
    Algorithm & Reference & Parameters & Property \\
    \midrule
    NSGA-II & \citep{996017} & The simulated binary crossover and the polynomial mutation are adopted;\newline{}The distribution index of crossover is set to 20; \newline{}The distribution index of mutation is set to 20;\newline{}The crossover probability is set to 1.0. \\
    \midrule
    CCGDE3 & \citep{6557903} & The number of subPopulations is 2;\newline{}The times of each subpopulation is 1. & Large-scale\\
    \midrule
    DGEA & \citep{9138459} & The number of sampling solutions in control variable analysis is 20;\newline{}The maximum number of tries required to judge the interaction is 6. & Large-scale\newline{} Many-objective\\
    \midrule
    WOF   & \citep{RN89} & The number of FEs for the optimization of each original problem t1 is set to 1000;\newline{}For the transferred problem, t2 is set to 500;\newline{}Parameter $q$ is set to $M$ + 1;\newline{}The number of groups $\gamma$ is set to 4;\newline{}The fraction of function evaluations to use for the alternating weight-optimisation phase is 0.5. & Large-scale\\
    \midrule
    LMOCSO & \citep{8681243} & \multicolumn{1}{l}{} & Large-scale \newline{} Many-objective \newline{} Constrained\\
    \midrule
    CMOEAD & \citep{6600851} & \multicolumn{1}{l}{} & Many-objective \newline{} Constrained \\
    \midrule
    CMOCSO & \citep{9861720} & Extension factor $\tau$ is 0.05;\newline{}Parameter to control the speed of reducing relaxation of constraints $cp$ is 2; & Large-scale \newline{} Constrained\\
    \midrule
    POCEA & \citep{9311862} & Parameter $K$ controls the neighborhood size is 5 & Large-scale \newline{} Constrained\\
    \midrule
    AB-SAEA & \citep{WANG2020317} & The parameter controlling the rate of change of penalty $\alpha$ is 2;\newline{} Number of generations before updating Kriging models $w_{max}$ is 20;\newline{}Number of re-evaluated solutions at each generation $\mu$ is 5; & Many-objective \newline{} Expensive\\
    \midrule
    CSEA & \citep{8281523} & Number of reference solutions $k$ is 6;\newline{} Number of solutions evaluated by surrogate model $g_{max}$ is 3000; & Many-objective \newline{} Expensive\\
    \midrule
    EmoDM & \citep{yan2024emodm} &  & Large-scale\\
    \midrule
    MOEA/D-LO & \citep{liu2023large} & LLM: GPT-3.5 Turbo & Many-objective\\
    \midrule
    PPM & Ours  & The ratio of sampling decision variables $f$ is set to 0.2;\newline{}The function evaluation $e$ for decision variable sampling is set to $0.01 \times E$. & Large-scale \newline{} Many-objective \newline{} Constrained \newline{} Expensive\\
    \bottomrule
    \end{tabular}%
    \end{adjustbox}
  \label{tab: full-para}%
\end{table}%

\section{Visualization on Benchmark ZDT6, LSMOP7, LSMOP8, and LSMOP9}
\label{app:lsmop}

\begin{figure*}[htbp]
    \centering
    \subfloat[{Problem ZDT6}]{\includegraphics[width=0.3\hsize]{POF/All_POF2_ZDT6_M2_D250.pdf}}
    \subfloat[{Problem ZDT6}]{\includegraphics[width=0.3\hsize]{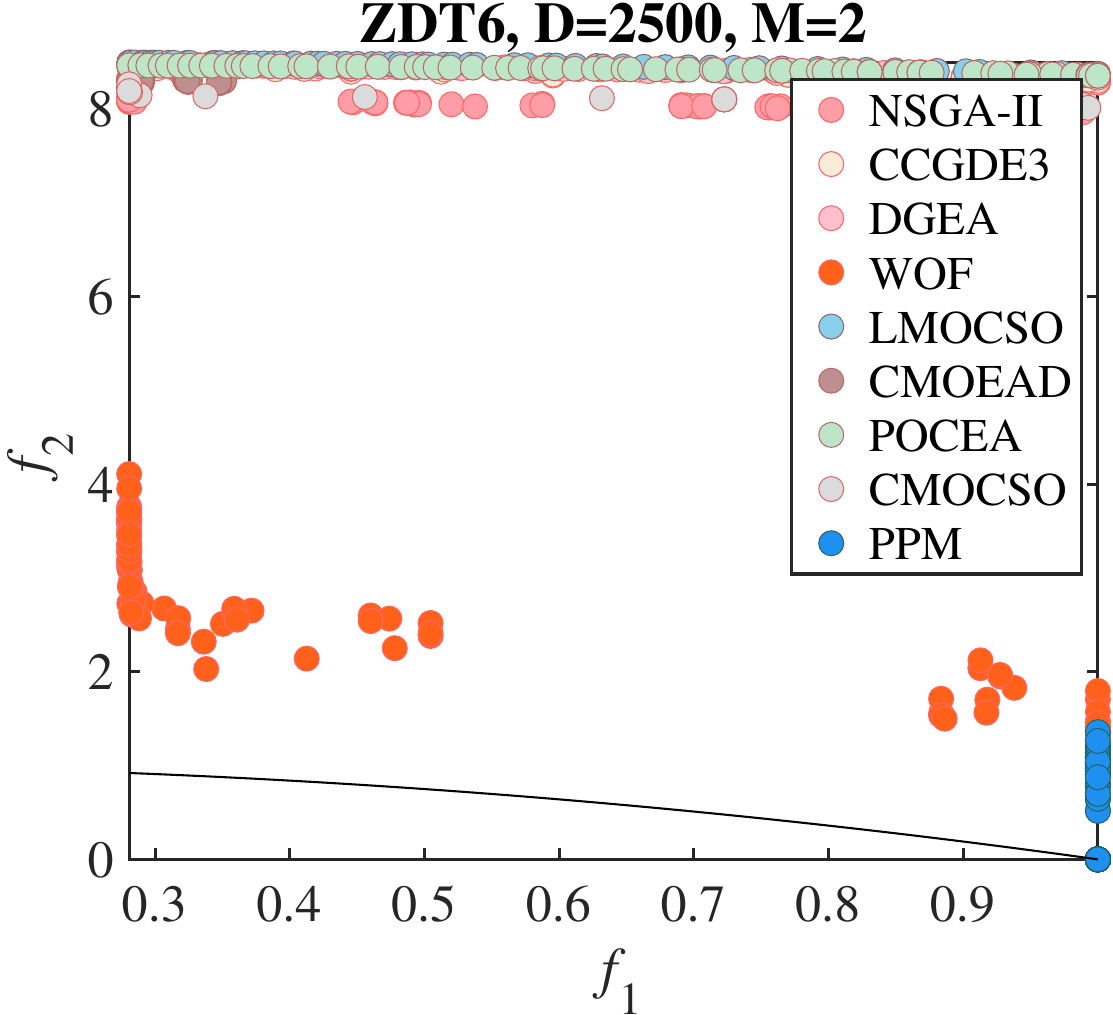}}
    \subfloat[{Problem ZDT6}]{\includegraphics[width=0.3\hsize]{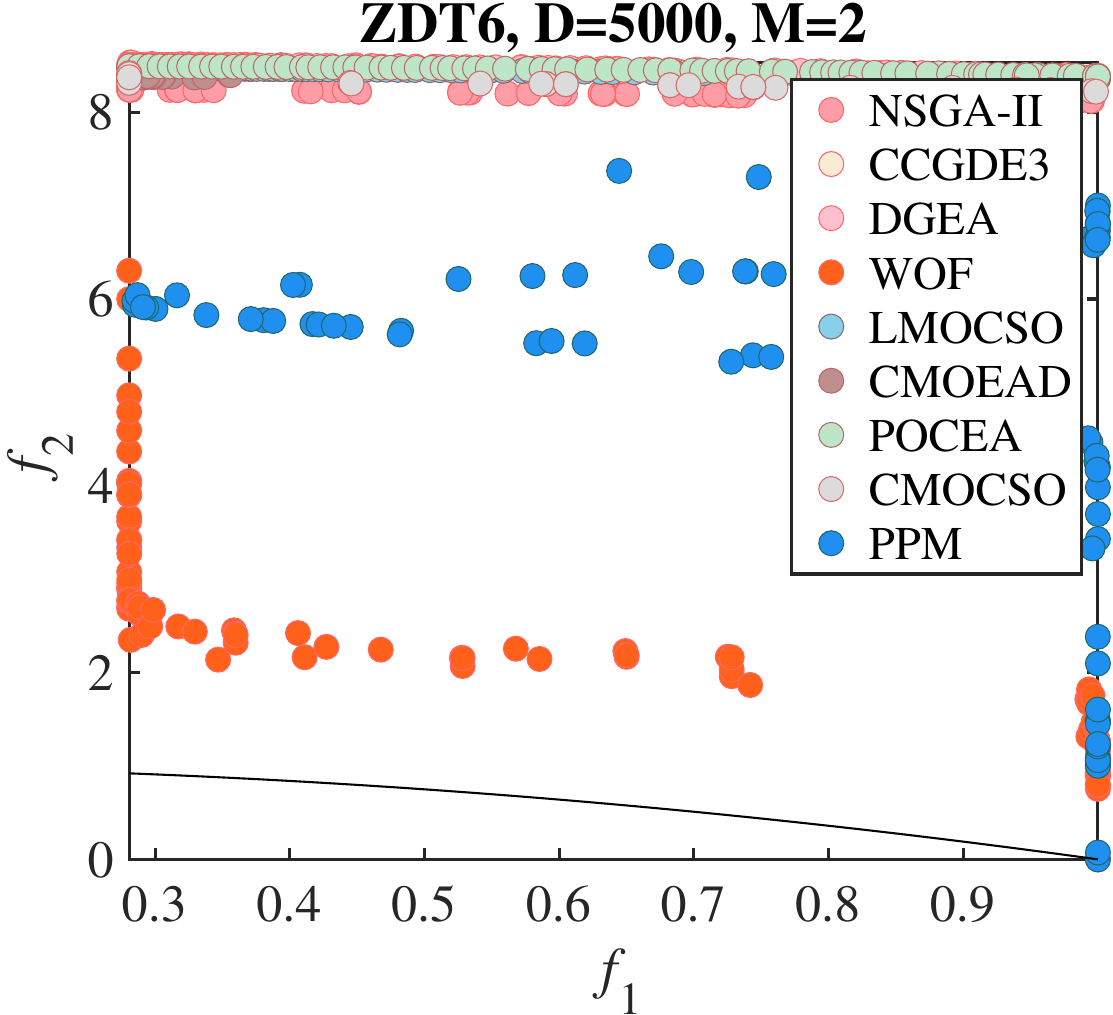}}
    \caption{Visualization of Non-dominated Solutions Obtained by Each Algorithm on Bi-objective ZDT6.}
    \label{fig: zdt}
\end{figure*}

\begin{figure*}[htbp]
    \centering
    \subfloat[{Problem LSMOP7}]{\includegraphics[width=0.3\hsize]{POF/All_POF2_LSMOP7_M2_D250.pdf}}
    \subfloat[{Problem LSMOP7}]{\includegraphics[width=0.3\hsize]{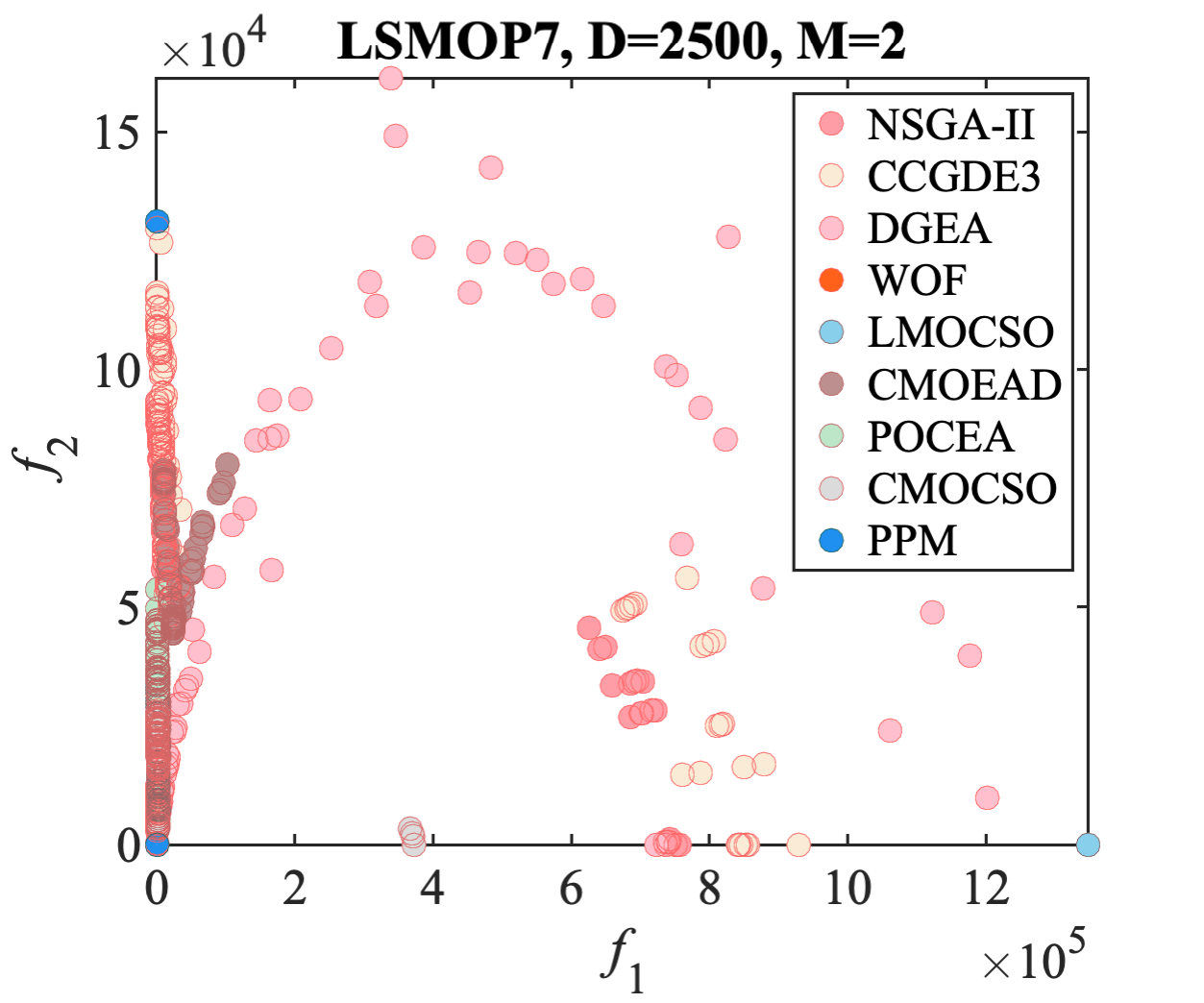}}
    \subfloat[{Problem LSMOP7}]{\includegraphics[width=0.3\hsize]{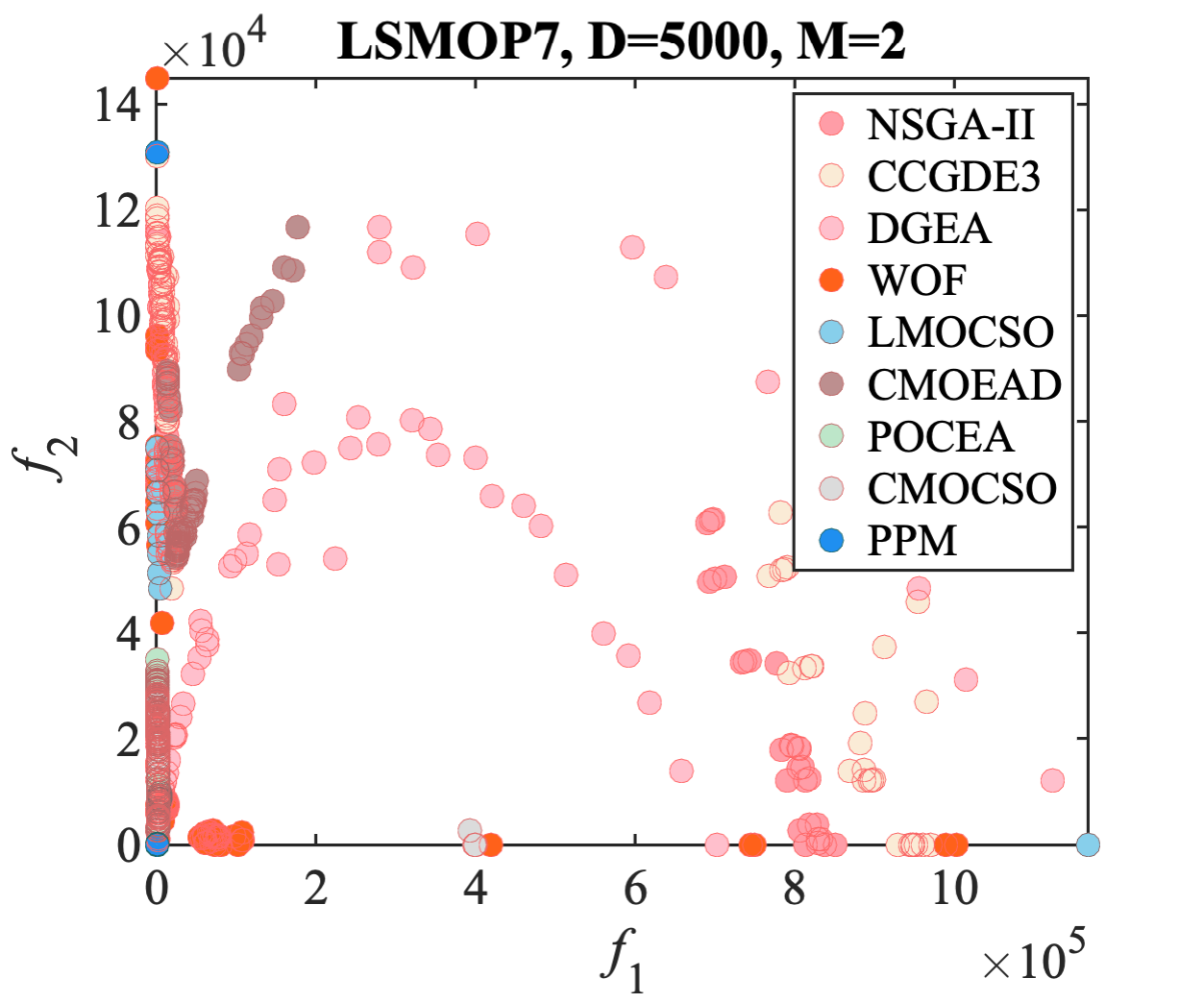}}
    \caption{Visualization of Non-dominated Solutions Obtained by Each Algorithm on Bi-objective LSMOP7.}
    \label{fig: lsmop7}
\end{figure*}

\begin{figure*}[htbp]
    \centering
    \subfloat[{Problem LSMOP8}]{\includegraphics[width=0.3\hsize]{POF/All_POF2_LSMOP8_M2_D250.pdf}}
    \subfloat[{Problem LSMOP8}]{\includegraphics[width=0.3\hsize]{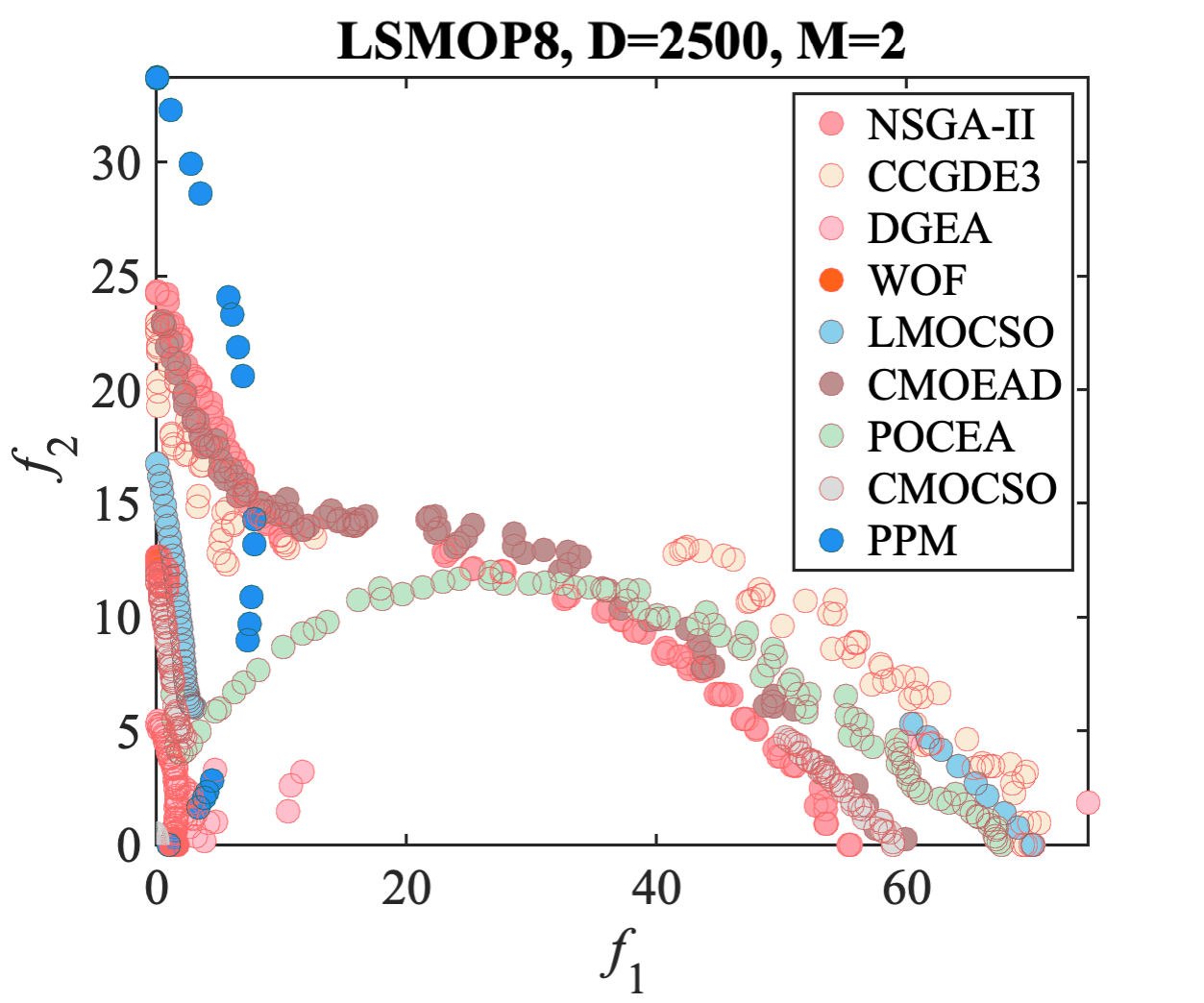}}
    \subfloat[{Problem LSMOP8}]{\includegraphics[width=0.3\hsize]{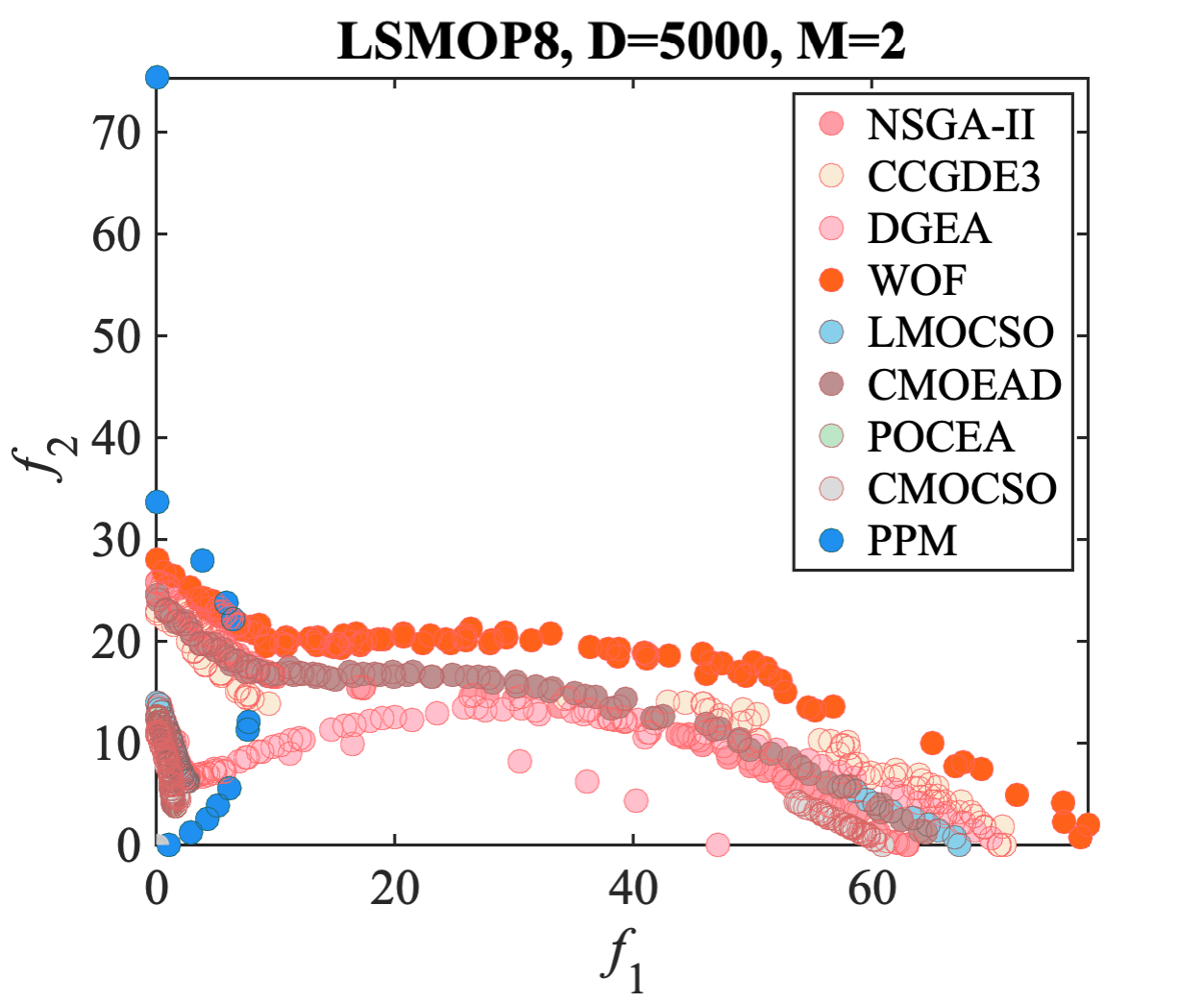}}
    \caption{Visualization of Non-dominated Solutions Obtained by Each Algorithm on Bi-objective LSMOP8.}
    \label{fig: lsmop8}
\end{figure*}

\begin{figure*}[htbp]
    \centering
    \subfloat[{Problem LSMOP9}]{\includegraphics[width=0.3\hsize]{POF/All_POF2_LSMOP9_M2_D250.pdf}}
    \subfloat[{Problem LSMOP9}]{\includegraphics[width=0.3\hsize]{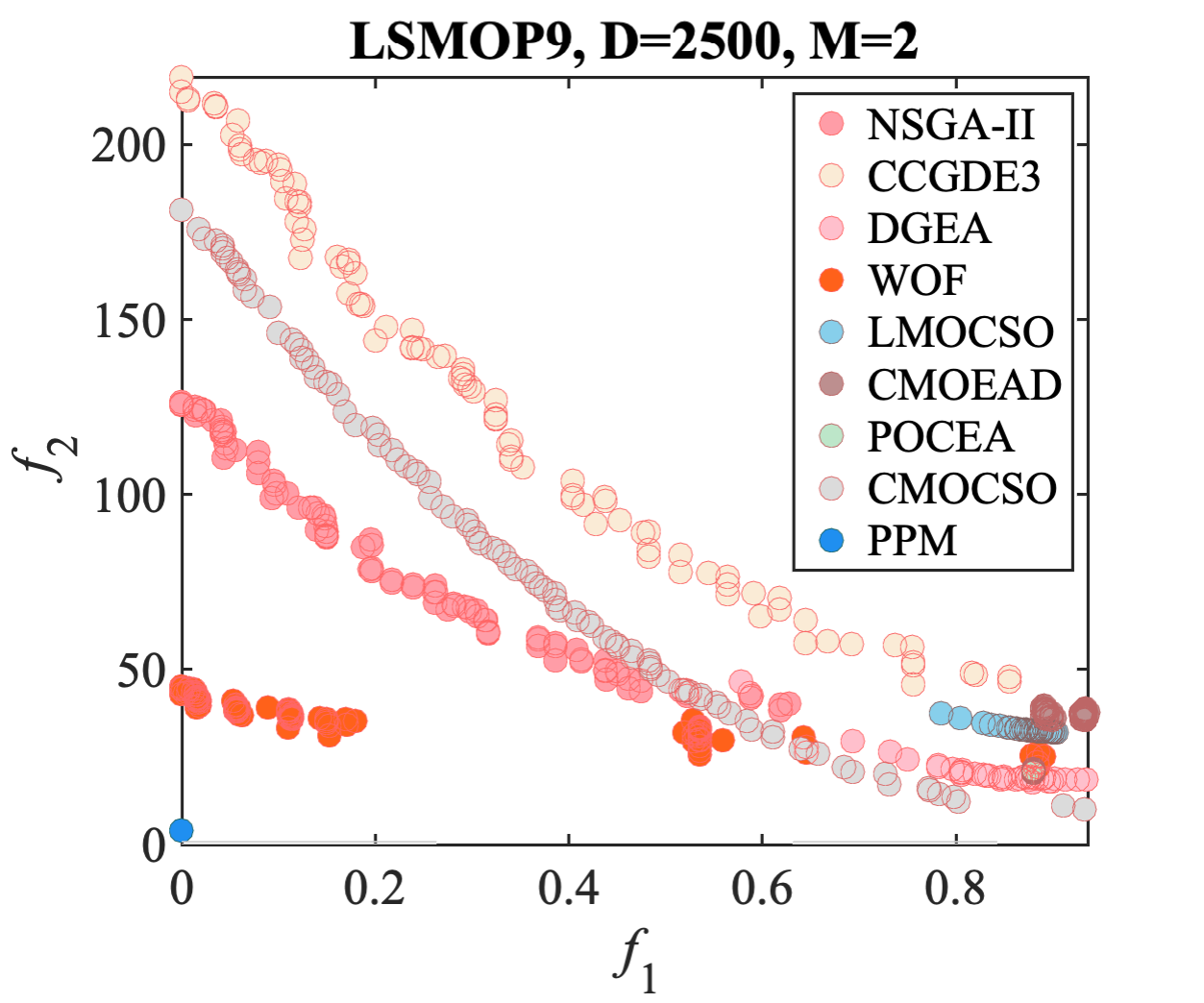}}
    \subfloat[{Problem LSMOP9}]{\includegraphics[width=0.3\hsize]{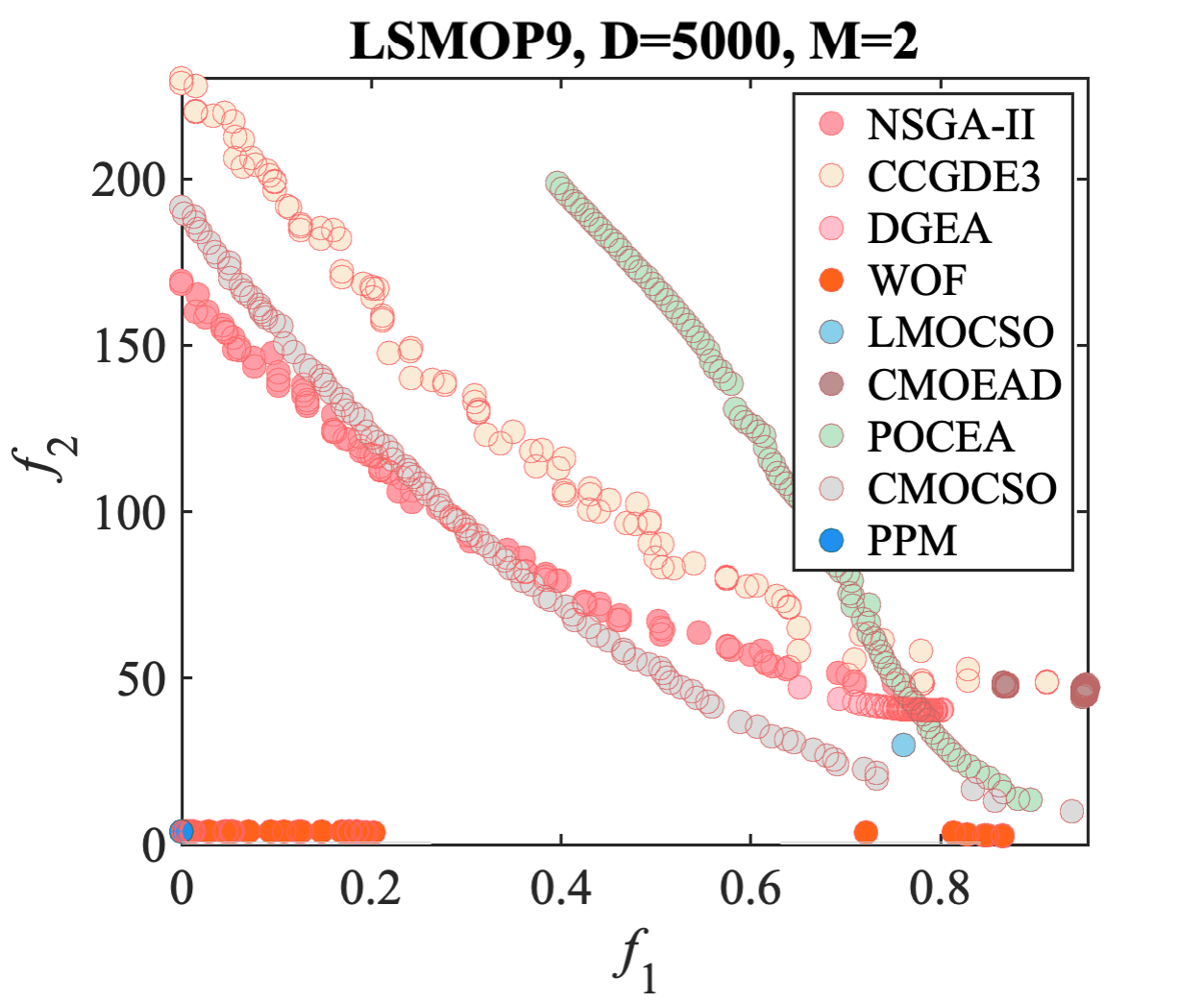}}
    \caption{Visualization of Non-dominated Solutions Obtained by Each Algorithm on Bi-objective LSMOP9.}
    \label{fig: lsmop9}
\end{figure*}

We visualize the nondominated solution sets obtained by all algorithms on LSMOP in Figures \ref{fig: zdt}, \ref{fig: lsmop7}, \ref{fig: lsmop8}, and \ref{fig: lsmop9}.

\clearpage
\section{Test on Benchmark LSMOP*}
\label{app:lsmops}
We further evaluated the performance of our proposed algorithm using the LSMOP* benchmark~\citep{10254122}, a variant of the original LSMOP benchmark that incorporates a translation transformation.
\par
Solutions of original LSMOP1-LSMOP9 near the boundary points $\mathbf{o}$ or $\mathbf{t}$ are proximate to or part of the PS, therefore most large-scale MOEAs can readily locate solutions near $\mathbf{o}$ or $\mathbf{t}$ by exploring specified directions and consistently achieving satisfactory results. The modified benchmark LSMOP* introduces a translation transformation to circumvent these 'shortcuts'.
\par
We present the experimental results on bi-objective and tri-objective LSMOP* in Tables~\ref{tab: lsmops-m2} and \ref{tab: lsmops-m3}, respectively. Following previous experiments, the proposed PPM is only pre-evolved on the original LSMOP1-6 and ZDT1-5. Under this modification to the benchmark, the proposed algorithm still finds better solutions on more problems (8 and 9 best results, respectively).

\begin{table}[htbp]
  \centering
  \caption{IGD Values Obtained By Compared Algorithms on 27 Bi-objective Instances From LSMOP* Test Suite. The Best Result in Each Row is Highlighted in Bold.}
    \begin{adjustbox}{width=1\hsize,left}
    \begin{tabular}{ccccccccccc}
    \toprule
    Problem & D     & NSGI-II & CCGDE3 & DGEA  & WOF   & LMOCSO & CMOEAD & POCEA & CMOCSO & PPM \\
    \midrule
    \multirow{3}[2]{*}{LSMOPS1} & 1000  & 6.05e+01- & 6.23e+01- & 5.60e+01- & 5.47e+01- & 5.66e+01- & 6.22e+01- & 5.43e+01- & 5.67e+01- & \textbf{4.50e+01} \\
          & 2000  & 6.20e+01- & 6.39e+01- & 5.78e+01- & 5.50e+01- & 5.67e+01- & 6.56e+01- & 5.57e+01- & 5.66e+01- & \textbf{4.51e+01} \\
          & 5000  & 6.49e+01- & 6.46e+01- & \textbf{4.08e+01+} & 6.46e+01- & 5.72e+01= & 6.68e+01- & 5.61e+01+ & 5.74e+01- & 5.72E+01 \\
    \midrule
    \multirow{3}[2]{*}{LSMOPS2} & 1000  & 6.28e-02= & 6.31e-02= & 6.20e-02= & \textbf{5.53e-02=} & 6.08e-02= & 6.68e-02= & 5.85e-02= & 6.09e-02= & 6.43E-02 \\
          & 2000  & 3.64e-02= & 3.66e-02= & 3.62e-02= & 3.71e-02= & \textbf{3.49e-02=} & 4.56e-02= & 3.52e-02= & 3.55e-02= & 3.72E-02 \\
          & 5000  & 2.01e-02= & 2.03e-02= & 2.00e-02= & 2.02e-02= & \textbf{1.95e-02=} & 2.98e-02= & 1.98e-02= & 1.98e-02= & 2.24E-02 \\
    \midrule
    \multirow{3}[2]{*}{LSMOPS3} & 1000  & 1.39e+02- & 1.41e+02- & 1.49e+02- & 1.37e+02- & 7.72e+03- & 1.38e+02- & 1.49e+02- & 1.45e+02- & \textbf{5.71e+01} \\
          & 2000  & 1.43e+02- & 1.45e+02- & \textbf{5.71e+01=} & 1.41e+02- & 1.04e+05- & 1.47e+02- & 1.54e+02- & 1.45e+02- & 5.71E+01 \\
          & 5000  & 1.48e+02- & 1.50e+02- & \textbf{5.71e+01=} & 6.48e+04- & 3.57e+04- & 1.53e+02- & 1.60e+02- & 1.47e+02- & 5.71E+01 \\
    \midrule
    \multirow{3}[2]{*}{LSMOPS4} & 1000  & 1.11e-01= & 1.11e-01= & 1.11e-01= & 1.12e-01= & 1.10e-01= & 1.17e-01= & \textbf{1.10e-01=} & 1.10e-01= & 1.12E-01 \\
          & 2000  & 6.51e-02= & 6.51e-02= & 6.50e-02= & 6.54e-02= & 6.42e-02= & 7.00e-02= & \textbf{6.41e-02=} & 6.44e-02= & 6.65E-02 \\
          & 5000  & 3.32e-02= & 3.33e-02= & 3.29e-02= & 3.37e-02= & \textbf{3.28e-02=} & 4.55e-02= & 3.30e-02= & 3.29e-02= & 3.45E-02 \\
    \midrule
    \multirow{3}[2]{*}{LSMOPS5} & 1000  & 1.46e+02- & 1.59e+02- & 1.53e+02- & 1.52e+02- & 1.49e+02- & 1.57e+02- & 1.46e+02- & 1.48e+02- & \textbf{9.42e+01} \\
          & 2000  & 1.54e+02- & 1.59e+02- & 1.51e+02- & 1.55e+02- & 1.50e+02- & 1.59e+02- & 1.48e+02- & 1.48e+02- & \textbf{6.67e+01} \\
          & 5000  & 1.58e+02- & 1.62e+02- & \textbf{6.67e+01+} & 1.47e+02+ & 1.51e+02+ & 1.66e+02- & 1.46e+02+ & 1.48e+02+ & 1.52E+02 \\
    \midrule
    \multirow{3}[2]{*}{LSMOPS6} & 1000  & 8.38e-01- & 8.41e-01- & 8.38e-01- & 8.25e-01= & 3.84e+04- & 1.04e+05- & 8.44e-01- & 8.38e-01- & \textbf{7.76e-01} \\
          & 2000  & 7.86e-01= & 7.86e-01= & 7.87e-01= & 7.80e-01= & 1.17e+04- & 7.86e-01= & 7.86e-01= & 7.85e-01= & \textbf{7.57e-01} \\
          & 5000  & 7.58e-01= & 7.58e-01= & \textbf{7.48e-01=} & 2.10e+05- & 1.91e+05- & 7.58e-01= & 7.58e-01= & 7.58e-01= & \textbf{7.48e-01} \\
    \midrule
    \multirow{3}[2]{*}{LSMOPS7} & 1000  & 9.44e+05- & 1.01e+06- & 1.10e+06- & \textbf{4.85e+05+} & 6.00e+05- & 9.76e+05- & 5.61e+05- & 6.26e+05- & 5.23E+05 \\
          & 2000  & 9.96e+05- & 9.14e+05- & 6.76e+05- & 9.18e+05- & \textbf{5.70e+05+} & 1.02e+06- & 6.16e+05- & 6.97e+05- & 5.73E+05 \\
          & 5000  & 1.08e+06- & 9.53e+05- & 1.06e+06- & 1.02e+06- & 6.22e+05+ & 1.11e+06- & \textbf{4.97e+05+} & 6.44e+05+ & 7.12E+05 \\
    \midrule
    \multirow{3}[2]{*}{LSMOPS8} & 1000  & 6.76e+01+ & 7.06e+01+ & 6.97e+01+ & 6.91e+01+ & 6.49e+01+ & 7.21e+01+ & \textbf{5.97e+01+} & 6.42e+01+ & 7.67E+01 \\
          & 2000  & 6.92e+01+ & 7.21e+01+ & 7.33e+01+ & 7.03e+01+ & 6.61e+01+ & 7.33e+01+ & \textbf{6.21e+01+} & 6.22e+01+ & 7.55E+01 \\
          & 5000  & 7.38e+01+ & 7.34e+01+ & 7.65e+01- & 7.42e+01+ & 6.28e+01+ & 7.60e+01- & \textbf{6.07e+01+} & 6.57e+01+ & 7.57E+01 \\
    \midrule
    \multirow{3}[2]{*}{LSMOPS9} & 1000  & 1.91e+02- & 2.16e+02- & 1.90e+02+ & 1.97e+02- & 1.93e+02- & 1.95e+02- & 1.90e+02+ & \textbf{1.89e+02+} & 1.91E+02 \\
          & 2000  & 2.15e+02- & 2.28e+02- & \textbf{1.93e+02+} & 2.11e+02- & 2.12e+02- & 2.14e+02- & 1.99e+02- & 2.02e+02- & 1.93E+02 \\
          & 5000  & 2.27e+02- & 2.41e+02- & \textbf{1.93e+02+} & 2.03e+02- & 2.24e+02- & 2.32e+02- & 2.00e+02- & 2.09e+02- & 1.93E+02 \\
    \midrule
    \multicolumn{2}{c}{(+/-/=)} & {3/16/8} & {3/16/8} & {7/10/10} & {5/14/8} & {6/14/7} & {2/17/8} & {7/12/8} & {6/13/8} &  \\
    \bottomrule
    \end{tabular}%
    \end{adjustbox}
  \label{tab: lsmops-m2}%
\end{table}%

\begin{table}[htbp]
  \centering
  \caption{IGD Values Obtained By Compared Algorithms on 27 Tri-objective Instances From LSMOP* Test Suite. The Best Result in Each Row is Highlighted in Bold.}
  \begin{adjustbox}{width=1\hsize,left}
    \begin{tabular}{ccccccccccc}
    \toprule
    Problem & D     & NSGI-II & CCGDE3 & DGEA  & WOF   & LMOCSO & CMOEAD & POCEA & CMOCSO & PPM \\
    \midrule
    \multirow{3}[2]{*}{LSMOPS1} & 1000  & 6.67e+01- & 7.22e+01- & 6.16e+01- & 6.50e+01- & 6.02e+01- & 6.55e+01- & 5.73e+01- & 6.33e+01- & \textbf{4.73e+01} \\
          & 2000  & 6.73e+01- & 7.02e+01- & 6.63e+01- & 6.24e+01- & 6.02e+01- & 6.48e+01- & 6.01e+01- & 6.42e+01- & \textbf{4.77e+01} \\
          & 5000  & 6.94e+01- & 7.12e+01- & 5.14e+01- & 7.02e+01- & 6.23e+01- & 6.89e+01- & 6.07e+01- & 6.45e+01- & \textbf{4.73e+01} \\
    \midrule
    \multirow{3}[2]{*}{LSMOPS2} & 1000  & 8.91e-02= & 8.13e-02= & 8.68e-02= & 8.13e-02= & 7.72e-02= & 8.26e-02= & 8.02e-02= & \textbf{7.70e-02=} & 8.93E-02 \\
          & 2000  & 6.90e-02= & 6.92e-02= & 7.05e-02= & 6.31e-02= & 6.03e-02= & 8.04e-02= & 6.36e-02= & \textbf{5.82e-02=} & 7.55E-02 \\
          & 5000  & 6.07e-02= & 6.10e-02= & 6.04e-02= & 4.97e-02= & 5.02e-02= & 6.29e-02= & 5.63e-02= & \textbf{4.85e-02=} & 6.78E-02 \\
    \midrule
    \multirow{3}[2]{*}{LSMOPS3} & 1000  & 1.24e+02- & 9.07e+01- & \textbf{3.94e+01+} & 1.01e+02- & 8.46e+02- & 8.54e+01- & 1.27e+02- & 1.49e+02- & 5.02E+01 \\
          & 2000  & 8.55e+01- & 9.95e+01- & 5.64e+01= & 8.29e+01- & 1.78e+03- & 8.16e+01- & 1.19e+02- & 1.56e+02- & \textbf{5.64e+01} \\
          & 5000  & 1.16e+02- & 8.46e+01- & \textbf{4.78e+01+} & 9.27e+01- & 9.83e+01- & 8.40e+01- & 9.87e+01- & 1.53e+02- & 5.63E+01 \\
    \midrule
    \multirow{3}[2]{*}{LSMOPS4} & 1000  & 1.71e-01= & 1.67e-01= & 1.71e-01= & 1.62e-01= & \textbf{1.61e-01=} & 1.71e-01= & 1.61e-01= & 1.63e-01= & 1.76E-01 \\
          & 2000  & 1.05e-01= & 1.06e-01= & 1.13e-01= & 1.08e-01= & 9.78e-02= & 1.09e-01= & 1.02e-01= & \textbf{9.74e-02=} & 1.08E-01 \\
          & 5000  & 7.69e-02= & 6.99e-02= & 6.68e-02= & 6.04e-02= & 6.23e-02= & 8.42e-02= & 6.34e-02= & \textbf{6.02e-02=} & 7.58E-02 \\
    \midrule
    \multirow{3}[2]{*}{LSMOPS5} & 1000  & 1.28e+02- & 1.34e+02- & 1.32e+02- & 1.24e+02- & 1.30e+02- & 1.28e+02- & 1.24e+02- & 1.26e+02- & \textbf{5.38e+01} \\
          & 2000  & 1.35e+02- & 1.32e+02- & 5.38e+01= & 1.29e+02- & 1.32e+02- & 1.33e+02- & 1.23e+02- & 1.30e+02- & \textbf{5.38e+01} \\
          & 5000  & 1.36e+02- & 1.34e+02- & 1.33e+02- & 1.38e+02- & 1.27e+02- & 1.36e+02- & 1.27e+02- & 1.31e+02- & \textbf{5.37e+01} \\
    \midrule
    \multirow{3}[2]{*}{LSMOPS6} & 1000  & 3.67e+05- & 4.23e+05- & 3.50e+05- & 3.19e+05+ & 3.79e+05- & 3.55e+05- & \textbf{2.65e+05+} & 3.61e+05- & 3.22E+05 \\
          & 2000  & 4.23e+05- & 4.48e+05- & 3.75e+05- & 3.70e+05- & 3.04e+05- & 4.24e+05- & \textbf{2.26e+05+} & 4.34e+05- & 2.97E+05 \\
          & 5000  & 4.50e+05+ & 4.93e+05+ & 5.75e+05+ & 4.72e+05+ & 4.31e+05+ & 4.26e+05+ & \textbf{2.58e+05+} & 3.86e+05+ & 6.04E+05 \\
    \midrule
    \multirow{3}[2]{*}{LSMOPS7} & 1000  & 1.17e+00= & 1.17e+00= & 1.13e+00= & 1.17e+00= & 1.17e+00= & 1.17e+00= & \textbf{1.08e+00=} & 1.17e+00= & 1.13E+00 \\
          & 2000  & 1.05e+00= & 1.05e+00= & 1.03e+00= & 1.05e+00= & 1.05e+00= & 1.05e+00= & 1.05e+00= & 1.04e+00= & \textbf{1.03e+00} \\
          & 5000  & 9.83e-01= & 9.83e-01= & 9.76e-01= & 9.82e-01= & 9.82e-01= & 9.83e-01= & 9.82e-01= & 9.83e-01= & \textbf{9.76e-01} \\
    \midrule
    \multirow{3}[1]{*}{LSMOPS8} & 1000  & 9.75e-01= & 9.75e-01= & 9.61e-01= & 9.72e-01= & 9.69e-01= & \textbf{7.29e-01+} & 9.72e-01= & 9.76e-01= & 9.61E-01 \\
          & 2000  & 9.70e-01= & 9.71e-01= & 8.01e-01+ & \textbf{6.30e-01+} & 8.19e-01+ & 7.43e-01+ & 7.97e-01+ & 9.72e-01= & 9.57E-01 \\
          & 5000  & 9.67e-01= & 9.66e-01= & 8.91e-01+ & 1.19e+00- & 8.02e-01+ & 6.94e+00- & \textbf{6.94e-01+} & 9.70e-01= & 9.54E-01 \\
    \midrule
    \multirow{3}[1]{*}{LSMOPS9} & 1000  & 4.21e+02+ & 4.37e+02+ & 4.26e+02+ & 4.14e+02+ & 4.32e+02+ & 4.16e+02+ & \textbf{4.10e+02+} & 4.48e+02- & 4.43E+02 \\
          & 2000  & 4.70e+02- & 4.97e+02- & 4.31e+02+ & 4.62e+02- & 4.71e+02- & 4.61e+02- & 4.27e+02+ & \textbf{4.27e+02+} & 4.55E+02 \\
          & 5000  & 4.85e+02- & 5.08e+02- & 4.25e+02+ & \textbf{4.19e+02+} & 4.68e+02- & 4.82e+02- & 4.29e+02+ & 4.36e+02+ & 4.40E+02 \\
    \midrule
    \multicolumn{2}{c}{(+/-/=)} & {2/13/12} & {2/13/12} & {8/7/12} & {5/12/10} & {4/13/10} & {4/14/9} & {8/9/10} & {3/12/12} &  \\
    \bottomrule
    \end{tabular}%
    \end{adjustbox}
  \label{tab: lsmops-m3}%
\end{table}%
\clearpage

\section{Visualization on Benchmark LSMOP*}
\label{app:lsmops2}

\begin{figure*}[htbp]
    \centering
    \subfloat[{Problem LSMOP1}]{\includegraphics[width=0.3\hsize]{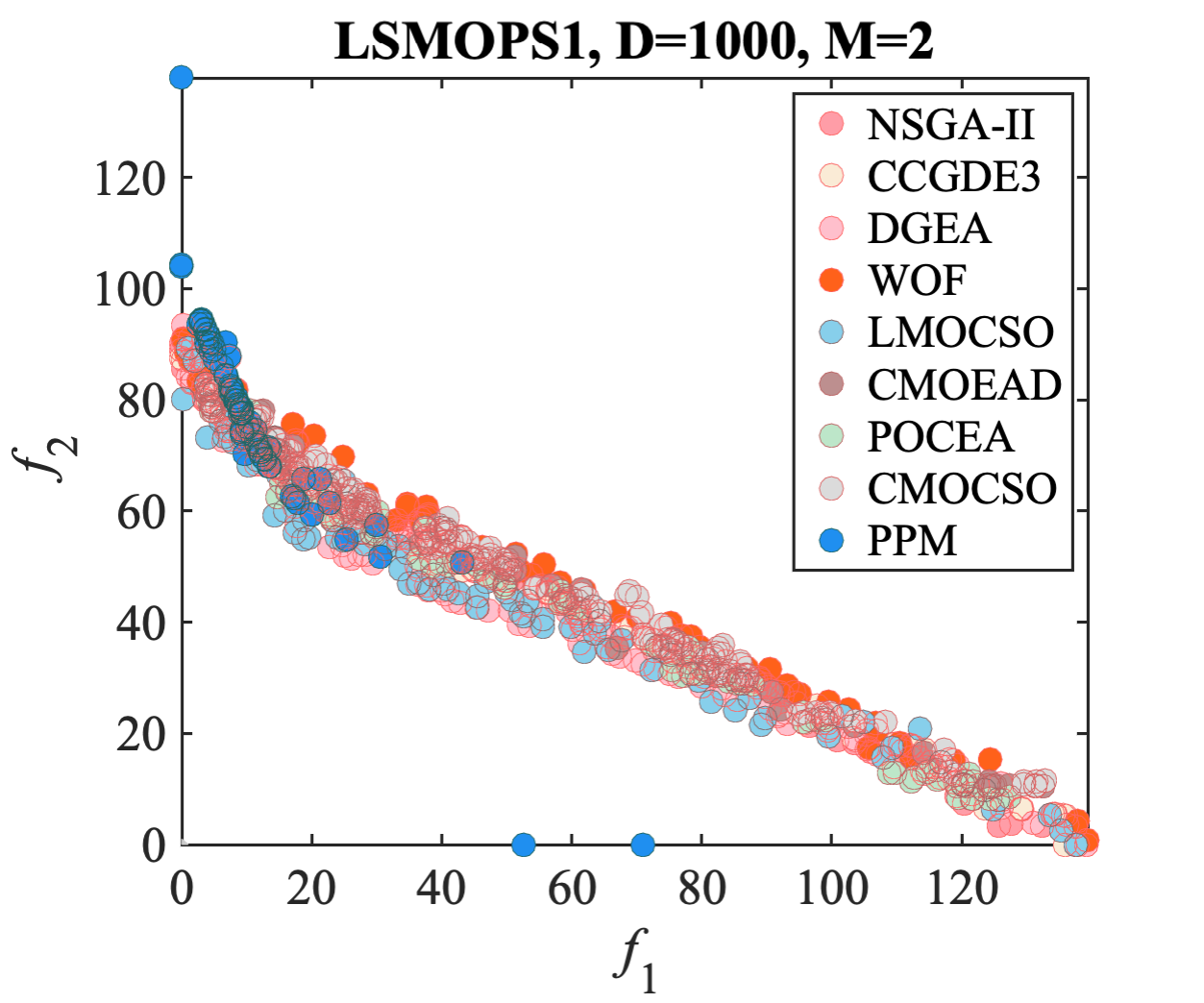}}
    \subfloat[{Problem LSMOP2}]{\includegraphics[width=0.3\hsize]{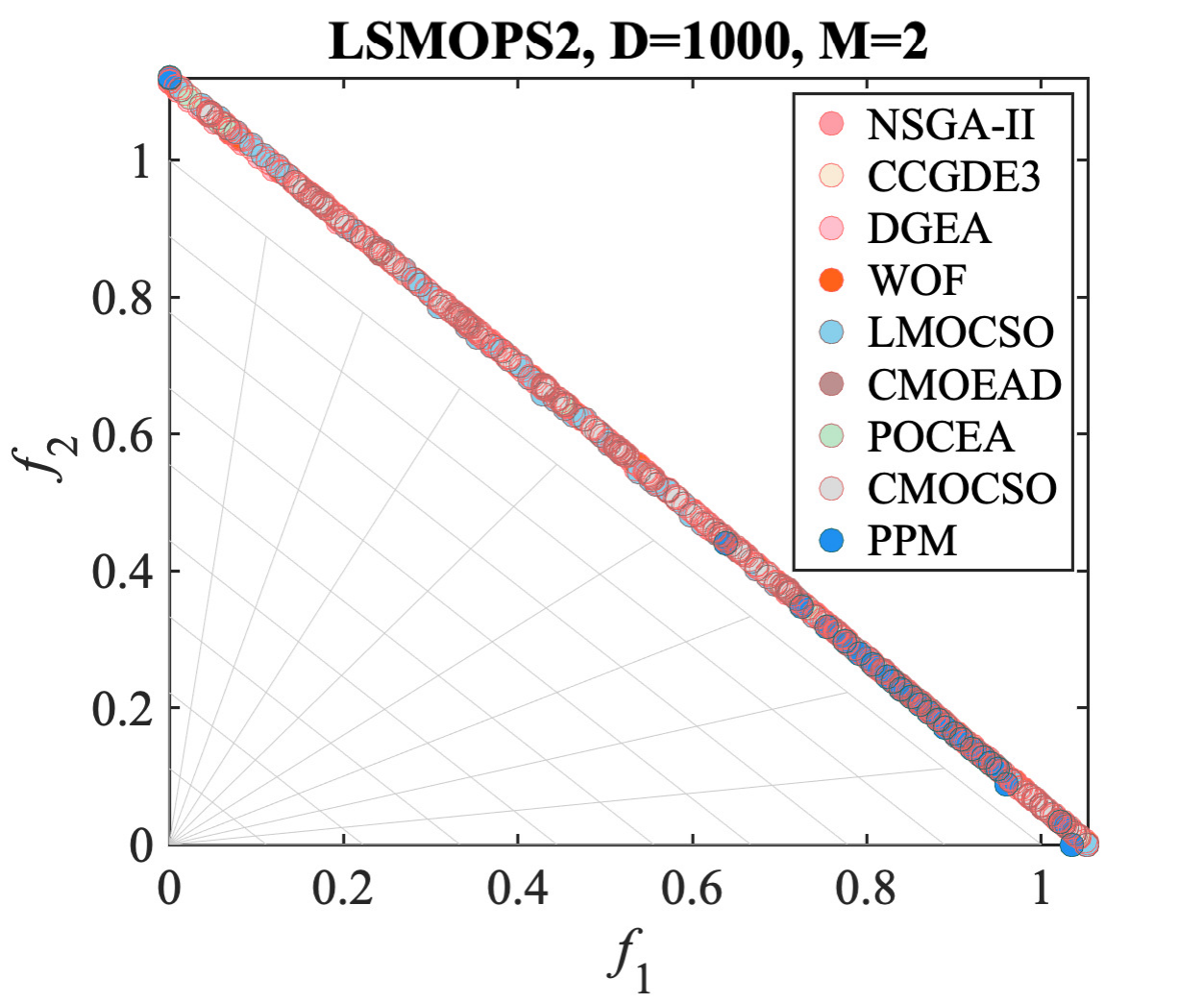}}
    \subfloat[{Problem LSMOP3}]{\includegraphics[width=0.3\hsize]{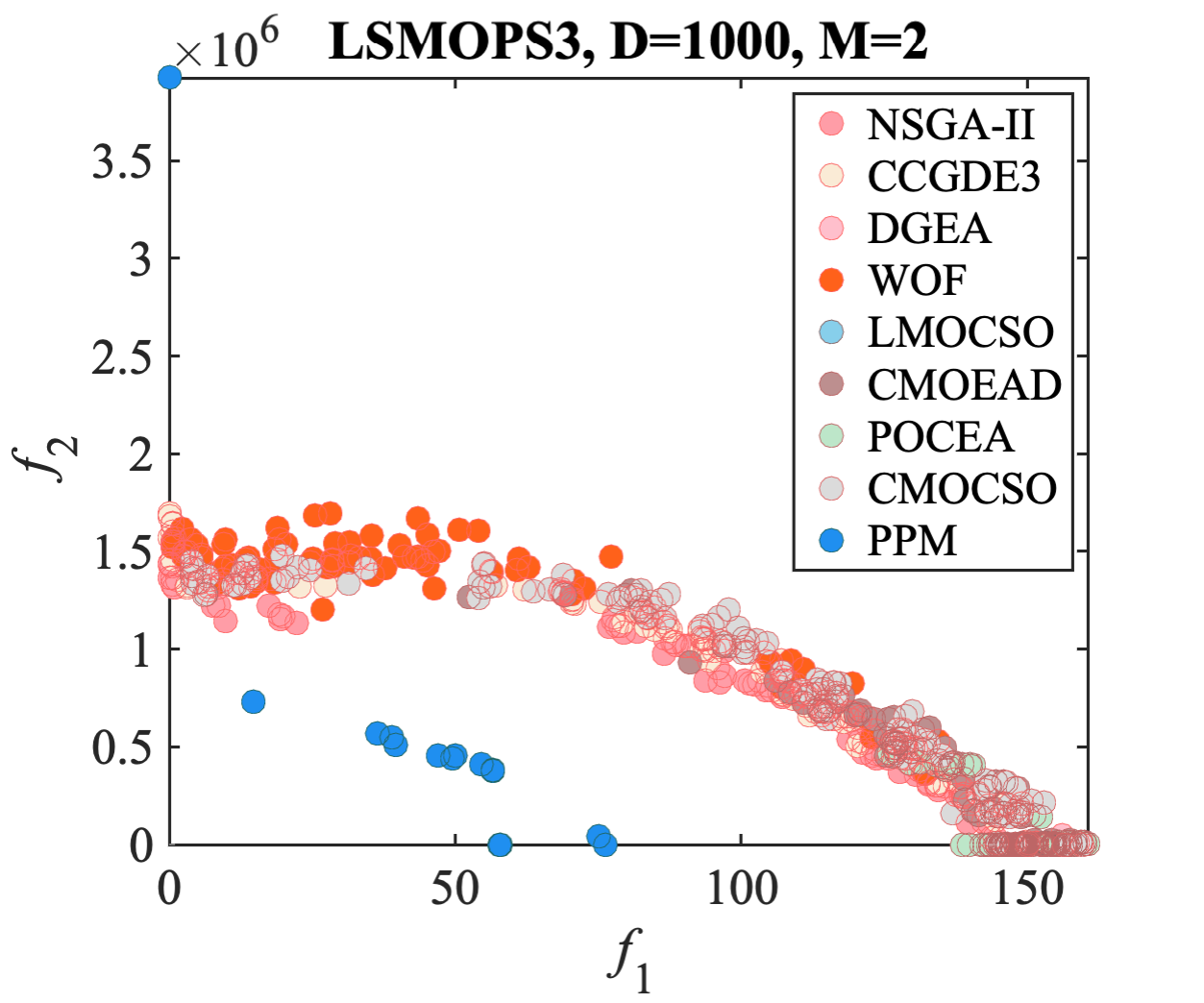}}
    \\
    \subfloat[{Problem LSMOP4}]{\includegraphics[width=0.3\hsize]{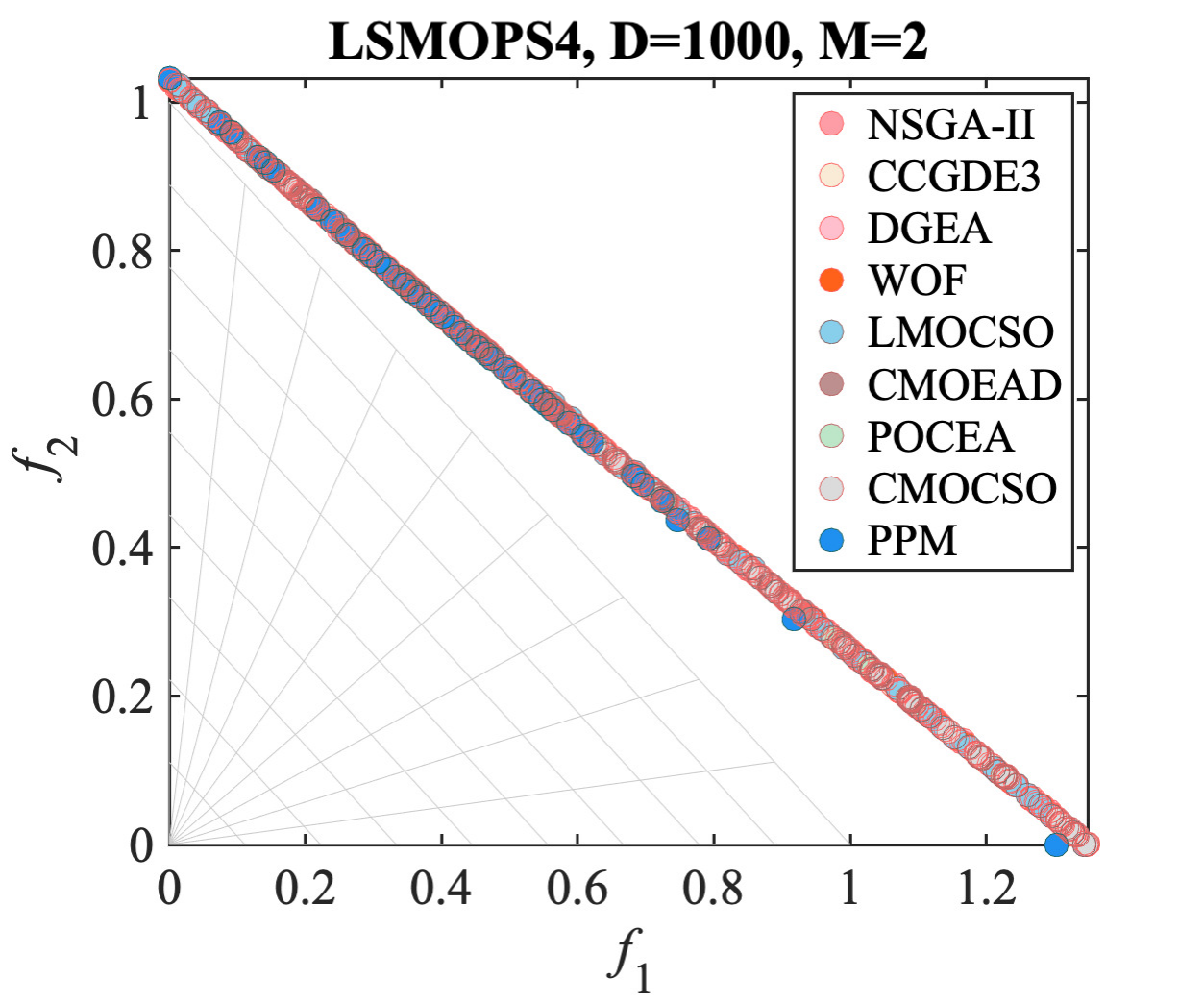}}
    \subfloat[{Problem LSMOP5}]{\includegraphics[width=0.3\hsize]{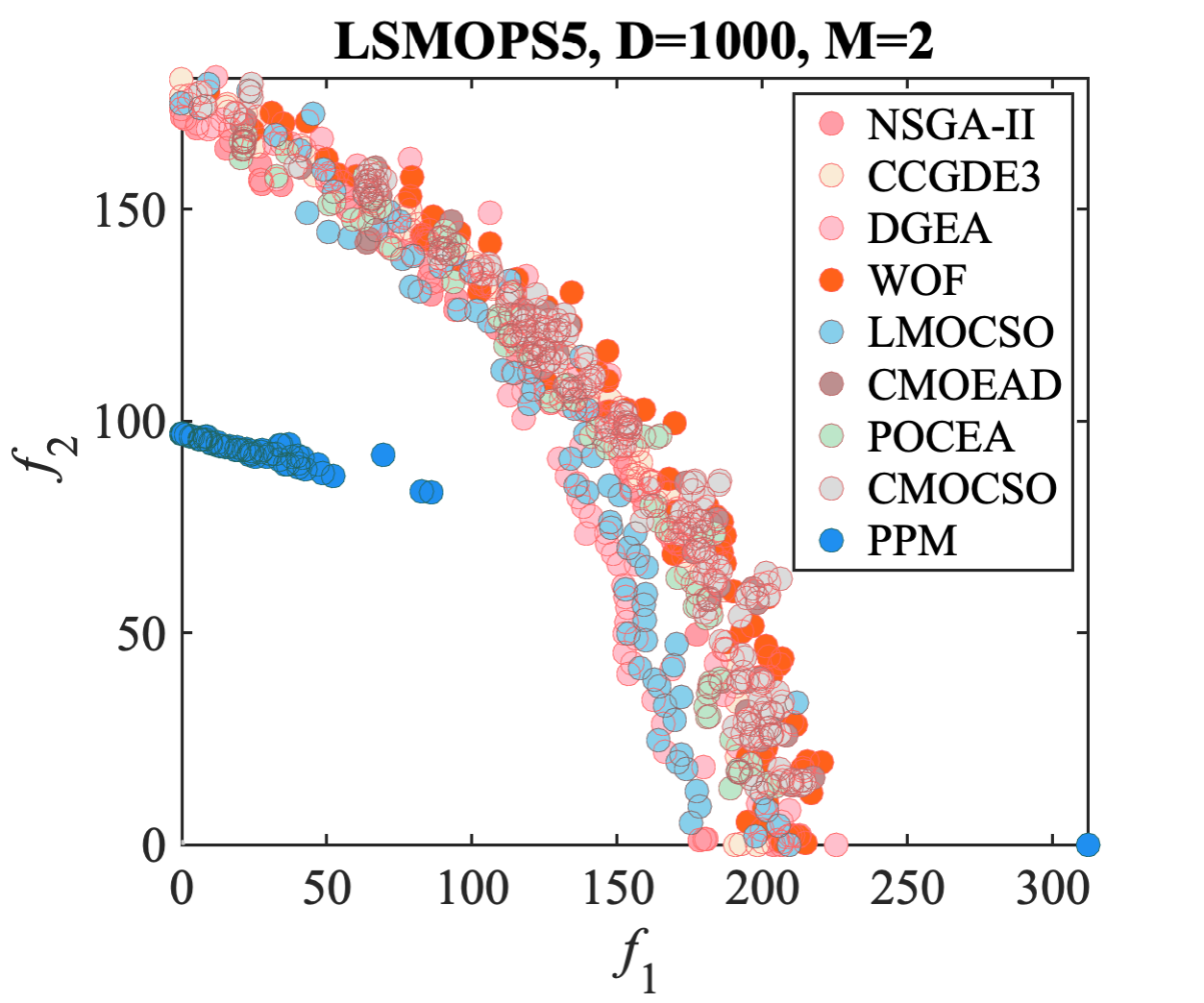}}
    \subfloat[{Problem LSMOP6}]{\includegraphics[width=0.3\hsize]{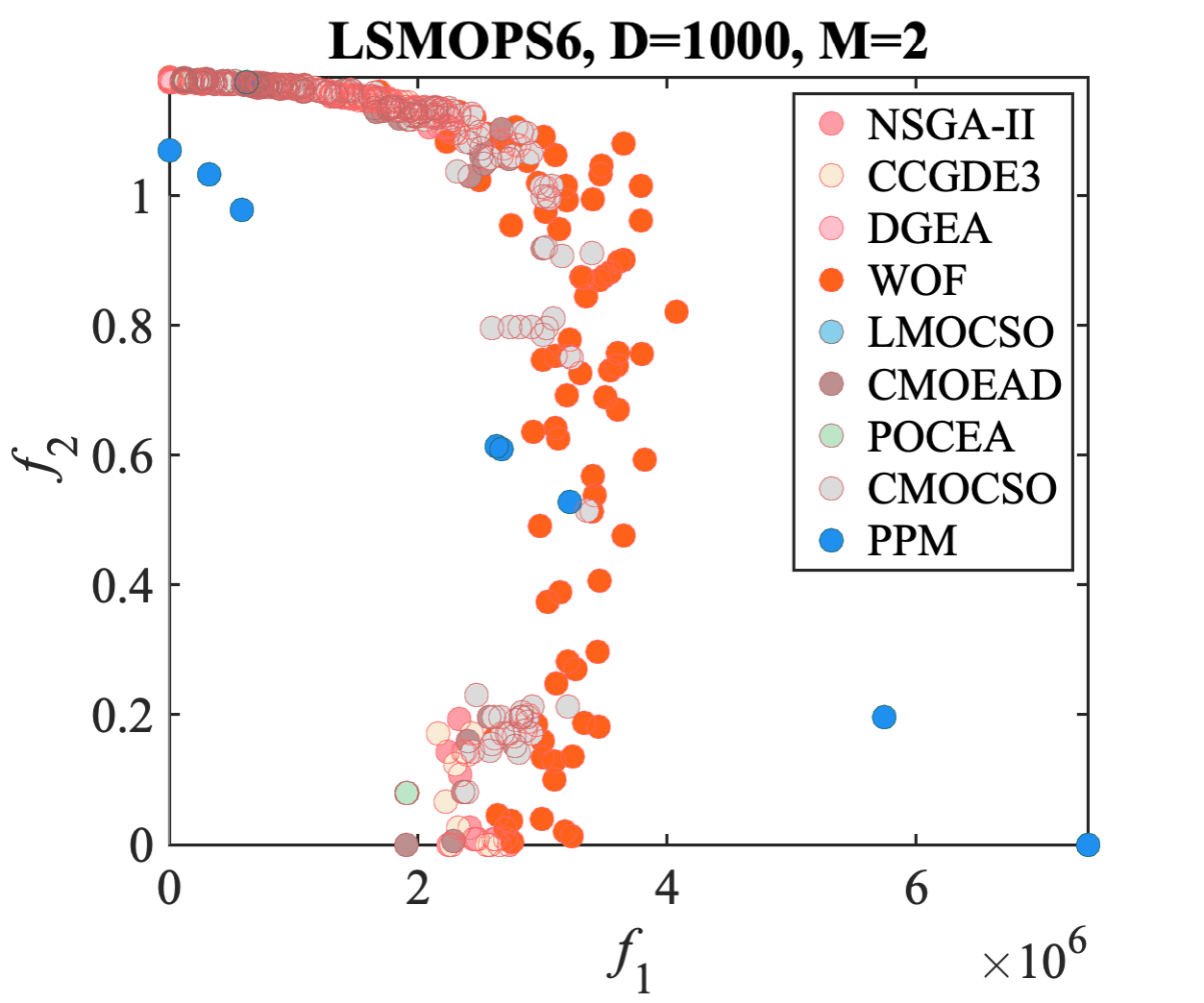}}
    \\
    \subfloat[{Problem LSMOP7}]{\includegraphics[width=0.3\hsize]{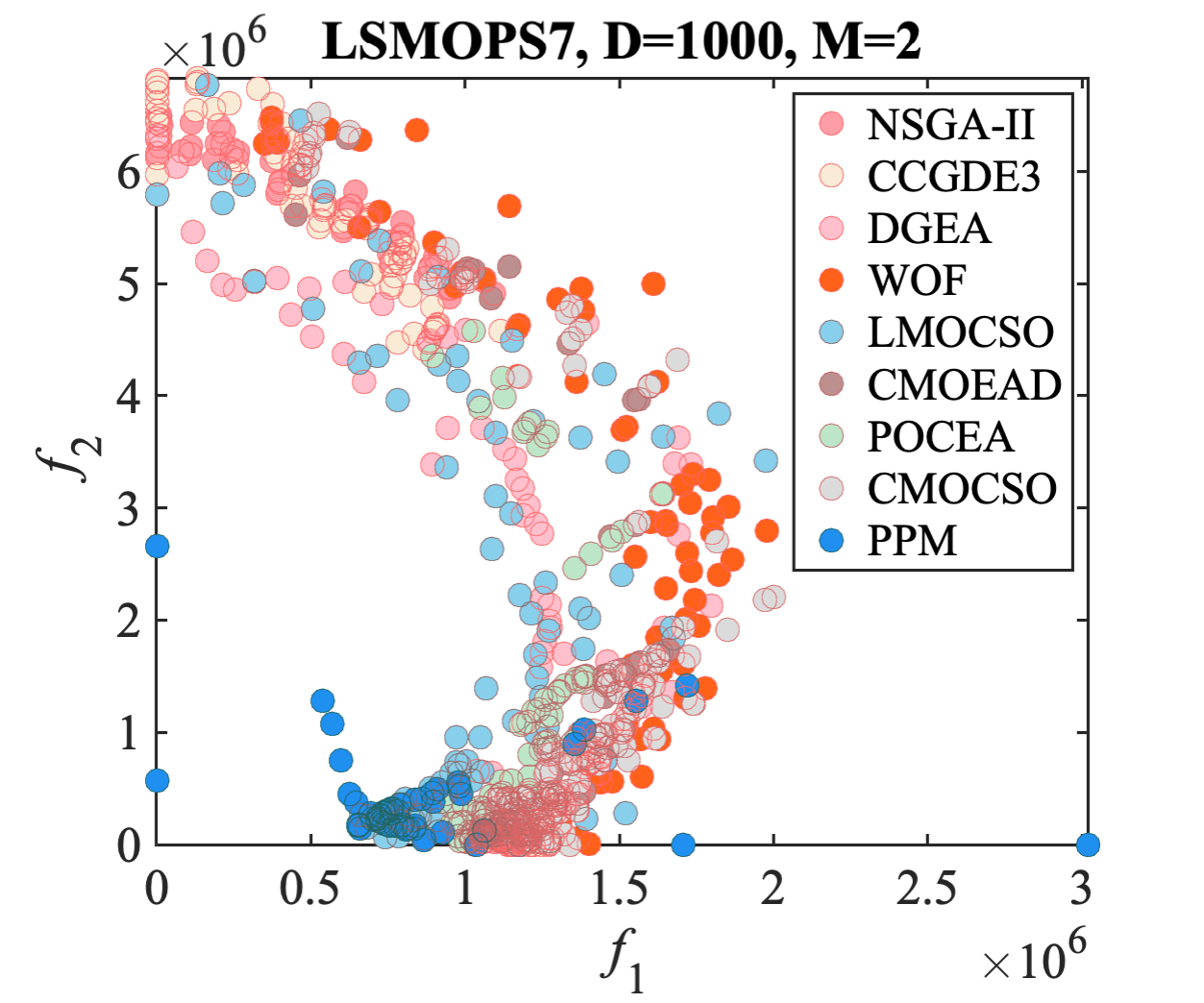}}
    \subfloat[{Problem LSMOP8}]{\includegraphics[width=0.3\hsize]{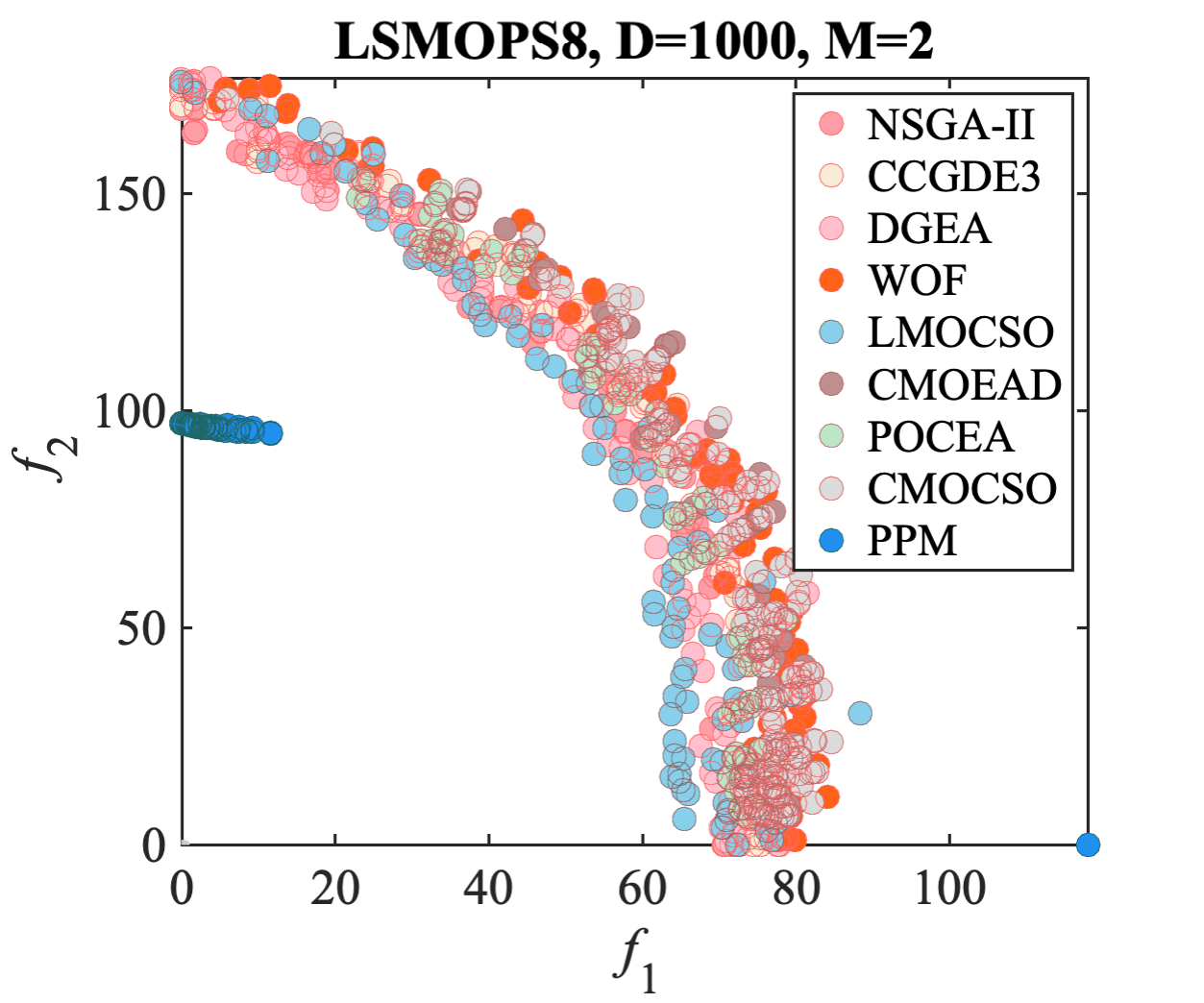}}
    \subfloat[{Problem LSMOP9}]{\includegraphics[width=0.3\hsize]{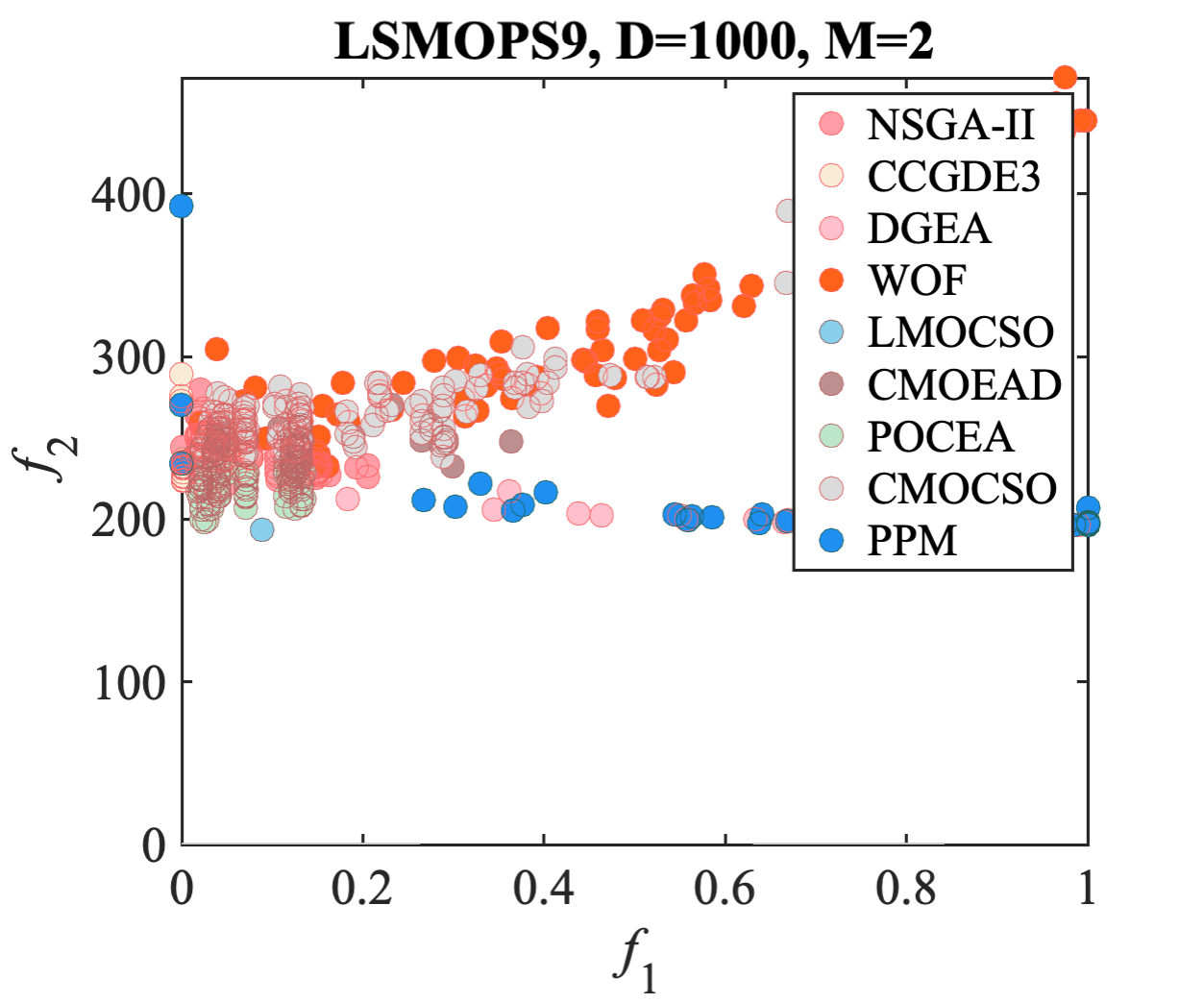}}
    \caption{Visualization of Non-dominated Solutions Obtained by Each Algorithm on 1000 Dimensional Bi-objective LSMOP*1 to LSMOP*9.}
    \label{fig: lsmop*1000}
\end{figure*}

\begin{figure*}[htbp]
    \centering
    \subfloat[{Problem LSMOP1}]{\includegraphics[width=0.3\hsize]{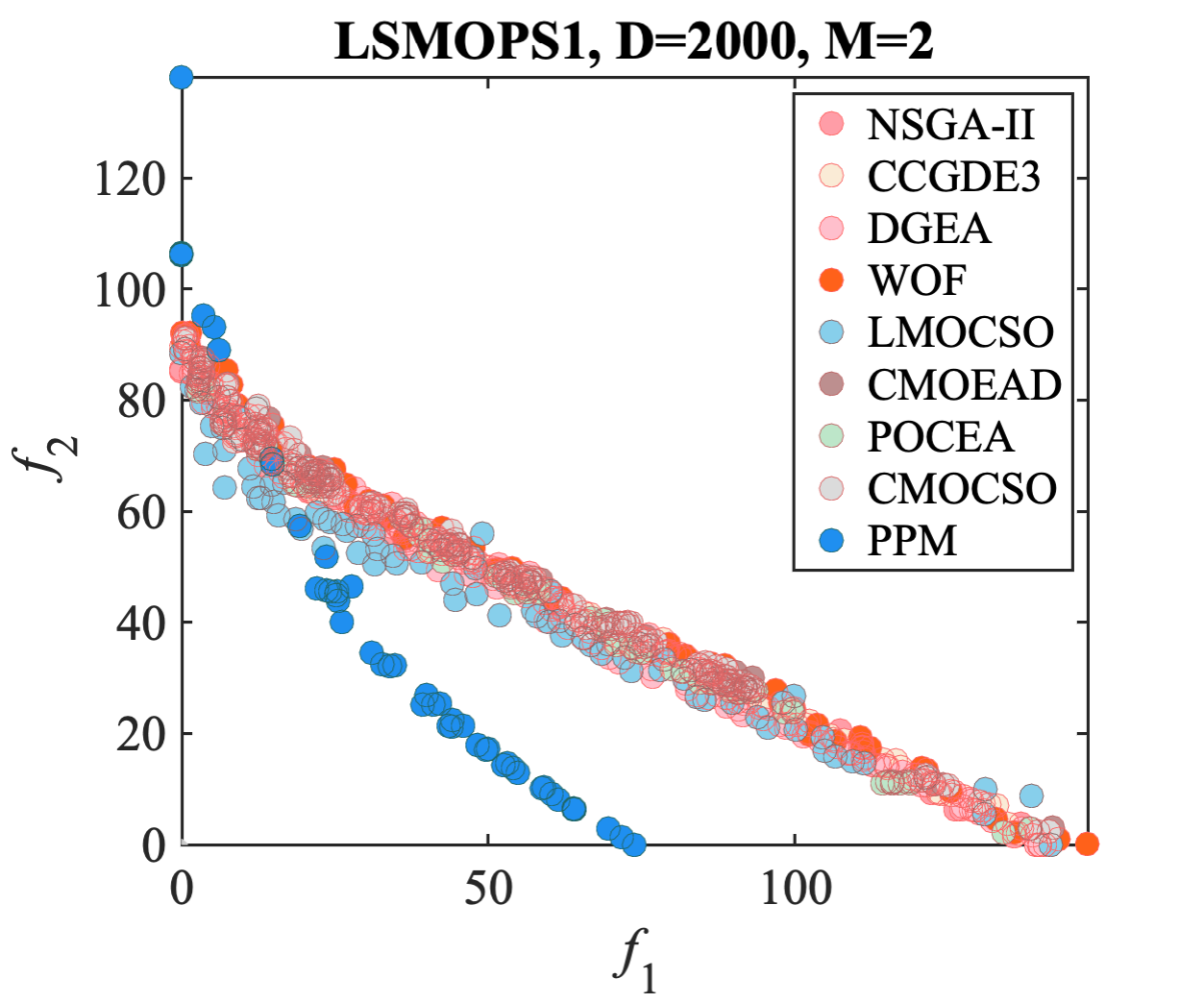}}
    \subfloat[{Problem LSMOP2}]{\includegraphics[width=0.3\hsize]{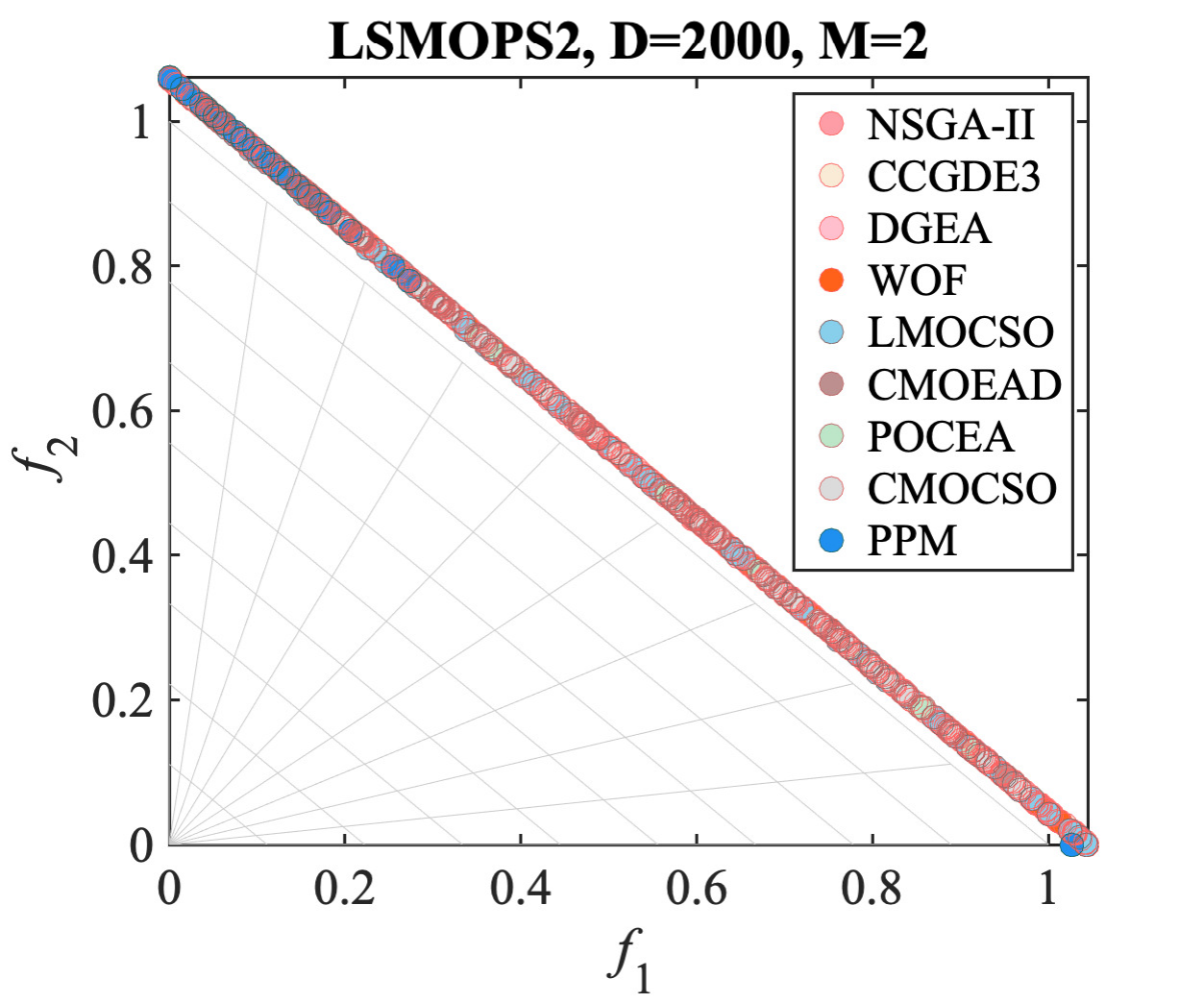}}
    \subfloat[{Problem LSMOP3}]{\includegraphics[width=0.3\hsize]{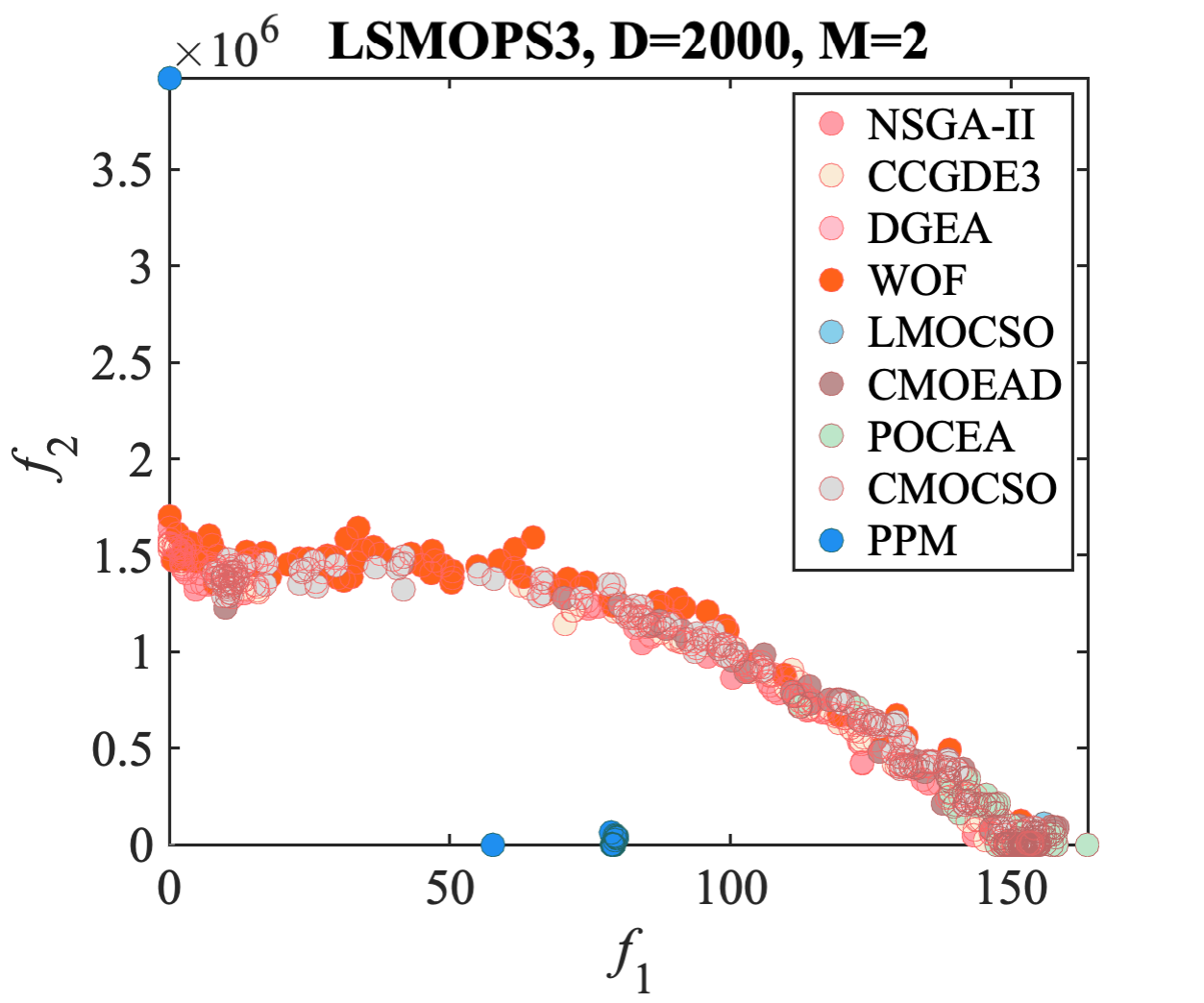}}
    \\
    \subfloat[{Problem LSMOP4}]{\includegraphics[width=0.3\hsize]{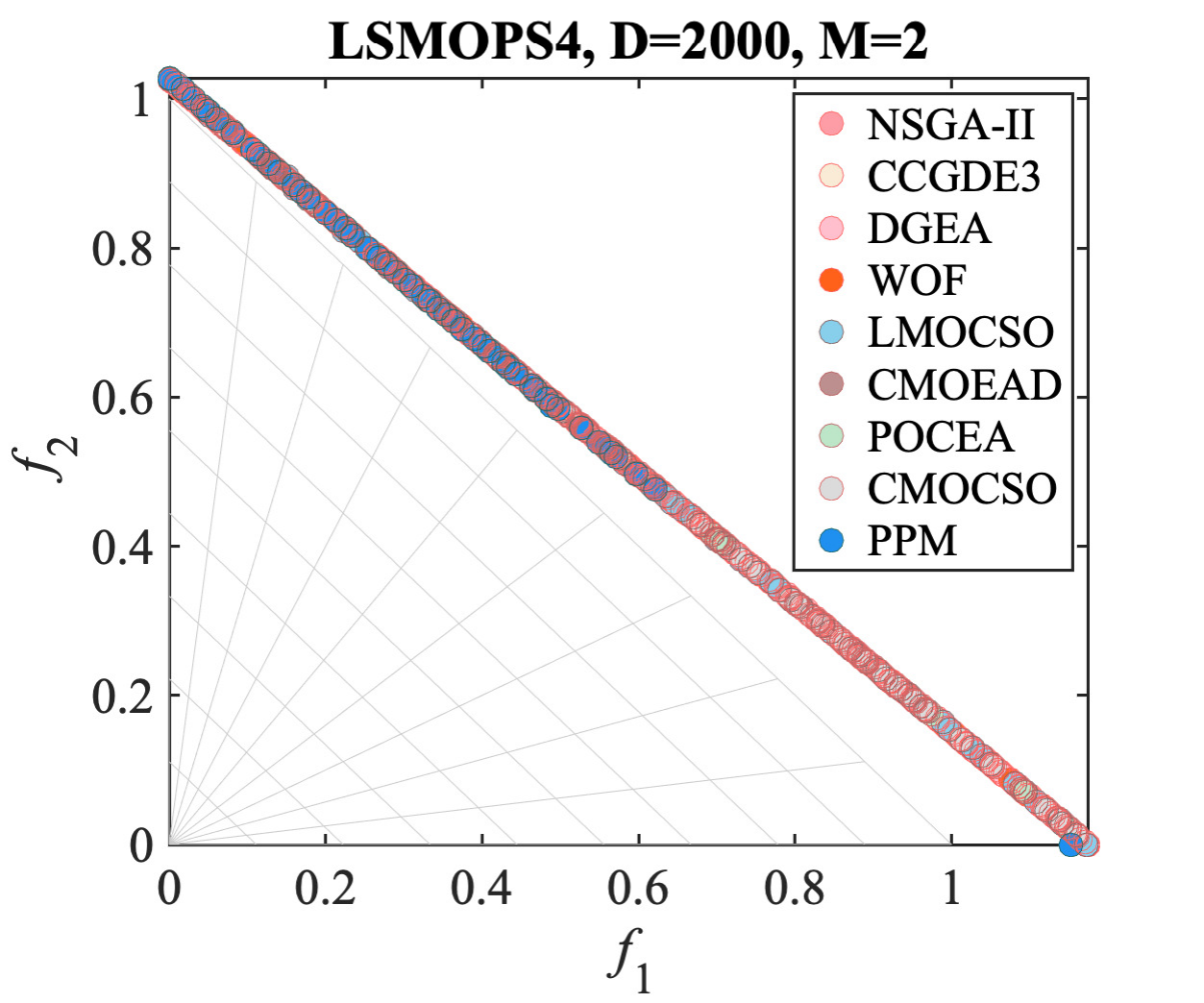}}
    \subfloat[{Problem LSMOP5}]{\includegraphics[width=0.3\hsize]{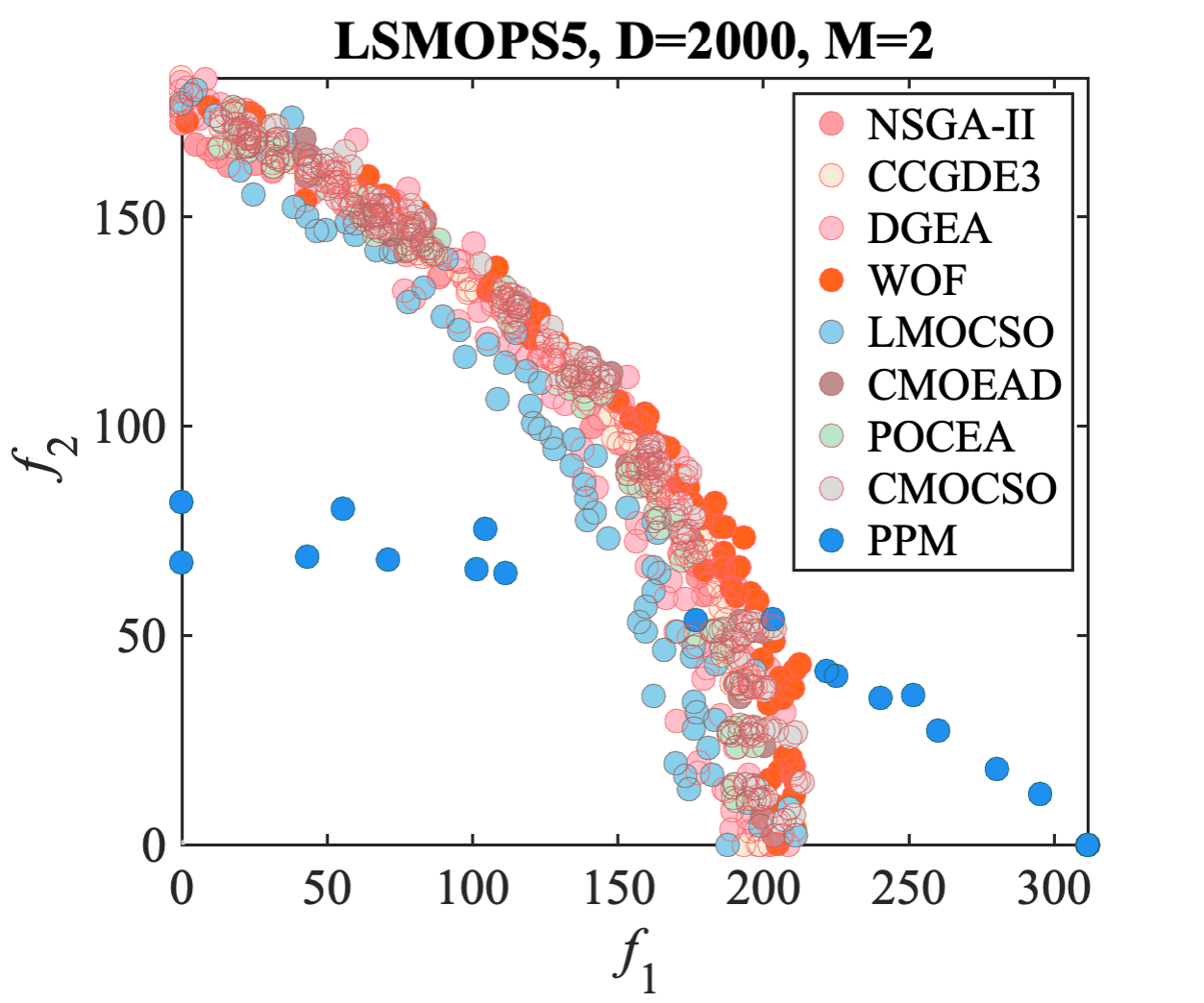}}
    \subfloat[{Problem LSMOP6}]{\includegraphics[width=0.3\hsize]{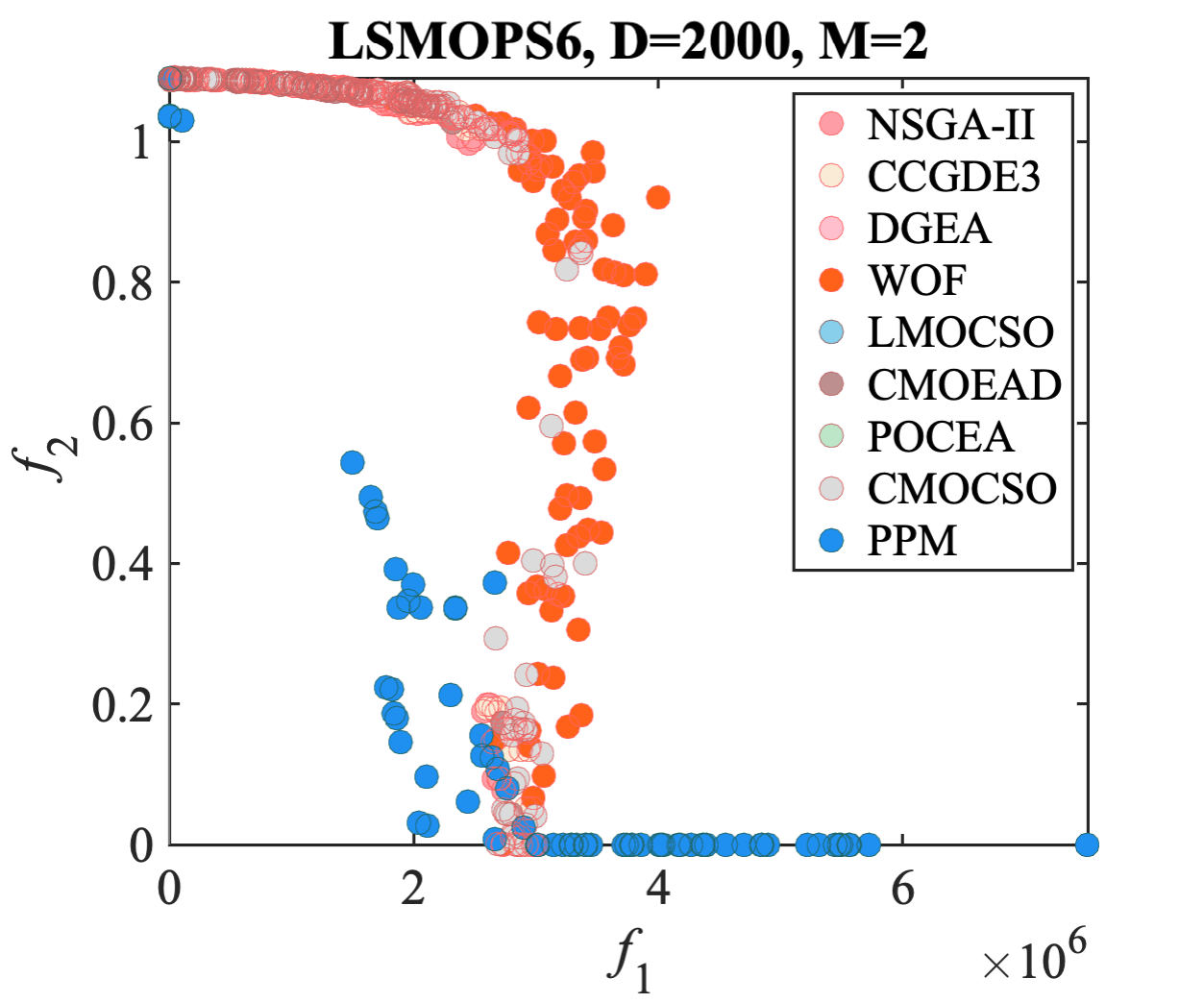}}
    \\
    \subfloat[{Problem LSMOP7}]{\includegraphics[width=0.3\hsize]{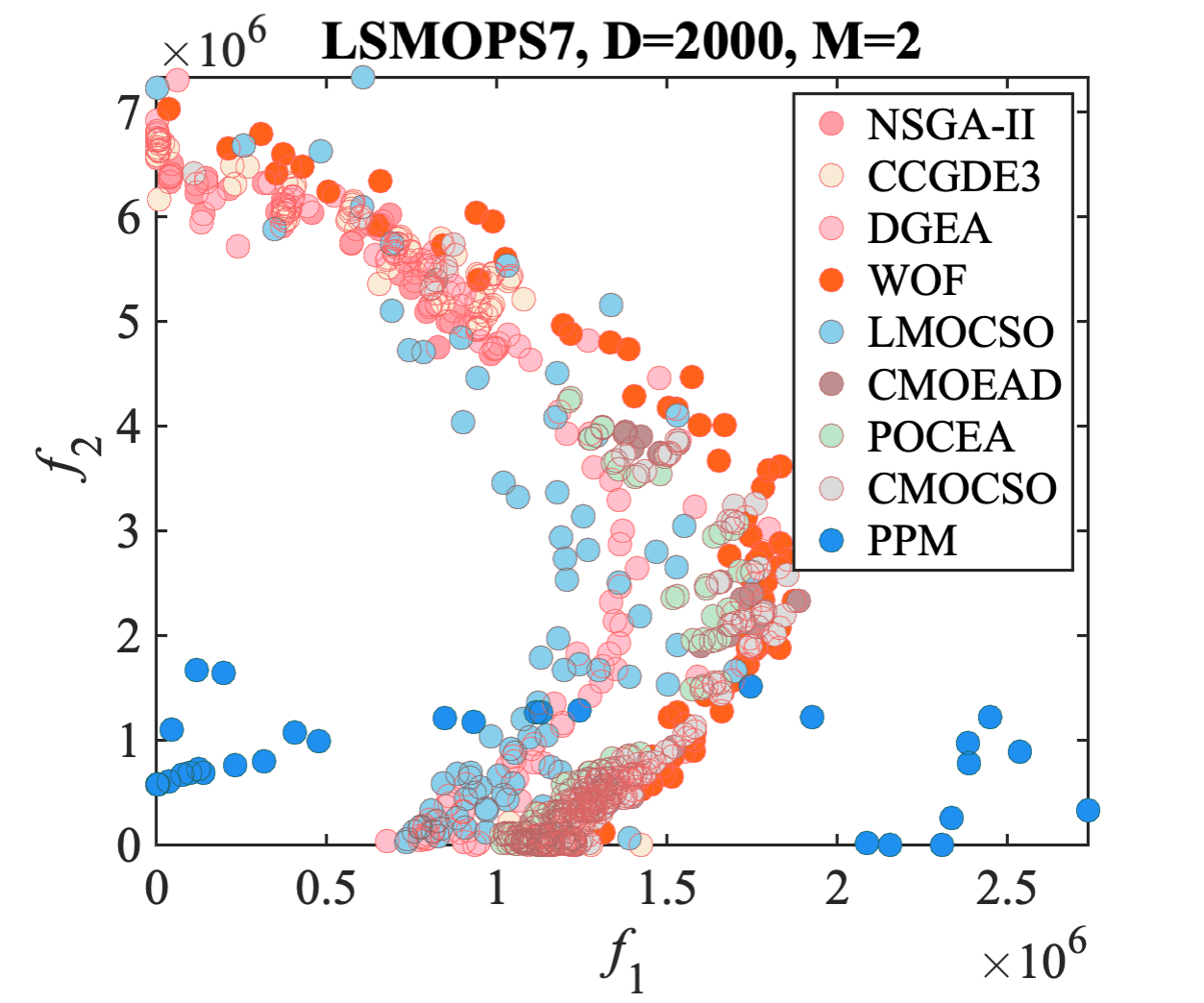}}
    \subfloat[{Problem LSMOP8}]{\includegraphics[width=0.3\hsize]{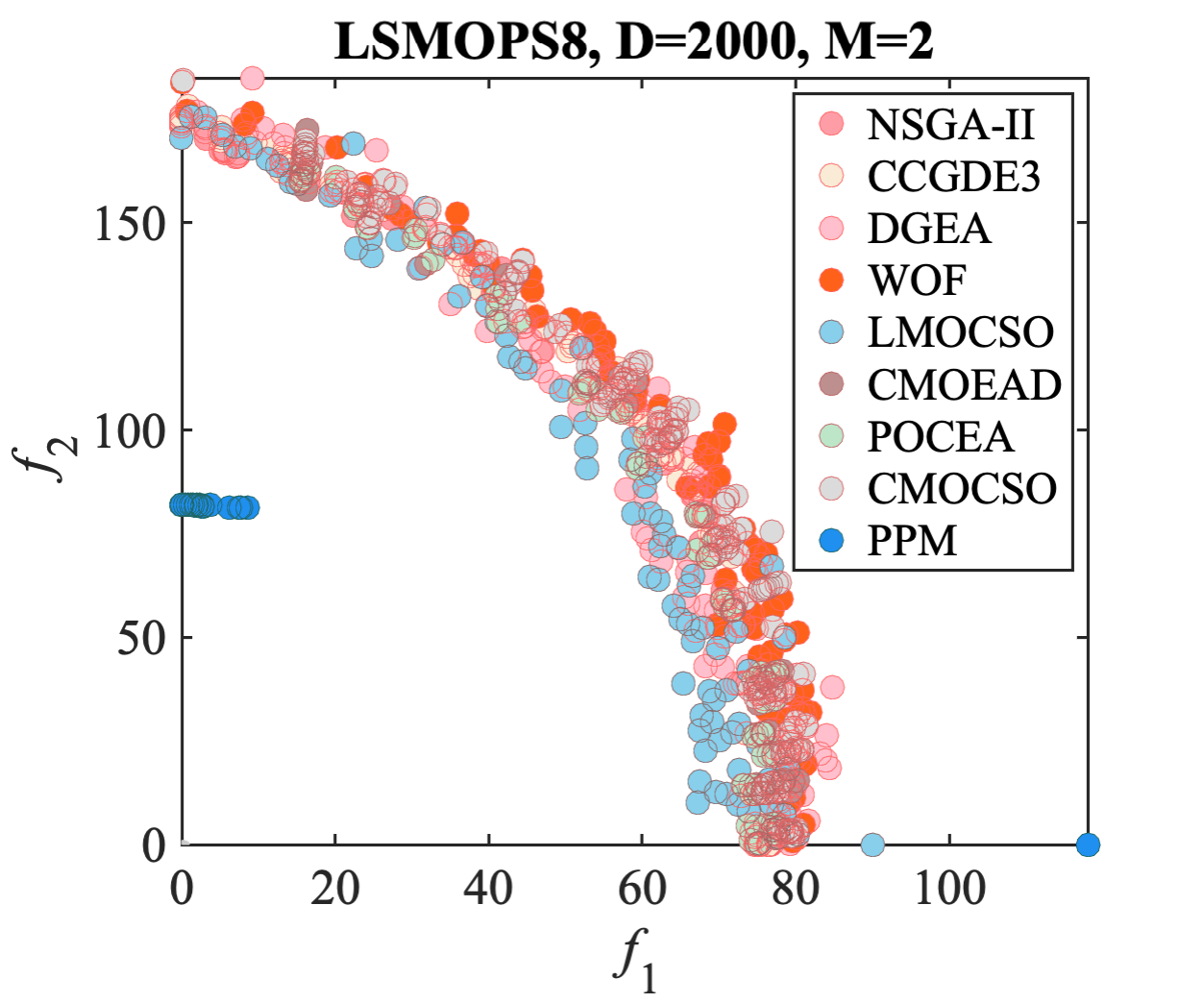}}
    \subfloat[{Problem LSMOP9}]{\includegraphics[width=0.3\hsize]{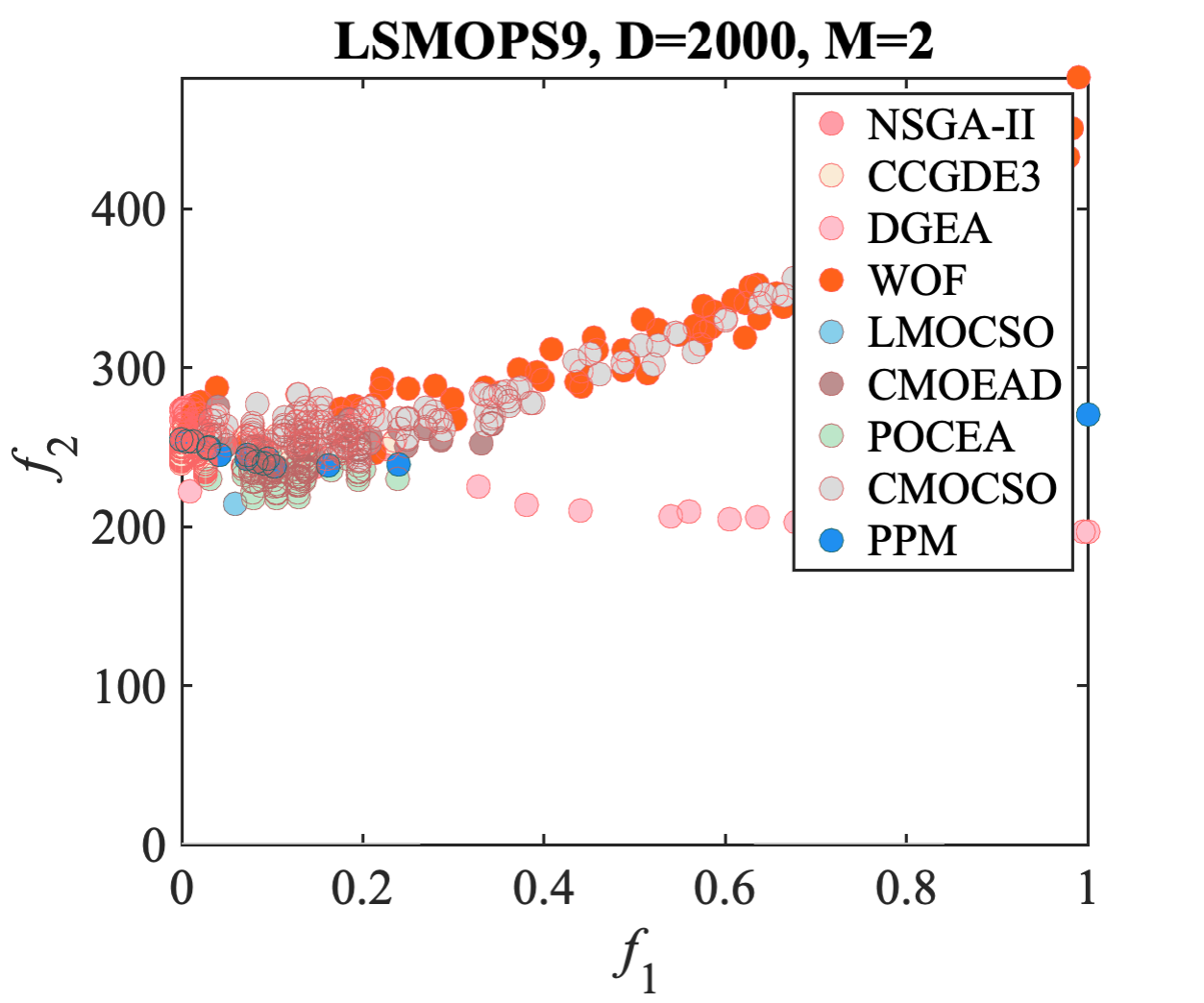}}
    \caption{Visualization of Non-dominated Solutions Obtained by Each Algorithm on 2000 Dimensional Bi-objective LSMOP*1 to LSMOP*9.}
    \label{fig: lsmop*2000}
\end{figure*}

\begin{figure*}[htbp]
    \centering
    \subfloat[{Problem LSMOP1}]{\includegraphics[width=0.3\hsize]{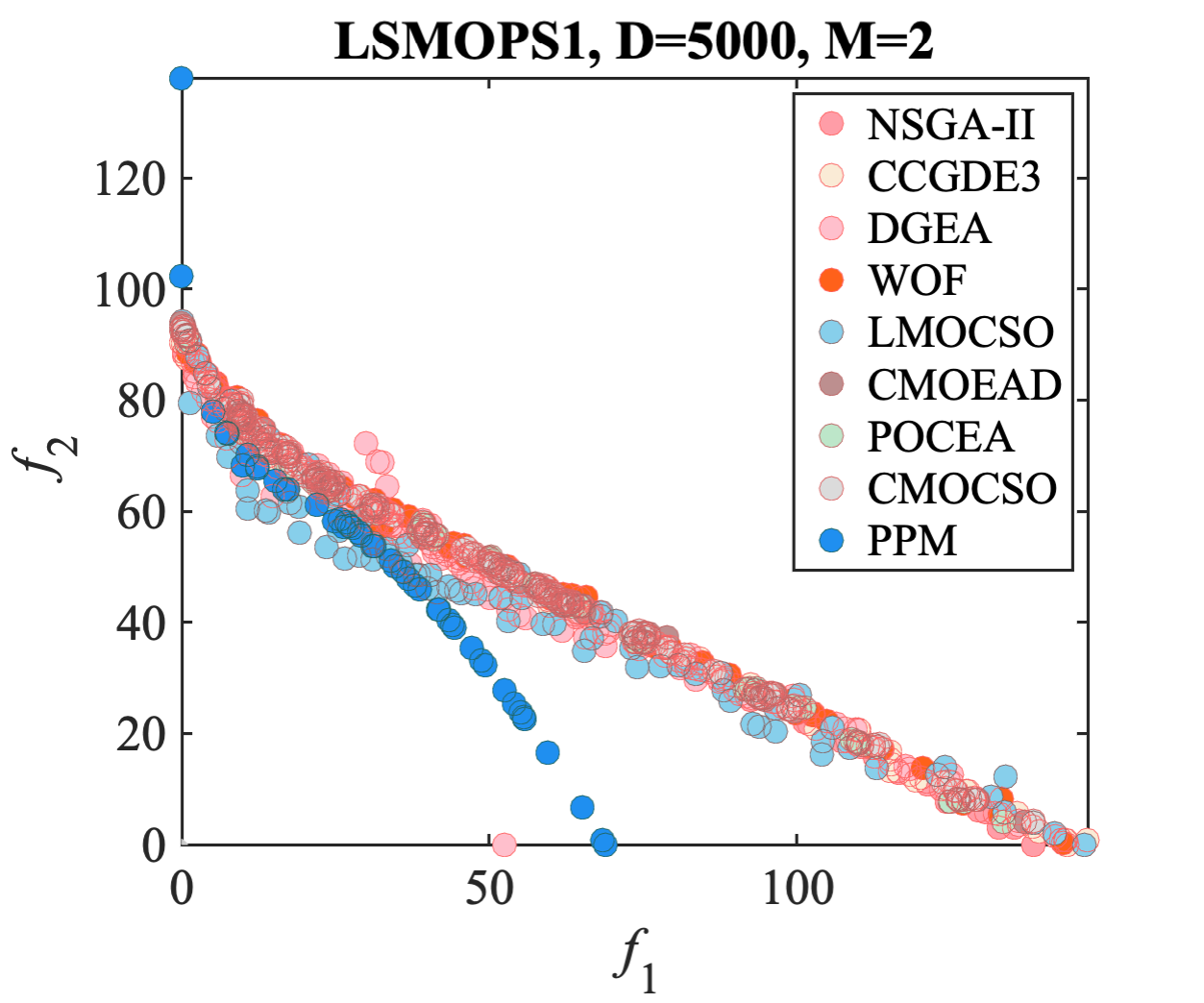}}
    \subfloat[{Problem LSMOP2}]{\includegraphics[width=0.3\hsize]{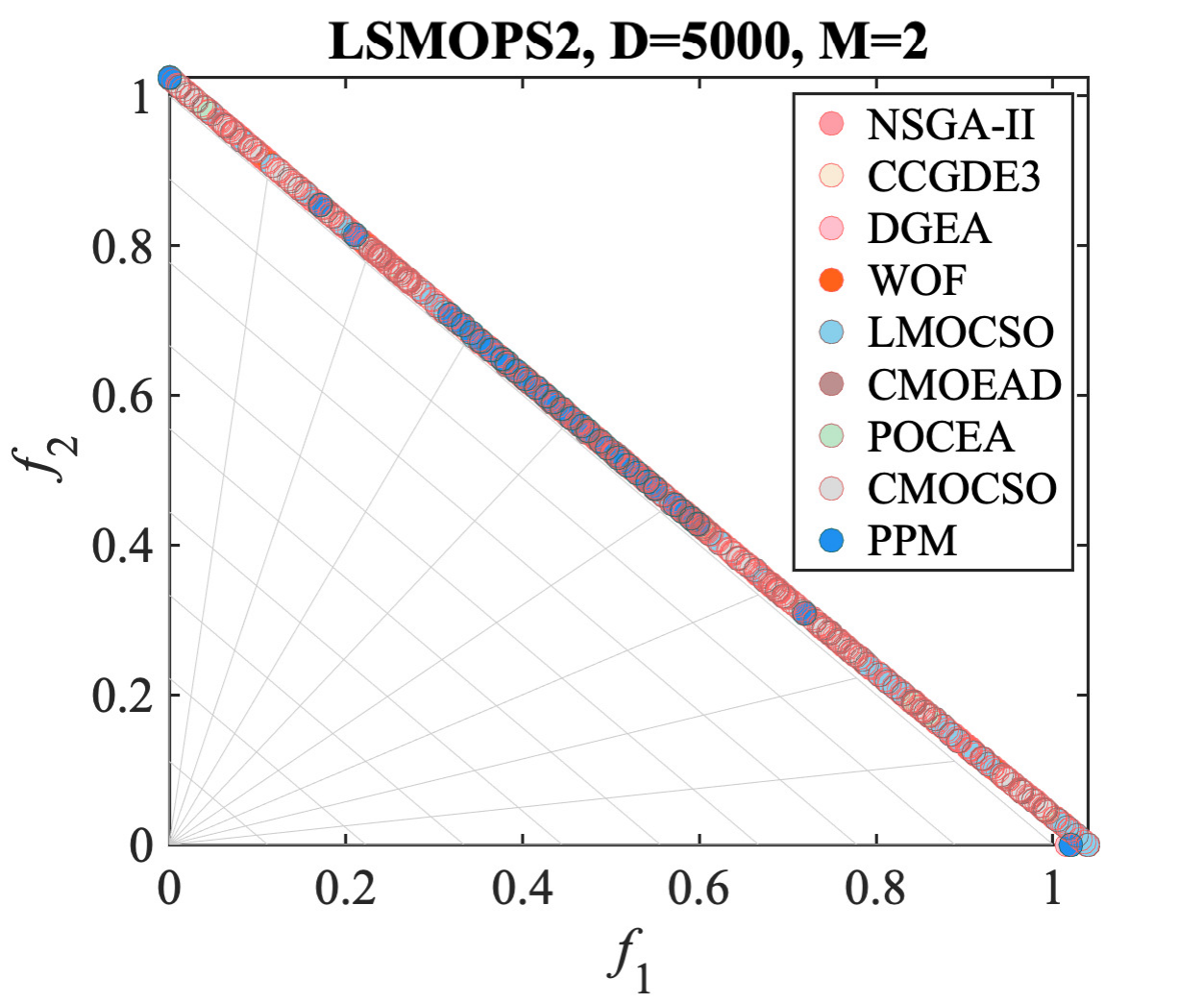}}
    \subfloat[{Problem LSMOP3}]{\includegraphics[width=0.3\hsize]{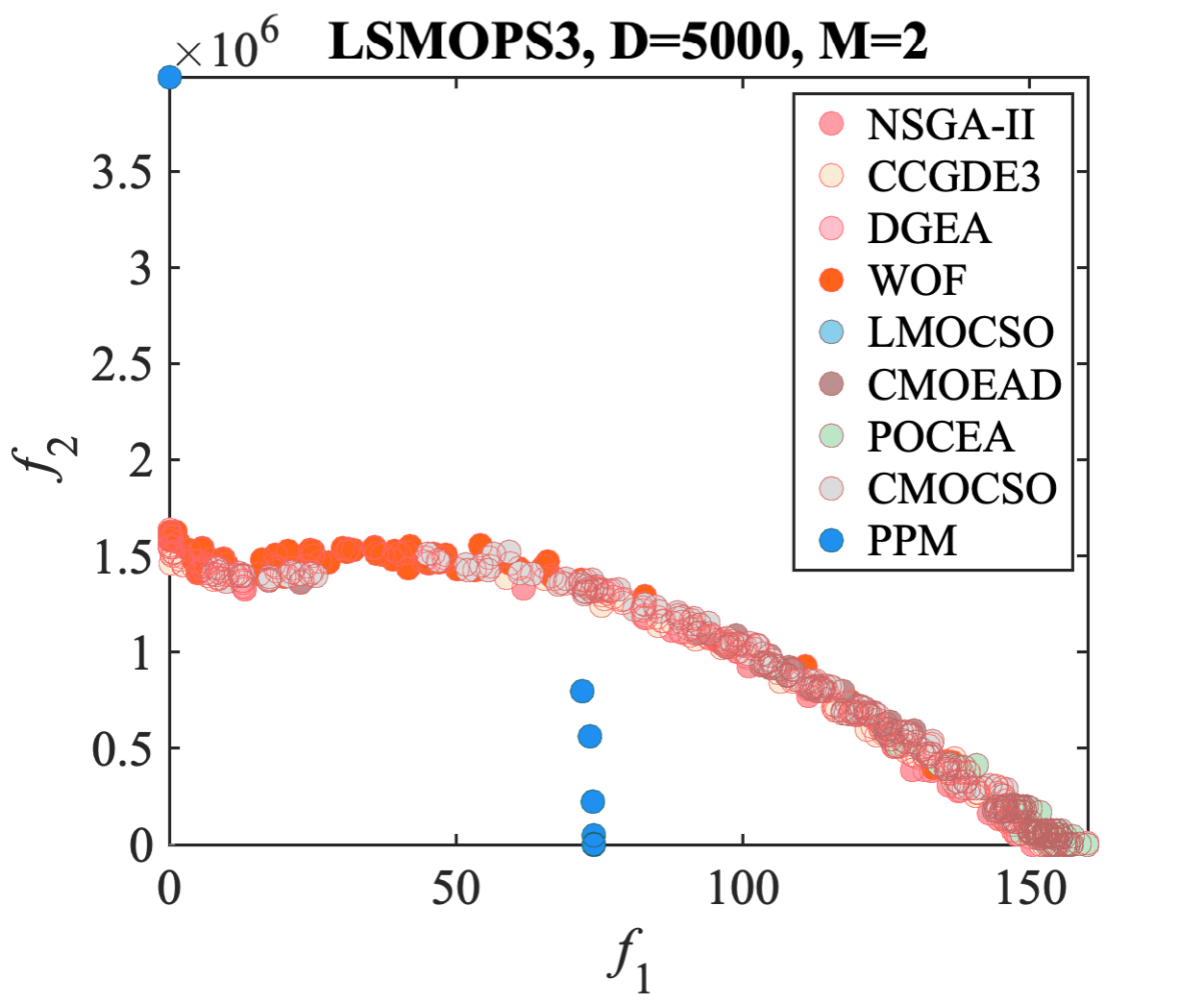}}
    \\
    \subfloat[{Problem LSMOP4}]{\includegraphics[width=0.3\hsize]{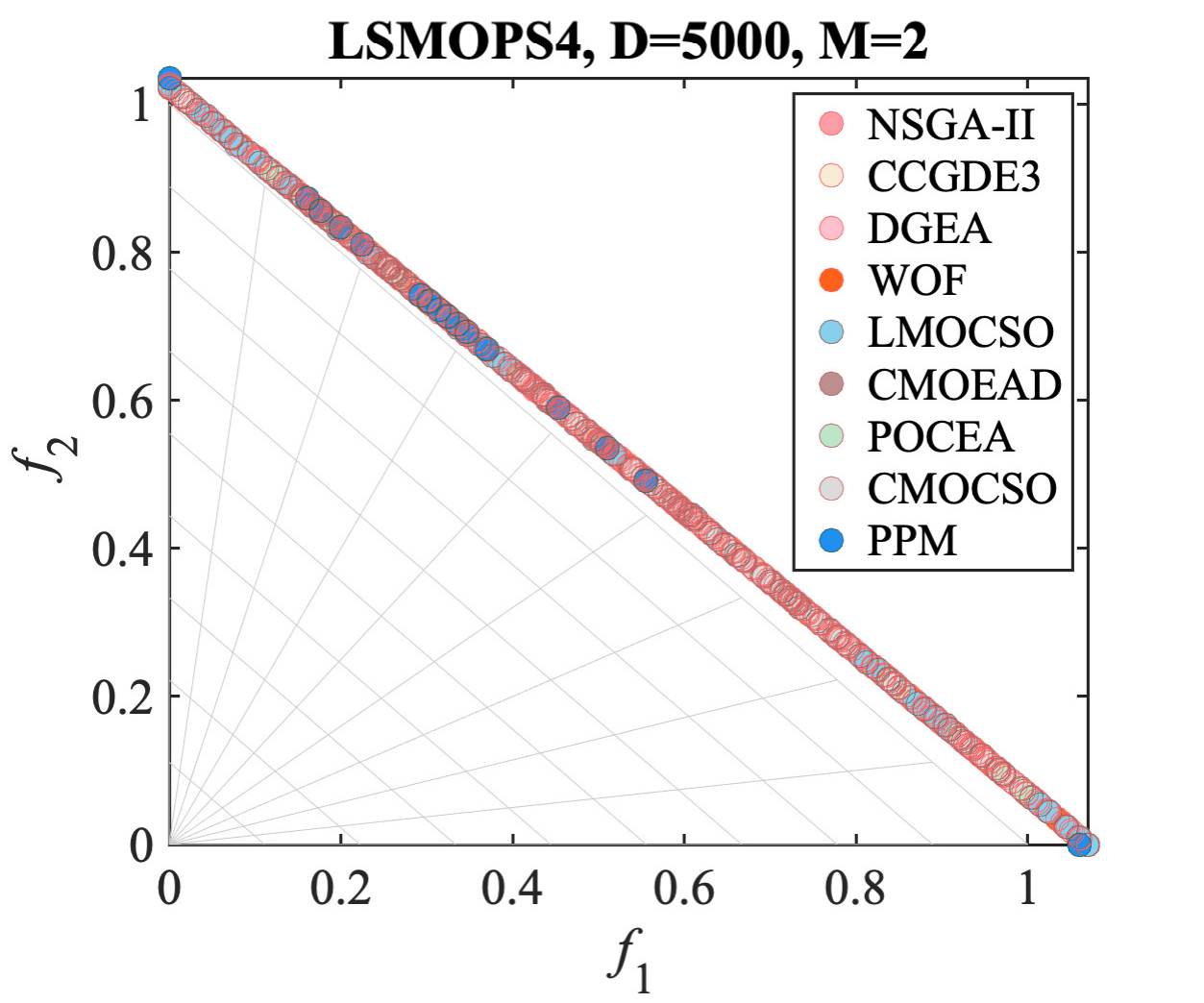}}
    \subfloat[{Problem LSMOP5}]{\includegraphics[width=0.3\hsize]{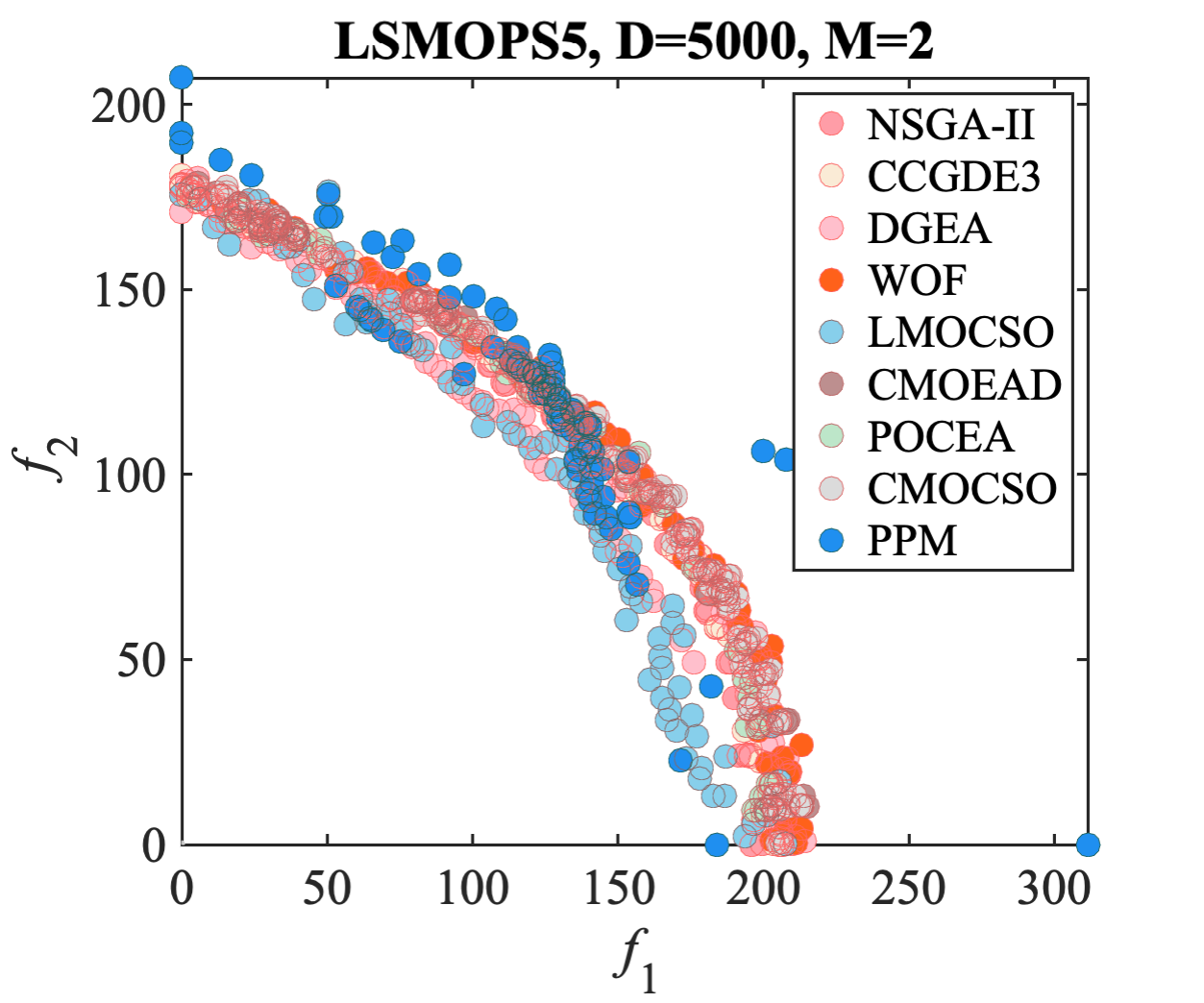}}
    \subfloat[{Problem LSMOP6}]{\includegraphics[width=0.3\hsize]{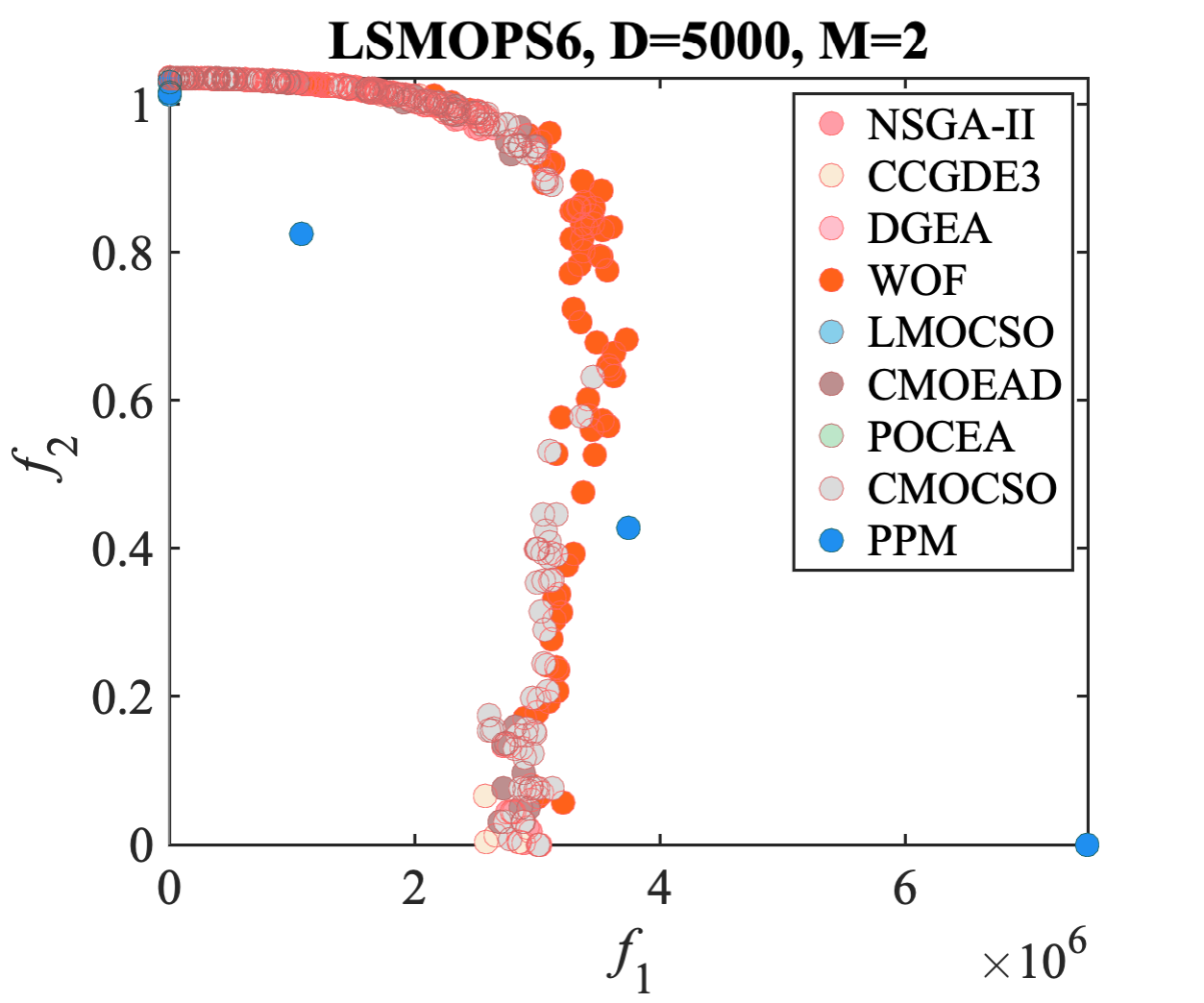}}
    \\
    \subfloat[{Problem LSMOP7}]{\includegraphics[width=0.3\hsize]{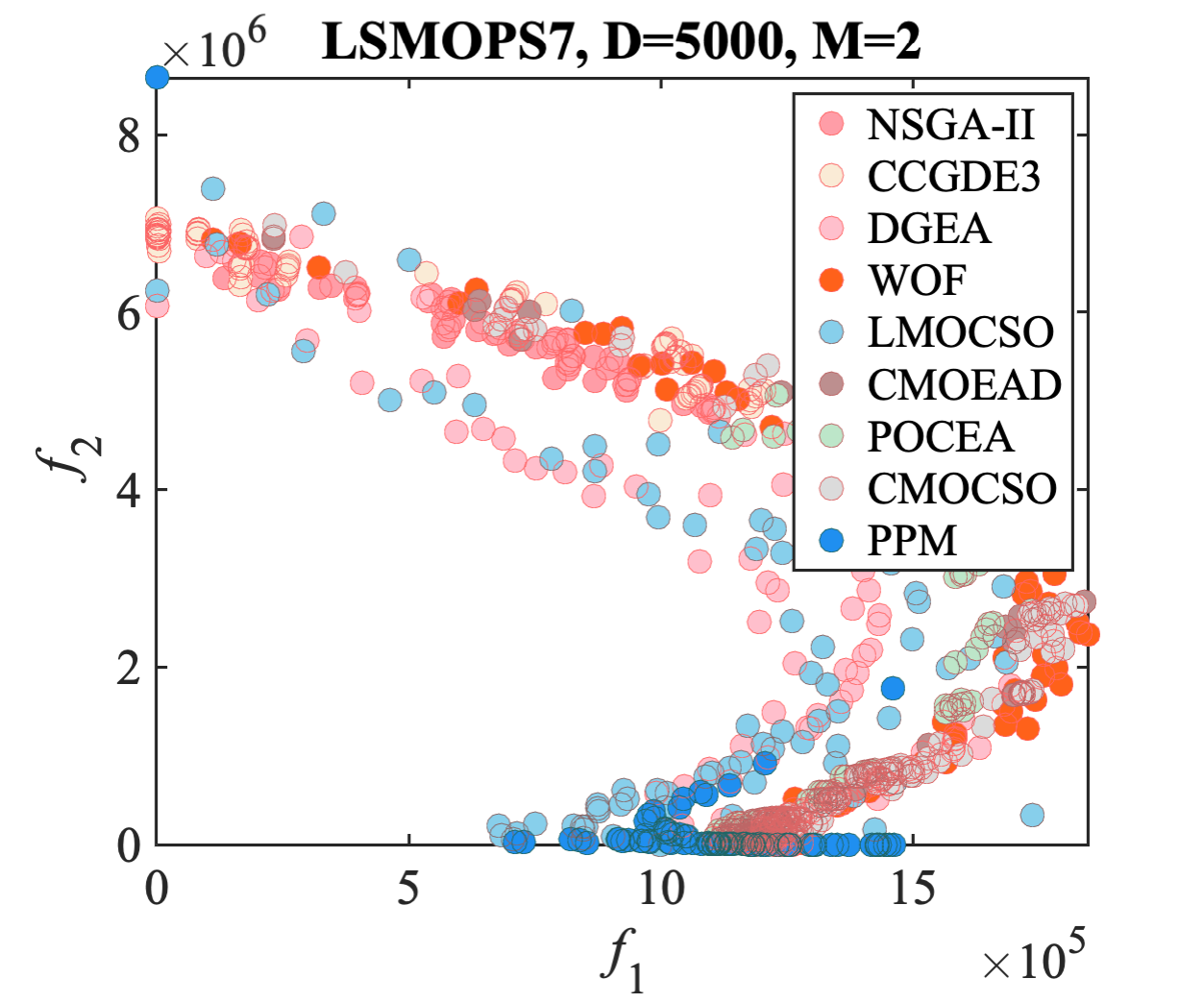}}
    \subfloat[{Problem LSMOP8}]{\includegraphics[width=0.3\hsize]{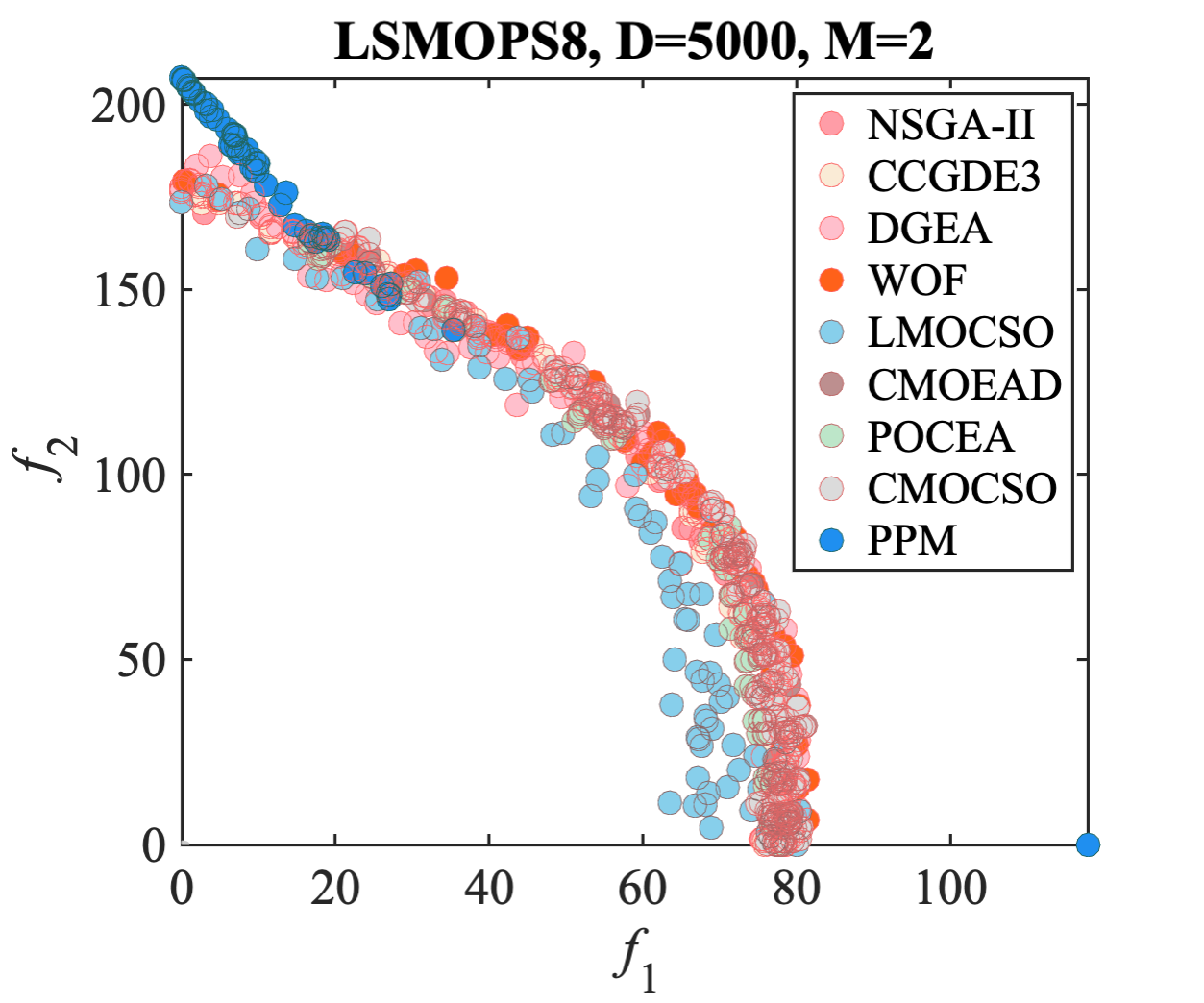}}
    \subfloat[{Problem LSMOP9}]{\includegraphics[width=0.3\hsize]{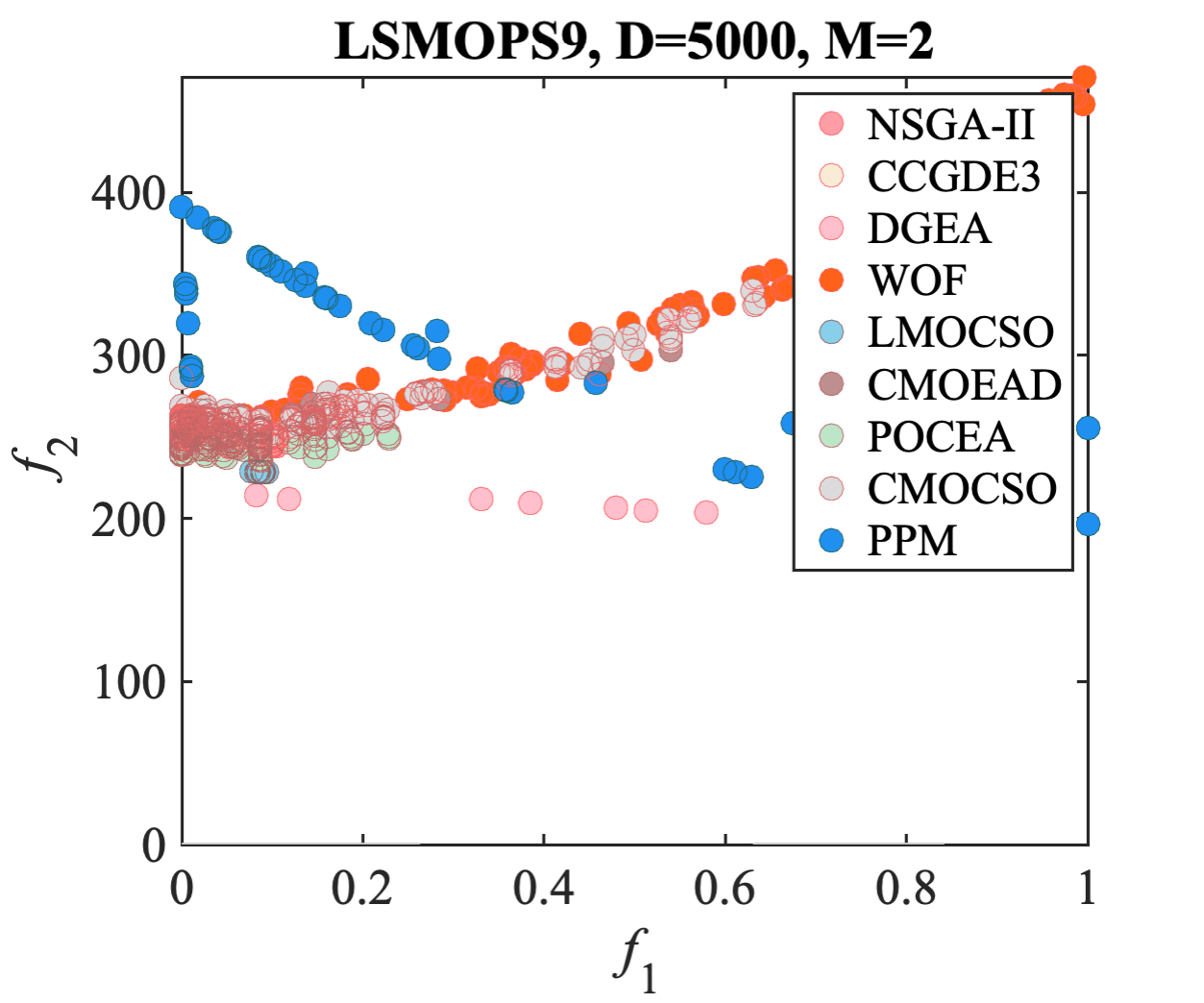}}
    \caption{Visualization of Non-dominated Solutions Obtained by Each Algorithm on Bi-objective LSMOP*1 to LSMOP*9.}
    \label{fig: lsmop*5000}
\end{figure*}

We visualize the nondominated solution sets obtained by all algorithms on LSMOP* in Figures \ref{fig: lsmop*1000}, \ref{fig: lsmop*2000}, and \ref{fig: lsmop*5000}. As can be seen from the figures, although the proposed PPM model has not been trained on LSMOP*1-9, compared with existing algorithms, it can locate solutions closer to the Pareto optimal under expensive constraints.

\end{document}